\documentclass[namedate,webpdf,traditional,medium]{oup-authoring-template}

\onecolumn 

\graphicspath{{Fig/}}

\theoremstyle{thmstyleone}%
\newtheorem{theorem}{Theorem}
\newtheorem{proposition}[theorem]{Proposition}%
\theoremstyle{thmstyletwo}%
\newtheorem{remark}{Remark}%
\newtheorem{lemma}{Lemma}%
\theoremstyle{thmstylethree}%

\usepackage{url}

\usepackage{xspace}
\usepackage{enumerate}
\newcommand{\refAssump}[2]{(\textbf{#1#2})}

\usepackage{amsmath}
\usepackage{bm}
\newcommand\numberthis{\addtocounter{equation}{1}\tag{\theequation}}

\usepackage{amssymb}
\usepackage{pifont}
\newcommand{\cmark}{\ding{51}}%
\newcommand{\xmark}{\ding{55}}%

\newcommand{\E}{\mathbb{E}}
\newcommand{\Epi}{\mathbb{E}_{\pi}}

\newcommand{\R}{\mathbb{R}}
\newcommand{\Rtwo}{\mathbb{R}^2}

\newcommand{\B}{\mathcal{B}}

\newcommand{\M}{\mathcal{M}}
\newcommand{\pr}{\mathbb{P}}

\newcommand{\ActSet}{\Pi}

\newcommand{\Ei}{\bar{A}_i}
\newcommand{\ei}{\bar{a}_i}

\newcommand{\EiHat}{\widehat{\bar{A}}_i}

\newcommand{\muTrue}{\mu(\ei; \Xouti, \outCoef_0)}
\newcommand{\muGeneral}{\mu(\Ei; \Xouti, \outCoef)}
\newcommand{\muGeneralAB}{\mu(\Ei; \Xouti, \outCoefb, \outCoeftx)}

\newcommand{\meanModelTrue}{\E[Y_i \mid \Xouti, \Ei; \outCoef_0]}

\newcommand{\Lim}[1]{\lim\limits_{#1}}

\newcommand{\Xouti}{\mathbf{X}^{out}_i}

\newcommand{\Xintj}{\mathbf{X}^{int}_j}

\newcommand{\trueFA}{f_A(\Xouti; \outCoeftx)}

\newcommand{\seq}[1]{\stackrel{#1}{=}}
\newcommand{\PhiAug}{\Phi_n^{\mathrm{aug}}}
\newcommand{\PM}{PM$_{2.5}$\xspace}
\newcommand{\SOt}{SO$_{2}$\xspace}

\newcommand{\ratioJn}{\mathcal{R}}

\newcommand{\thetaMatrix}{
\begin{bmatrix}
        \hat{\outCoeftx}-\outCoeftx \\
        \hat{\outCoefb}-\outCoefb
    \end{bmatrix} }

\newcommand{\TotEff}{\texttt{TotalEffect}\xspace}
\newcommand{\EstTotEff}{\widehat{\texttt{TotalEffect}}\xspace}
\newcommand{\EstTotEffjk}{\widehat{\texttt{TotalEffect}_{jt}}\xspace}

\newcommand{\SNR}{3\xspace}

\newcommand{\cntmax}{38,372\xspace}
\newcommand{\ratemax}{55.30\xspace}
\newcommand{\cntten}{15,025\xspace}
\newcommand{\rateten}{23.37\xspace}
\newcommand{\cntninety}{37,928\xspace}
\newcommand{\rateninety}{55.00\xspace}
\newcommand{\totalCost}{43,969,292,000\xspace}

\newcommand{\Err}{\texttt{Bias}\xspace}
\newcommand{\TEErr}{\texttt{RMSE}\xspace}

\newcommand{\CostBenefit}{\texttt{BenefitCost}\xspace}
\newcommand{\EstCostBenefit}{\widehat{\texttt{BenefitCost}}\xspace}
\newcommand{\TotPop}{\texttt{TotPop}\xspace}
\newcommand{\TotHeatInp}{\texttt{TotHeatInput}\xspace}
\newcommand{\TotOp}{\texttt{TotOpTime}\xspace}
\newcommand{\TotLoad}{\texttt{Total Load}\xspace}
\newcommand{\TotSOt}{\texttt{Total \SOt Emissions}\xspace}

\newcommand{\outCoef}{\pmb{\theta}}
\newcommand{\outCoefb}{\pmb{\alpha}}
\newcommand{\outCoeftx}{\pmb{\beta}}
\newcommand{\intCoef}{\pmb{\gamma}}

\newcommand{\XJ}{\mathbf{\mathbb{X}}_{1:J}}
\newcommand{\Xj}{\Xintj}

\newcommand{\ql}{Q-Learning\xspace}
\newcommand{\al}{A-Learning\xspace}

\newcommand{\policy}{\pi}

\newcommand{\T}{\mathbf{H}}
\newcommand{\Tinst}{H_{ij}}
\newcommand{\I}[1]{\textbf{1}\{ #1 \}}

\newcommand{\etaEst}{\hat{\eta}}
\usepackage[capitalise]{cleveref}

\usepackage{anyfontsize}
\hypersetup{bookmarksdepth=subsubsection}

\usepackage{etoolbox}

\begin{document}

\journaltitle{Journal Title Here}
\DOI{DOI HERE}
\copyrightyear{2022}
\pubyear{2019}
\access{Advance Access Publication Date: Day Month Year}
\appnotes{Paper}

\firstpage{1}

\title[Policy Learning under Bipartite Network Interference]{Towards Optimal Environmental Policies:\\ Policy Learning under Arbitrary Bipartite Network Interference}


\author[1,$\ast$]{Raphael C. Kim}
\author[2]{Falco J. Bargagli-Stoffi}
\author[1]{Kevin L. Chen}
\author[1]{Rachel C. Nethery}

\authormark{Kim et al.}

\address[1]{\orgdiv{Department of Biostatistics}, \orgname{Harvard T.H. Chan School of Public Health}, \orgaddress{\street{677 Huntington Ave}, \postcode{02115}, \state{MA}, \country{USA}}}
\address[2]{\orgdiv{Department of Biostatistics}, \orgname{UCLA Fielding School of Public Health}, \orgaddress{\street{650 Charles E Young Dr S}, \postcode{90095}, \state{CA}, \country{USA}}}

\corresp[$\ast$]{Corresponding author. \href{email:raphaelkim@fas.harvard.edu}{raphaelkim@fas.harvard.edu}}

\received{5}{2}{2025}
\revised{10}{8}{2025}

\abstract{The substantial effect of air pollution on cardiovascular disease and mortality burdens is well-established. Emissions-reducing interventions on coal-fired power plants---a major source of hazardous air pollution---have proven to be an effective, but costly, strategy for reducing pollution-related health burdens. Targeting the power plants that achieve maximum health benefits while satisfying realistic cost constraints is challenging. The primary difficulty lies in quantifying the health benefits of intervening at particular plants. This is further complicated because interventions are applied on power plants, while health impacts occur in potentially distant communities, a setting known as bipartite network interference (BNI). In this paper, we introduce novel policy learning methods based on Q- and A-Learning to determine the optimal policy under arbitrary BNI. We derive asymptotic properties and demonstrate finite sample efficacy in simulations. We apply our novel methods to a comprehensive dataset of Medicare claims, power plant data, and pollution transport networks. Our goal is to determine the optimal strategy for installing power plant scrubbers to minimize ischemic heart disease (IHD) hospitalizations under various cost constraints. We find that annual IHD hospitalization rates could be reduced in a range from \rateten--\ratemax per 10,000 person-years through optimal policies under different cost constraints.
}

\maketitle


\section{Introduction}
\label{sec:intro}
Airborne pollutants, most notably fine particulate matter of an aerodynamic diameter smaller than 2.5$\mu m$ (\PM), pose a major risk to human health. Many studies have directly associated \PM exposure with cardiovascular morbidity and mortality \citep{Samet2000,Tsai2003,Koken2003,Dominici2006,Nethery2020,Wu2020,Henneman2023}, providing sufficient evidence for causality according to the U.S. Environmental Protection Agency (EPA) --- see \cite{usepa2022a}. 
Furthermore, marginalized groups, including low-income and racialized minority communities, have been shown to experience disproportionate health burdens from \PM compared to non-marginalized groups \citep{Zeger2008,Hajat2015,Kioumourtzoglou2015,Di2017,bargaglistofficre,jbaily2022air,Josey2023}. 
 Therefore, regulatory bodies, including the EPA, seek to design policies to reduce \PM exposure and, in turn, improve population health \citep{usepa2022a,usepa2022b}.

Coal-fired power plants are the largest source of sulfur dioxide emissions in the U.S., which is a major contributor to secondary \PM formation \citep{Massetti2017}. A recent study found that coal-fired power plant emissions caused approximately 460,000 deaths in the U.S. Medicare population---i.e., U.S. individuals older than 65 years of age---alone from 1999-2020 \citep{Henneman2023}. An effective intervention strategy for reducing \PM concentrations is installing flue gas desulfurization (FGD) equipment (so called ``scrubbers'') on coal-fired power plants \citep{Zigler2020}.  Scrubbers, in fact, have proven to be an effective solution, removing at least 90\% of sulfur dioxide emissions from power plants on which they are applied \citep{Srivastava2001} and, consequently, reducing the health burdens associated with exposure to power plants emissions \citep{Zigler2020}.

Given this evidence, if scrubbers were cost-free, the optimal environmental policy would be to install scrubbers at every coal-fired power plant. However, such a policy is not feasible due to the high installation and upkeep costs of scrubbers. Thus, a more nuanced intervention strategy is necessary---i.e., given a financial budget, which power plants should be `targeted' for scrubbers to achieve the greatest reduction in health burdens? To illustrate this problem more clearly, consider two hypothetical power plants: {power plant A} is a high-emitting power plant near a densely populated, urban community with high daily temperatures (and therefore elevated susceptibility to outdoor air pollution) and {power plant B} is a high-emitting power plant near a sparsely populated, rural community. Based solely on this information, a scrubber in power plant A might be prioritized over power plant B. Now suppose further that pollution originating from power plant B is known to reach ``downwind'' densely populated urban communities due to air pollution transport patterns. Also, suppose that the cost of a scrubber at power plant B is much higher than that at power plant A. In this complex but realistic scenario, it is not clear whether power plant A or power plant B should be prioritized under a given financial budget. In the even more complex reality of hundreds of power plants on which to potentially intervene, sophisticated statistical analyses are needed to design an optimal intervention strategy.

From a statistical perspective, policy learning methods for emissions-reducing policies must accommodate two key problem features that differ significantly from more conventional causal inference settings. First, intervention strategies are implemented in coal-fired power plants (``intervention units''), but health impacts are measured in surrounding communities (``outcome units''). This creates a so-called \textit{bipartite} data structure. Second, the complex process in which airborne pollutants react in the atmosphere and are transported (e.g., by the wind) means that intervening at a single power plant can potentially affect health in many distant communities, resulting in a network of connections between power plants and proximal communities. In causal inference, the presence of such connections creates interference, where one unit's treatment can spill to (possibly many) untreated units and can affect their potential outcomes. The aggregate of these two features has been described as \textit{bipartite network interference} (BNI) \citep{Zigler2021}. Although a small number of papers have proposed approaches for conducting causal inference under BNI, optimal policy learning methods under BNI have not been explored to date.


\subsection{Related Work}
The literature on causal analysis under BNI is relatively sparse. Here, we highlight primary works in BNI, non-BNI, and spatial settings. The causal BNI literature began with \cite{Zigler2021}, introducing estimators for causal effects under partial interference, adapting standard inverse probability weighted estimators to this simplified setting. The BNI literature expanded to accommodate more general causal estimands under experimental \citep{PougetAbadieNeurips2019,doudchenko2020causal} and observational settings \citep{Zigler2021, Zigler2020,chen2024environmental}, including average and heterogeneous treatment effects. More recent work considers the estimation of causal effects from longitudinal data \citep{chen2024differenceindifferences,song2024bipartite}. To the best of our knowledge, no work has considered the problem of policy learning in the context of BNI. 
The closest related works consider policy learning under interference in non-BNI settings \citep{suModelingEstimation,viviano2020policy2, viviano2024policy, zhang2024individualized}. 

From a methodological perspective, our proposed approach significantly differs from recent works in the field of policy learning and policy optimization. 
\cite{viviano2020policy2} propose experimental designs to study policy learning under interference when the interference network is unknown, but data are organized into a finite number of large clusters. \cite{viviano2024policy} proposes policy learning methods under an experimental setup for networks with an anonymous, limited degree of interference. \cite{zhang2024individualized} propose policy learning methods under clustered interference, in which the interference is restricted to a finite cluster size $m$. To overcome high variance during estimation, they propose a semiparametric heterogeneous additive model, similar to ours, that practically restricts the level of interference.  Furthermore, \cite{suModelingEstimation} proposes estimation and inference for network models under anonymous interference using a similar additive heterogeneous model to our approach without clustering assumptions. However, \cite{suModelingEstimation}'s approach assumes binary connections between units and a model setup that is specific to non-BNI settings.
Methodologically, these works are significantly different from our proposed approach and our scientific application since 
(i) the BNI setting involves separate intervention units and outcome units meaning the estimands, treatment policy, and models are different as well, (ii) finite dependence clustered network interference is not practical for our setting as interference may extend past $m$ finite units, even if the spillover might be weak, and (iii) the exposures experienced by outcome units are variable based on climate patterns, and we wish to incorporate this information into our model.

From a theoretical perspective, we employ the multiplier central limit theorem to derive inferential results as \cite{suModelingEstimation} and spatial mixing along the lines of \cite{Jenish2009} to derive regret bounds for treatment effect functions under arbitrary interference. \cite{Jenish2009} derive general asymptotic results for random fields but do not consider causal estimation. We assume a subset of the infinite-sampling regime and mixing from \cite{Jenish2009}, and employ an independent blocking technique from \cite{Bernstein1927} adapted to spatial domains to derive regret bounds. Spatial literature tends to assume finite $m-$dependence or more general dependence via spatial mixing. \cite{Pinkse07} utilizes independent blocking for spatial data but does not present the assumption set in an interpretable or verifiable manner 
and does not apply their methods to causal estimation. \cite{JENISH12} consider asymptotics for near-epoch dependent fields, a popular notion in time-series literature, and prove results on $MA(\infty)$ models, but do not consider causal estimation. \cite{Guan21} consider learning treatment regimes under spatial data for malaria control, but consider limited $m-$dependence for interference and non-bipartite settings.
Other works have considered policy learning problems under safety or budget constraints, but operate under non-BNI and non-interference settings \citep{Labersafety18, qiu2022individualized,qi2024jrssb}. Adapting such work to BNI settings is left for future work.


\subsection{Contributions and Organization}
In this paper, we develop and apply methods to learn optimal power plant scrubber installation policies to minimize hospitalization burdens for ischemic heart disease (IHD) under cost constraints. Our approach will leverage a rich dataset of Medicare claims, power plant locations and features, power plant pollution transport networks, community features, and scrubber installation costs for the eastern U.S. Building methodologically on the work of \cite{Robins2004}, we propose \ql and \al methods for policy learning under \textit{arbitrary} BNI. In contrast to much of the existing work on interference, our approach enables estimation and inference for the average treatment effects of intervening on intervention units without the need for (1) clustering assumptions or (2) causal estimands that might suffer of subjective choices of the \textit{exposure mapping} such as spillover or direct effects \citep{Hudgens2008,bargagli2024heterogeneous}. We derive the asymptotic properties of these methods and demonstrate their finite sample performance via simulation studies, finding robustness of \al to model misspecification in terms of estimation and inference, compared to \ql. We use the proposed methods to determine the power plant scrubber installation strategy that would optimally reduce IHD hospitalization outcomes within the United States, under cost constraints, finding that annual IHD hospitalizations could be reduced in a range from \cntten (\rateten per 10,000 person-years) to \cntmax (\ratemax per 10,000 person-years) based on different financial budget constraints.

The rest of the paper is organized as follows: \cref{sec:methods} introduces our Q- and A-Learning methods and their theoretical properties. \cref{sec:simulation} details an extensive semi-simulated Monte Carlo study carried out to assess performance of the methods in finite samples. In \cref{sec:rwd}, we present the details and results of our scrubber installation policy analysis. Finally, we end with discussion and limitations in \cref{sec:conclusions}. Proofs are deferred to Appendix \labelcref{sec:pfs}.
\section{Methods}\label{sec:methods}

\subsection{Setup and Objective}

\subsubsection{Setup and Notation} Suppose we have $J$ intervention units indexed by $j \in [J]$ and $n$ outcome units indexed by $i \in [n]$. 
Let $\Xouti$ denote the vector of covariates corresponding to outcome unit $i$, 
and $\Xintj$ denote the vector of covariates corresponding to intervention unit $j$. $\XJ$ will denote the covariates for all intervention units or $\XJ=\{ \Xintj \}_{j \in [J]}$. 
$\T \in \mathbb{R}^{n\times J}$ or $\{ \T_{ij} \}$ denotes the `\textit{interference map}' or bipartite adjacency matrix
. The elements of $\T$ correspond to the strength of pollution transport from a particular intervention unit $j$ to each outcome unit $i$. $\T_j^\top=(\T_{1j} \dots \T_{ij} \dots \T_{nj})^\top$ denotes a particular column of the interference map. Similarly, $\T_i=(\T_{i1} \dots \T_{ij} \dots \T_{iJ})$ denotes a particular row of the interference map---that is, $\T_{i}$ represents the strength of connections between a particular outcome unit $i$ and each of the $J$ intervention units. 
In our scientific application, $\T$ is derived from meteorological variables (primarily wind patterns), measured with high accuracy from the National Oceanic and Atmospheric Administration (NOAA) Air Resources Laboratory, aggregated over space and time \citep{Henneman2019}. 
$\mathbf{A}$ denotes the treatment status (or intervention status) vector. In particular, $\mathbf{A}=(A_1 \dots A_j \dots A_J) \in \{ 0, 1 \}^J$, where 1 corresponds to treating a particular power plant and 0 corresponds to not treating. Outcomes will depend on the exposure mapping \citep{AronowSamii2017}, which is given by $\Ei=\frac{1}{J} \sum_{j=1}^J \Tinst A_j$--- that is, the linear combination of treatments at the intervention unit level weighted by the strength of the connection between the outcome unit $i$ and intervention unit $j$ (interference map). 
Let $Y_i(\ei)$ denote the potential outcome for unit $i$ under exposure level $\ei$ and $Y_i$ denote the observed outcome for outcome unit $i$. Note that, in this work, we will consider smaller values of $Y_i$ as desirable---e.g., a smaller number of IHD hospitalizations. 
Let the policy function be given by $\pi: \eta(\Xouti, \T_j) \mapsto [0,1]$, where $\eta$ is a functional of $\Xouti, \T_j$. Additionally define $e_j$ as the propensity score for intervention unit $j$, or $e_j=\mathbb{P}(A_j=1 \mid \Xj)$.
Notationally, $\E$ will denote the mean with respect to any variables not explicitly conditioned on including $\Xouti, \Xintj, \T_i$ and $Y(\ei)$, and $\E_\pi$ will mean the expectation taken under policy $\pi$. $\E_n$ denotes the empirical mean as in the empirical process literature \cite{kosorok2008introduction}. $\lesssim$ will mean $\leq$ up to a finite constant.


\subsubsection{Objective}  The objective is to determine the optimal treatment policy $\policy^* \in \ActSet$ for some policy class $\ActSet$, which minimizes $\Epi[Y_i(\ei)]$, the mean outcome under policy $\policy$--- e.g. the treatment policy that yields the smallest number of IHD hospitalizations. The causal estimands depend on a realization of $\T$ and $n, J$. As done in \cite{Ogburn24} and preceding works, the causal estimands and estimators condition on $\T$ and $n, J$.
\begin{equation}\label{eq:theoreticalOptObj}
     \policy^*=\arg \min_{\policy \in \ActSet} \Epi[Y_i(\ei)]
\end{equation}

\subsection{Nonparametric Identification}
We state the identification assumptions required to identify our estimand \cref{eq:theoreticalOptObj} from data:

\paragraph{Identification Assumptions}
\begin{enumerate}[{\bfseries ({I}1)}] 
    \item \textit{ Consistency of outcome units}: \label{ass:consistency} $Y_i=Y_i(\ei)$.
    \item \textit{ Positivity}: \label{ass:positivity} 
    $P(\ei \mid \Xouti, \XJ, \T_i) > 0$.
    \item \textit{ Unconfoundedness}: \label{ass:noConfounding}
    $Y_i(\ei) \perp \ei \mid \Xouti, \XJ, \T_i$.
    \item \textit{ Intervention covariates are independent of potential outcomes given the exposure mapping $\ei$ and outcome covariates}:
    \label{ass:interferenceMapRichness} 
    $Y_i(\ei) \perp \XJ \mid \Xouti, \T_i, \ei $.
\end{enumerate}

Assumptions \refAssump{I}{\labelcref{ass:consistency}} -- \refAssump{I}{\labelcref{ass:noConfounding}} 
are adaptations of the standard consistency, positivity, and `no unmeasured confounding' assumptions in causal inference to our BNI setting. 
\refAssump{I}{\labelcref{ass:interferenceMapRichness}} 
encodes domain knowledge from air pollution policy: outcome unit potential outcomes are determined by aggregate exposure levels $\Ei$ rather than individual treatment values of intervention units $A_j$.

\begin{remark} Assumption 
\refAssump{I}{\labelcref{ass:consistency}} 
is made in terms of our exposure level $\ei$, not at the intervention level $A_j$. Under this formulation, it is possible for two treatments $\mathbf{A}^{(1)}=(A_{1}^{(1)} \dots A_{J}^{(1)}) \neq \mathbf{A}^{(2)}=(A_{1}^{(2)} \dots A_{J}^{(2)}) $ to yield the same exposure level $\ei$. This does not threaten the validity of our consistency assumption and results developed below, since we reason at the exposure level. 
\end{remark}

\begin{remark}
We comment further on \refAssump{I}{\labelcref{ass:interferenceMapRichness}}: 
Reiterating our justification above, such an assumption is reasonable in our setting since health outcomes are impacted by total pollution levels, not individual power plant treatment statuses. Further, the predominant mechanism through which power plant covariates affect our health outcomes is air pollution levels. To account for additional confounders such as socioeconomic factors, we additionally adjust for $\Xouti$ in the model. Thus, when learning the distribution of $Y_i(\ei)$ given $\Ei$ and $\Xouti$, it is not necessary to condition on $\XJ$. 
\end{remark}


\begin{proposition}[Identification] \label{thm:identification}
    Under \refAssump{I}{\labelcref{ass:consistency}} -- \refAssump{I}{\labelcref{ass:interferenceMapRichness}}, we have 
    \begin{equation}
        \E[Y_i(\ei)]= \E\big[\E[Y_i \mid \Xouti, \T_i, \Ei=\ei]\big]
    \end{equation}
\end{proposition}

This means that we can instead consider the following objective function
\begin{equation}\label{eq:optObj}
     \policy^* = \arg \min_{\policy \in \ActSet} \E_n\big[\E[Y_i \mid \Xouti, \T_i, \Ei=\ei]\big]
\end{equation}
Note that, in the real world, we may have cost constraints---e.g., a total budget for installation cost of scrubbers. Let $C$ denote the total cost constraint and for $j \in [J],$ let $c_j=\mbox{Cost}(A_j=1)-\mbox{Cost}(A_j=0)$ denote the cost difference between treating and non treating the intervention unit $j$. 
 To enable identification of cost-constrained optimal policies, we require these costs $c_j$ to be known: 
     $ c_j \geq 0$. 
\begin{enumerate}[\bfseries ({C}1)]
\label{ass:costAssumptions}
    \item \label{ass:CostKnown} \textit{ Known treatment costs:}
    For $j \in [J]$, $c_j$ is known and
    $ c_j \geq 0$. 
\end{enumerate}

\begin{proposition}[Cost-constrained Optimal Policy]\label{thm:optObjCost}
    Assume \refAssump{I}{\labelcref{ass:consistency}} -- \refAssump{I}{\labelcref{ass:interferenceMapRichness}} and \refAssump{C}{\labelcref{ass:CostKnown}}.
    Then, the optimal cost constrained optimization problem is given by:
    \vspace{-1em}
    \begin{align*} \label{eq:optObjCost}
    \policy^* = \arg \min_{\policy \in \ActSet} \quad & \E_n[\E[Y_i \mid \Xouti, \T_i, \Ei=\ei]] \numberthis \quad \mbox{ s.t. } \sum_{j=1}^J c_j \pi_j \leq C \mbox{ }
    \end{align*}
\end{proposition}

\subsection{Optimal Policy Learning using Parametric, Linear Exposure Models}

In this section, we propose \ql and \al methods for solving \cref{eq:optObjCost}. Motivated by domain knowledge that outcome unit potential outcomes are determined by an aggregate exposure level, we assume the outcome model is linear in $\Ei$. 
This assumption not only simplifies the policy optimization, but enhances the interpretability of the optimal solution. For illustration purposes, we present methods based on parametric forms. Please note that such results can be generalized. Namely, the methods below can be extended to semi/nonparametric models under appropriate smoothness conditions and general outcome model forms. See Appendix \labelcref{sec:generalizations} for details. 
At a high level, we are able to derive inferential results by employing the multiplier central limit theorem as \cite{suModelingEstimation}, and regret bounds by spatial mixing assumptions of the outcome units as \cite{Jenish2009}. We now describe our setup in more detail, while the formal proofs can be found in the appendix. Here, we assume the following outcome model.

\paragraph{Modeling Assumption}
\begin{enumerate}[\bfseries ({O}1)]
    \item \label{ass:outcomeModel} \textit{ Outcome model is linear in the exposure mapping}:
    \begin{align*}
    Y_i(\ei; \outCoef_0) &= \muTrue+\epsilon_i\\
        &= f_0(\Xouti, \outCoefb_0)+ \ei \cdot f_A(\Xouti, \outCoeftx_0) +\epsilon_i \numberthis \label{eq:meanModel}
    \end{align*}
\end{enumerate}
\noindent for $\outCoef_0 = (\outCoefb_0, \outCoeftx_0) \in \Theta \subseteq \R^{|\outCoef_0|}$, which parameterize $f_0(\Xouti; \outCoefb_0)$ and $f_A(\Xouti; \outCoeftx_0)$ respectively, and mean zero independent random variable $\epsilon_i$. Above, $f_0$ captures the main effect of the covariates and $f_A$ captures the treatment effect. By \cref{eq:meanModel}, we can express the objective function as follows:
\begin{align*}
    \frac{1}{n} \sum_{i=1}^n \meanModelTrue & = \frac{1}{n} \sum_{i=1}^n f_0(\Xouti; \outCoefb_0) + \frac{1}{n} \sum_{i=1}^n \Ei f_A(\Xouti; \outCoeftx_0) \\
    & = \frac{1}{n} \sum_{i=1}^n f_0(\Xouti; \outCoefb_0) + \frac{1}{n} \sum_{j=1}^J  A_j \frac{1}{J} \sum_{i=1}^n \Tinst f_A(\Xouti; \outCoeftx_0) 
\end{align*}
Let
\begin{equation}
    \TotEff_j= \frac{1}{J} \sum_{i=1}^n  \Tinst f_A(\Xouti; \outCoeftx_0), \quad j\in [J] \label{eq:TEj}
\end{equation}
Hence the effect of treating an intervention unit $j$ is fully captured by $\TotEff_j$, precluding the need for defining a direct or spillover effect or making clustered interference assumptions. The quantity in \cref{eq:TEj} will serve as a key object of interest throughout our analyses below.

\subsubsection{Optimal Policy under a Linear Exposure Model.} Since smaller values of $Y$ are preferred (e.g. less hospitalizations), the optimal regime $\policy^*$ in the absence of cost constraints would be given by
\begin{equation}\label{eq:optimalPolicy}
    \policy_j^*(\TotEff_j)=\I{\TotEff_j < 0} 
\end{equation}

\noindent We are now set up to state our \ql and \al approach to obtaining statistical inference procedures. At a high-level, \ql and \al posit regression models for the mean outcome and determine the optimal policy by using a plug-in estimate for the decision rule. \ql relies on a correctly-specified outcome regression model, while \al is a variant of \ql that fits a modified estimating equation to endow double robustness \citep{Schulte_2014}.
In our case of \cref{eq:meanModel}, \ql requires $\mu=(f_0, f_A)$ to be correctly specified to obtain the optimal policy and inference. As we will see in our approach below, \al requires $f_A$ to be correctly specified but only requires correct specification of only one of baseline model $f_0$ or propensity score model $e$ to obtain valid inference and the optimal policy.

\subsubsection{\ql}\label{sec:QLearning}
Parametric \ql proceeds by proposing a model $\E[Y_i \mid \Xouti, \Ei; \outCoef_0]$, and solving \cref{eq:optObjCost} using the plug-in estimate $\hat{\outCoef}$.

\paragraph{Estimation}
To estimate $\outCoef_0$, we perform least squares regression, i.e. we solve the following estimating equations:
\begin{equation}\label{eq:qLearningEE}
     \frac{1}{n} \sum_{i=1}^n \phi(Y_i; \outCoef)=0 
\end{equation}
\noindent where $\phi(Y_i; \outCoef)=(Y_i - \muGeneral)d_i(\outCoef)$ and $d_i(\outCoef) = \frac{\partial \muGeneral}{\partial \outCoef}$.
\paragraph{Inference}
%
Under suitable regularity conditions, we find for covariance matrix $\Sigma$ defined in Appendix \labelcref{pf:qLearning}, \label{thm:qLearning}
        $$ \sqrt{n}(\hat{\outCoef}-\outCoef_0) \xrightarrow{d} N(0, \Sigma) $$

The proof follows in a similar manner to \cref{thm:aLearning} below invoking the multiplier central limit theorem \cite{kosorok2008introduction}. We defer the details to Appendix \labelcref{pf:qLearning}.

\subsubsection{\al} \label{sec:ALearning}

\ql requires the full model $\mu=(f_0, f_A)$ to be correctly specified in order to obtain the optimal policy and inference. In this section, we propose a doubly robust estimator for $\outCoeftx$. Below, we are explicit about the dependence on $(\outCoefb,\outCoeftx)$ since they are estimated using different estimating equations.
Let the propensity score for intervention unit $j$, $e_j$, be parameterized by $\intCoef_0 \in \Gamma \subseteq \R^{|\intCoef_0|}$, or $e_j=\mathbb{P}(A_j=1 \mid \Xj; \intCoef_0)$.

\paragraph{Estimation}
We propose to estimate $\outCoeftx_0$ by solving for the roots of the following estimating equation:
\begin{align*}  \label{eq:alearning_beta}
\Phi_n(\outCoeftx, \outCoefb, \intCoef) &= \frac{1}{n}\sum_{i=1}^n \lambda(\Xouti, \T_i; \outCoeftx) (Y_i- \muGeneralAB) (\Ei - \EiHat) \numberthis
\end{align*}
where $\EiHat=\frac{1}{J} \sum_{j=1}^J e_j(\Xintj, \hat{\gamma}) \Tinst$ and $\lambda(\Xouti, \T_i;\outCoeftx)$, following the notation of Eq. 31 in \citep{Schulte_2014}, is an arbitrary function that is of dimension $|\outCoeftx|$.

We solve this estimating equation jointly with the following estimating equations for $\bm{\alpha_0}, \bm{\gamma_0}$. For $\alpha$, we solve
\begin{equation}
    \frac{1}{n}\sum_{i=1}^n \frac{\partial}{\partial \outCoefb} f_0(\Xouti, \outCoefb) (Y_i-\muGeneralAB) =0 \label{eq:alearning_alpha}
\end{equation}
For $\gamma$, we solve
\begin{equation}
    \frac{1}{J}\sum_{j=1}^J \frac{\partial}{\partial \intCoef} e(\Xj, \intCoef) (A_j - e(\Xj, \intCoef) ) =0  \label{eq:alearning_gamma}
\end{equation}

\begin{remark}
In \cref{eq:alearning_beta}, the choice of $\lambda$ is flexible. For example, one can take this to be $\lambda(\Xouti, \T_i; \outCoeftx)=c_i \frac{\partial}{\partial \outCoeftx} f_A(\Xouti, \outCoeftx)$ for $c_i \in \mathbb{R}$. This is similar to \citep{tsiatisSemiparametricClass03, Robins2004, Schulte_2014}, in causal inference literature on defining a `\textit{class}' of estimating equations that may satisfy consistent estimation, asymptotic normality, and double-robustness. During our simulations and analysis, we take $c_i=\frac{1}{J}\sum_{j=1}^J \Tinst$.
\end{remark}

\begin{remark}
    In our specification of the propensity score, we may augment $\mathbf{X}_j^{int}$ to incorporate information from $\Xouti, \T_i$ by including weighted summaries of outcome unit covariates, weighted by the interference level $\T_i$ (e.g., $\mathbf{X}^{out}_j=\frac{1}{n}\sum_{i=1}^n \Tinst \mathbf{X}^{out}_i$). Such an approach is natural for incorporating outcome level covariates in the intervention model, up- or down-weighting the influence of outcome unit characteristics based on how much a given outcome unit is affected by the intervention unit in question.
\end{remark}

This formulation yields an approach to estimating $\outCoeftx$ that is doubly-robust, as shown by the lemma below; if either $e$ or $f_0$ is correctly specified, we have consistent parameter estimation for $\outCoeftx$ \citep{bangRobins2005}. We clarify that the following approach still requires $f_A$ to be correctly specified, or the true treatment effects will not be recovered.

\begin{lemma}[Double Robustness]\label{thm:unbiased}
    If $e$ or $f_0$ is correctly specified, then under the identification conditions 
    \refAssump{I}{1}-\refAssump{I}{4},
    $$ \E[\Phi_n(\hat{\outCoeftx}, \hat{\outCoefb}, \hat{\intCoef})]=0$$
\end{lemma}

\paragraph{Inference}
    \begin{theorem}[Asymptotic Normality for \al]\label{thm:aLearning}
        Under suitable regularity conditions and the assumption that $\frac{J}{n} \rightarrow \ratioJn$ a fixed constant, as $J,n \rightarrow \infty$, we have for covariance matrix $\Omega$ defined in \cref{pf:aLearning}, 
        \begin{align*} \sqrt{n}\begin{bmatrix}
           \hat{\outCoefb}-\outCoefb_{0} \\
            \hat{\outCoeftx}-\outCoeftx_0 
         \end{bmatrix} \xrightarrow{d} N(0, \Omega) 
        \end{align*}
    \end{theorem}
    
\subsection{Estimating the Optimal Policy} \label{sec:howToOptPolicy}

With the treatment effect estimates in hand via $\hat{\outCoeftx}$ from either \ql or \al, we now describe how to estimate the optimal policy. In \cref{sec:newUnits}, we describe how to utilize our learned treatment effect parameters to perform policy learning on future, possibly independent samples.

\subsubsection{Unconstrained Optimal Policy.} From  \cref{eq:optimalPolicy}, we use the plug-in $\hat{\outCoeftx}$ to find:
\begin{equation}
    \hat{\pi}_j^*(\TotEff(\hat{\outCoeftx})_j) = \I{\TotEff(\hat{\outCoeftx})_j < 0}
\end{equation}

\subsubsection{Cost-Constrained Optimal Policy.} The constrained optimal policy is given by solving \cref{eq:optObjCost} with plug-in $\hat{\outCoeftx}$.
If we assume that the costs and cost constraints are positive (i.e. $c_j, C > 0$), we can obtain an efficient $O(J)$ computable solution to \cref{eq:optObjCost} via the fractional knapsack solution, taking $c_j$ as weights and $-Y_i$ as profits 
(see Ch. 15 \cite{10.5555/1614191}). To translate this to our setup, we define the benefit-cost ratio of installing a scrubber at a power plant $j$ relative to the cost, as 
\begin{equation}
    \CostBenefit_j=\frac{\TotEff_j}{c_j}
\end{equation}
  Such ``relative benefits'' play a key role in determining cost-constrained optimal intervention strategies, as $\TotEff_j$ does for the unconstrained counterpart. We will estimate this ratio using our model output to explain and contextualize our findings in \cref{sec:optCostConstrainedPolicy}.

\subsection{Regret Bound Analysis}
In the context of policy learning, it is key to study the regret of the estimated optimal policy with respect to the true optimal policy. These regret bounds guarantee that our policy learned acts optimally as our sample size grows, and characterizes the rate at which optimality is achieved.

Since the optimal policy relies on estimating $\TotEff_j$, we first show consistency and estimation rates for this quantity. With this in hand, regret bounds follow. For brevity, let $\hat{\eta}_j=\TotEff(\hat{\outCoeftx})_j$ and $\eta_j=\TotEff(\outCoeftx)_j$. Throughout, we rely on a spatial setup and mixing assumptions that are standard in spatial statistics literature, similar to that in \cite{Jenish2009}. 
    The full setup and assumptions are described in \cref{pf:TE_Consistency}. To provide some more context into our setup, \refAssump{S}{\labelcref{ass:samplingRegime}} formalizes how we observe spatial units, saying that asymptotically, the number of units we observe grows in space, while \refAssump{S}{\labelcref{ass:mixing}} says that units that are distance $r$ apart have dependency that decays sufficiently fast. 
    
    These assumptions are well-suited for our scientific setting, where interference from air pollution exposure exhibits spatial spillovers that are not adequately captured by traditional finite-clustering assumptions. With these in hand, we find the following.

\begin{theorem}[Consistency of Treatment Effect Estimation]\label{thm:TE_Consistency}
Assume spatial mixing (\textbf{S\labelcref{ass:samplingRegime}})-(\textbf{S\labelcref{ass:mixing}}), and suitable regularity conditions. Then we can estimate the treatment effect function for intervention $j$ at $\sqrt{n}$ rates, or 
    $$\E \frac{1}{n} |\etaEst_j -\eta_j| \lesssim \sqrt{\frac{1}{n}} $$
\end{theorem}

With this in hand, we may conclude that our regret is asymptotically 0. I.e.
\begin{theorem}[Convergence to the Optimal Policy]\label{thm:RegretBd}
Assume spatial mixing (\textbf{S\labelcref{ass:samplingRegime}})-(\textbf{S\labelcref{ass:mixing}}), and suitable regularity conditions. Denote our estimated policy from utilizing $\hat{\eta}_j$ by $\hat{\pi}^*$. Then our policy is asymptotically optimal (with regret approaching 0), or 
    $$|\E_{\hat{\pi}^*}[Y_i(\ei)]-\E_{\pi^*}[Y_i(\ei)]| \xrightarrow{p} 0$$
\end{theorem}

\section{Simulation Study}\label{sec:simulation}

In this section, we conduct an empirical Monte Carlo study to validate the methods in \cref{sec:methods}. The covariates and interference matrix are the same as from our data application across all simulations. Next to these ``empirical variables'', we randomly generate treatment assignments and outcomes according to a fixed Signal-to-Noise Ratio (SNR). The \Err of $\outCoeftx$ (\cref{eq:betaError}), the \TEErr of $\EstTotEff$ (\cref{eq:teError}), and the coverage probabilities are reported across all simulations. The equations for the metrics used in the simulation are found in \cref{sec:simDetails}.
Throughout simulations, we assume the following form for the true outcome and propensity score models:

\paragraph{Outcome Model.}
We assume that $f_0, f_A$ are quadratic in $\Xouti$, where $\alpha_{\mathrm{intercept}} \in \mathbb{R}$ and $\outCoefb_1, \outCoefb_2 \in \mathbb{R}^{\mathrm{dim}(\Xouti)}$. Similarly, $\beta_{\mathrm{intercept}} \in \mathbb{R}$ and $\outCoeftx_1, \outCoeftx_2 \in \mathbb{R}^{\mathrm{dim}(\Xouti)}$. 
\begin{equation} \label{eq:outBaseModelSim}
    f_0(\Xouti, \alpha)=\alpha_{\mathrm{intercept}} + \Xouti  \outCoefb_{1}  + (\Xouti)^2 \outCoefb_2 
\end{equation}
\begin{equation}\label{eq:outTxModelSim}
    f_A(\Xouti, \beta)= \beta_{\mathrm{intercept}} +  \Xouti \outCoeftx_{1}  + (\Xouti)^2 \outCoeftx_2 
\end{equation}

\paragraph{Propensity Score Model.}
 We model $e_j$ using a logistic regression model where $\gamma_{\mathrm{intercept}} \in \mathbb{R}$ and $\intCoef_1, \intCoef_2 \in \mathbb{R}^{\mathrm{dim}(\Xouti)}$.
\begin{align*} \label{eq:intPropModelSim}
    \log \left(\frac{e_j}{1-e_j}\right) &= \gamma_{\mathrm{intercept}} + \Xintj \intCoef_{1} + (\Xintj)^2 \intCoef_2 
    \numberthis
\end{align*}

In the simulations, we consider scenarios with fully correct model specification and with $f_0$ and/or $e$ misspecified to demonstrate the double robustness property of A-learning. Misspecification of each component is induced by fitting models imposing linear relationships with covariates in a given component instead of quadratic. Specifically, outcome model misspecification is carried out by setting $\hat{\outCoefb}_2=0$. The misspecification of the propensity score model is introduced by setting $\hat{\intCoef}_2=0$. Throughout, to construct an empirically grounded simulation framework that accurately reflects the real-world dynamics of our data application, $\intCoef$ is chosen so that simulated average of propensities $\frac{1}{J} \sum_{j=1}^J e_j$ is within 0.01 of the observed empirical average of treatments in the dataset (see final row of \cref{tab:pp_covs}). Further, $\outCoef=(\outCoefb,\outCoeftx)$ is chosen so that the simulated average $\frac{1}{n}\sum_{i=1}^n Y_i$, based on the expected exposure level $\bar{A}=\frac{1}{J}\sum_{j=1}^J \Tinst e_j$ from Step \labelcref{sim:stepGenTx}, is within 0.01 of the empirical average in the dataset (see penultimate row of \cref{tab:zip_covs}). 

Our simulation setup is as follows:
\begin{enumerate}
    \item \textbf{ Preprocess data}: Standardize the covariates of $\Xouti, \Xintj$.
    \item \label{sim:stepGenTx} \textbf{ Generate intervention unit treatments}: $A_j \sim \mathrm{Bernoulli}(e_j)$, where $e_j$ follows \cref{eq:intPropModelSim}.  
    \item \textbf{ Compute exposure mapping}: $\Ei=\frac{1}{J} \sum_{j \in [J]}\Tinst A_j$
    \item \textbf{ Generate outcomes}: $Y_i(\Ei)$ such that the SNR is \SNR, i.e. 
        $$Y_i=f_0(\Xouti, \outCoefb) + \Ei \cdot f_A(\Xouti,\outCoeftx) + \epsilon_i$$ 
        where 
         $\mathbb{V}(\epsilon_i)=\mathbb{V}\left(\frac{\E[Y_i \mid \Xouti, \T_i, \Ei; \outCoef]}{\SNR}\right)$ 
        and $f_0$, $f_A$ follow \cref{eq:outBaseModelSim,eq:outTxModelSim} respectively. 
    \item \textbf{ Estimate parameters}: $\outCoef=(\outCoefb, \outCoeftx)$ using each of the following:
    \begin{itemize}
    \item Q-learning with correctly specified outcome model $\mu=(f_0, f_A)$
    \item Q-learning with misspecified $f_0$, correctly specified $f_A$
    \item A-learning with correctly specified outcome $\mu=(f_0, f_A)$ and propensity score $e$
    \item A-learning with correctly specified outcome $\mu=(f_0, f_A)$ and misspecified $e$
    \item A-learning with $f_0$ misspecified but $f_A$ and $e$ correctly specified
    \item A-learning with both $f_0$ and $e$ misspecified but $f_A$ correctly specified
    \end{itemize}
    where the Q-learning models are estimated using least squares regression and the A-learning models solve \cref{eq:alearning_alpha,eq:alearning_gamma,eq:alearning_beta} jointly. \label{sim:stepEstTheta} 
    \item \textbf{ Iterate}: Repeat steps (\labelcref{sim:stepGenTx}-\labelcref{sim:stepEstTheta}) 1000 times.
    \begin{enumerate}[(i)]
        \item Compute $\Err$, $\TEErr$,
        and the average empirical 95\% confidence interval coverage probabilities for $\hat{\outCoeftx}$ (across all the dimensions of $\mbox{dim}(\Xouti)$). 
    \end{enumerate}
\end{enumerate}

The results from 1000 simulations are shown in \cref{tab:simResults_func}. The correctly specified \ql model achieves the lowest estimation error, consistent with common knowledge that a correctly-specified outcome model tends to outperform other methods. \al with correctly specified models is competitive with \ql correctly specified. Under misspecification of the propensity score model ($e$), \al gives nearly equivalent performance to \al under a fully correct model specification (first and second rows of \al). As we introduce misspecification of the outcome model in \ql and \al, the error metrics increase. 

Based on these simulations, under equivalent misspecification forms, the methodology seems less sensitive to propensity score model misspecification than outcome model misspecification. This phenomenon was also observed in simulations on doubly robust causal effect estimation under BNI in \cite{chen2024environmental}.
Regardless, under equivalent baseline model misspecification (second row of \ql and third row of \al), \al results in lower estimation error than \ql. 
\al yields good empirical coverage for significance level $\alpha=0.05$, even with misspecified models.  In contrast, under equivalent outcome model misspecification, \ql has extremely low coverage, affecting inference more severely than \al. 

In general, we find a lower estimation error and a higher coverage under various levels of misspecification in \al compared to misspecified \ql, validating the theoretical results shown above in \cref{sec:QLearning} and \cref{sec:ALearning}.
\section{Optimal Power Plant Scrubber Installation Policy}\label{sec:rwd}

We applied the optimal policy methods developed above to real data to determine the allocation of power plant scrubbers that minimizes the hospitalization rates for IHD among Medicare beneficiaries in the United States given a financial budget. The data, analysis of policy learning, and the results are described below. 
Additional analyses on non-linear outcome model fitting and equivalent count analysis are found in the Supplement.

\subsection{Data Description}
\label{sec:data}
Our data consist of three parts: (i) outcome level data with Medicare beneficiary information,  (ii) intervention level data with power plant characteristics and scrubber cost information, and (iii) the interference map or characterization of air pollution transport effects between outcome units and intervention units via a climate model known as \textit{HYSPLIT Average Dispersion} (HyADS) \citep{Henneman2019}. We detail each component below. 

\subsubsection{Outcome Level Data: Medicare Beneficiary Data}

The outcome data are derived from Medicare inpatient claims data for all 27,312,190 fee-for-service beneficiaries residing in the U.S. in 2005. The health outcome of interest is the ZIP code-level ischemic heart disease (IHD) inpatient hospitalization rate per 10,000 person-years in 2005 among these Medicare beneficiaries. For each ZIP code, U.S. Census socioeconomic and demographic features (from the 2000 decennial census), meteorological \citep{Kalnay1996}, and smoking rate \citep{DwyerLindgren2014} covariate data were obtained. 
Further, due to the relatively small number of power plants in the Western U.S., we remove ZIP codes in the U.S. states of Wyoming, Colorado, Idaho, Utah, Arizona, New Mexico, Nevada, California, Oregon, Montana, and Washington. After initial cleaning, we have $n=26,674$ ZIP codes. The ZIP code covariates used in our analysis, and corresponding descriptive statistics, are given in \cref{tab:zip_covs}.

The ZIP code-level IHD hospitalization rate per 10,000 person-years is used as the outcome in our statistical models, in order to account for different amounts of person-time of exposure across ZIP codes. Our primary results shown below are based on minimizing rates of hospitalization. The results stated in terms of total counts of hospitalizations are detailed in \cref{sec:analysisCounts}. 

\subsubsection{Intervention Level Data: Power Plant Data}

Data on scrubber status in the year 2005 and plant characteristics for $J=409$ coal–fired power plants concentrated in the Eastern US, which serve as our intervention units of interest, were obtained from the U.S. Environmental Protection Agency Air Markets Program Database. The plant covariates used in our analyses, along with descriptive statistics, are given in \cref{tab:pp_covs}. Scrubber installation cost information for power plants that had a scrubber installed by 2021 was obtained from the Energy Information Administration (EIA) website. 

\paragraph{Interference Map: HyADS Climate Model}

To characterize the interference structure or the bipartite adjacency matrix, we utilize the HyADS model. HyADS is a pollution transport model that yields a unit-less metric quantifying the amount of emissions from an individual power plant that were transported (e.g., by wind) to a particular ZIP code \citep{Henneman2019}. HyADS is based on highly accurate meteorological measures from the HYSPLIT model by the National Oceanic and Atmospheric Administration (NOAA) Air Resources Laboratory. 
These values, which were calculated for all power plant pairs and ZIP codes in the data, form the elements of $H$.

\subsection{Data Preprocessing}

We perform log transformations of several covariates (\TotHeatInp, \TotPop, and \TotOp) in order to reduce skew in their distribution and standardization of covariates from non-categorical outcome and intervention units. Prior to final analysis for \al, to reduce instability we perform trimming of intervention units with low estimated probability of treatment (i.e., very small estimated propensity scores) \citep{StrumerTrimSim2010}. Specifically, we trim below the bottom 5th percentile of estimated propensity scores (approximately $0.018$). After trimming intervention units, we end up with $J=388$ power plants. 

\subsection{Cost Model Development}

In order to find the optimal allocation of scrubbers under cost-constraints, we need to quantify the cost of installing a scrubber at each power plant. Because only a subset of plants has scrubbers installed (and therefore has observed installation cost information), we build a model to impute scrubber installation costs for power plants without scrubbers based on covariates available from our intervention dataset. Since the cost data from the EIA include information from various stacks within a given power plant facility, we define the cost ($c_j$ in \cref{eq:optObjCost}) to be the average cost of installing a scrubber at a given power plant. 
 
The EIA data report scrubber installation costs for 135 of the power plants represented in our data. These 135 samples serves as our training set. We follow a standard machine learning approach to build our cost model. First, we split our training data ($J=135$) into a (sub) training and validation set using an 80/20 split, resulting in ($J=108$) and ($J=27$) respectively. Then, we test three models---namely, random forest, support vector regression, and linear regression---on the validation set. The model with the best performance is selected. Finally, the selected model is then trained on our full training data, and used to generate scrubber installation cost predictions on power plant data with missing cost information ($J=225$). Performance metrics on the validation set, variables' importance metrics from our final random forest model, and cost predictions are shown in \cref{sec:CostModel}. Under our cost model, the total cost for installing a scrubber at all of the 388 power plants is approximately \$\totalCost. 

We clarify that the $J=135$ power plants considered in our training set above represents the number of treated power plants by 2021. In contrast, the number of treated power plants in 2005 is 73 (see \cref{tab:pp_covs}). When building our cost model, we utilize all power plants treated with scrubbers by 2021 to build the strongest cost model we can.

\subsection{Policy Analysis}

We apply the approaches developed above to (1) quantify the effects of FGD scrubber installation at each power plant on IHD hospitalizations in 2005 and (2) identify optimal scrubber installation policies under cost constraints. We perform our policy analysis under a hypothetical ``clean slate'' scenario in which none of the power plants have scrubbers installed. In other words, we identify the scrubber installation schemes that would be optimal to pursue (under various cost constraints) if we ``re-set'' to a world in which no scrubbers had yet been installed. By providing recommended policies and corresponding gains across a range of cost constraints, we offer insights into the potential benefits and how they compare to the factual policy implementation. In Appendix \labelcref{sec:addOnAnalysis}, we provide details on an alternative analysis in which the optimal policies instead build upon the factual 2005 scrubber installation landscape.
 Throughout, we use \al to protect against model misspecification. For shorthand, let $\EstTotEff_j=\TotEff_j(\hat{\outCoeftx})$ and $\EstCostBenefit_j=\frac{\EstTotEff_j}{c_j}$

\subsubsection{\TotEff Rates Analysis}

\begin{figure}[ht!]
\centering
\resizebox{\textwidth}{!}{
\begin{tabular}{cc}
     \includegraphics[width=.7\textwidth]{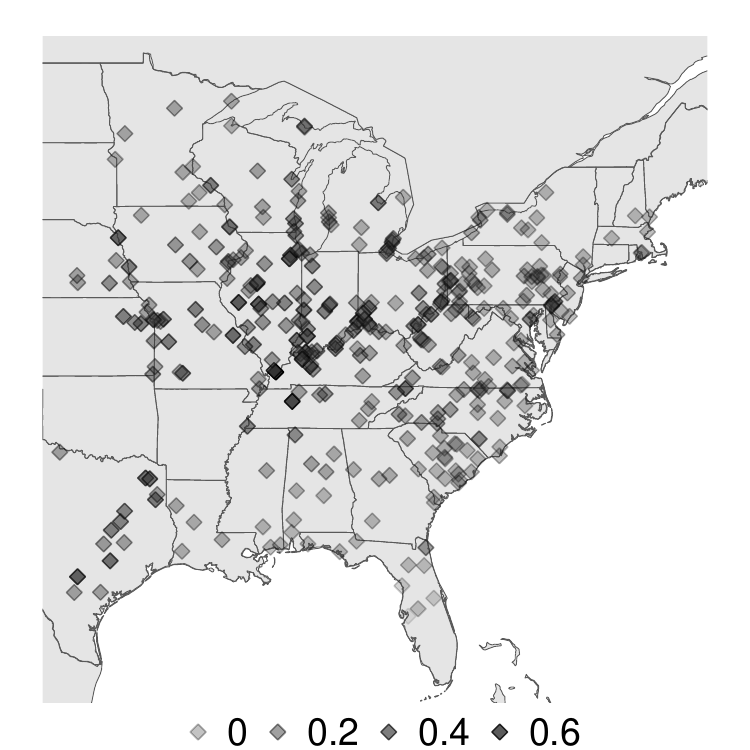} 
     &    \includegraphics[width=.75\textwidth]{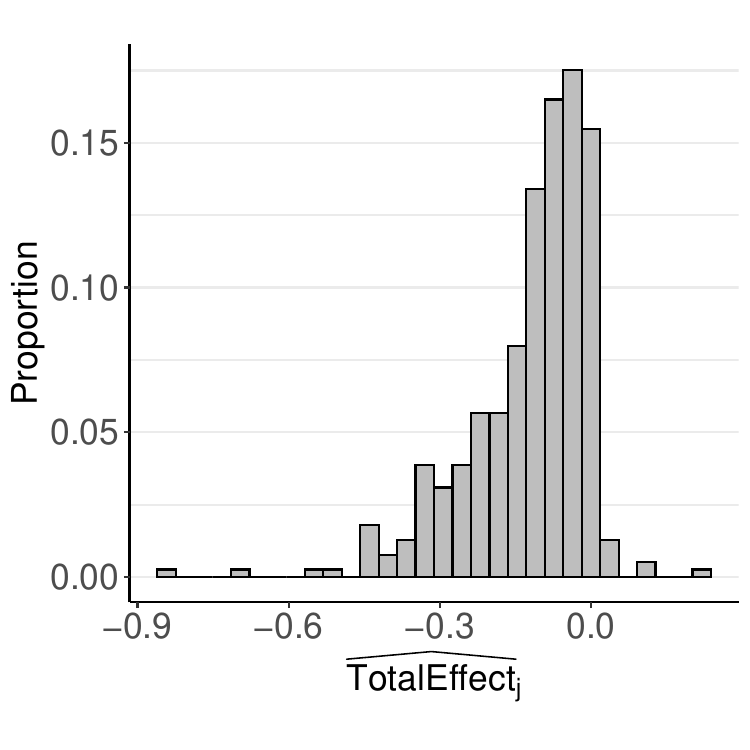} 
     \\
     (a)  & (b) 
\end{tabular}
}
\caption{\textmd{(a) displays the plot of $\EstTotEff_j$ on a U.S. map. (b) displays the histogram of $\EstTotEff_j$.} \label{fig:TEs}}
\end{figure}

For interpretability, we build an outcome model and a propensity score model akin to that specified in \cref{sec:simulation}, respectively, but linear in covariates instead of quadratic. We additionally augment $\Xintj$ with weighted summaries of outcome level covariates given by $\mathbf{X}^{out}_j=\frac{1}{n}\sum_{i=1}^n \Tinst \mathbf{X}^{out}_i$ since outcome covariates and the interference map may impact treatment decisions. Sensitivity analyses with more complex models are reported in Appendix \labelcref{sec:sens_analysis_generalize}. 
Upon fitting the \al model, we find that $96\%$ of power plants have estimated protective effects if treated, indicating that scrubber installation would reduce IHD hospitalizations. Graphically \cref{fig:TEs} shows a plot of $\EstTotEff_j$ on the U.S. map and in a histogram. 
Installing scrubbers at most power plants tend to have small effects, save for a handful of influential power plants in the Midwest. For example, consider the power plant that has the greatest estimated reduction in IHD hospitalization rates if a scrubber in installed, found in Tennessee. If we install a scrubber at this power plant, we would find a .69 (448) 
reduction in IHD hospitalization rate per 10,000 person-years (counts) in 2005 among Medicare beneficiaries. 

When determining which power plants to implement interventions on, policymakers may wish to incorporate statistical uncertainty into the decision-making process. Thus, to enhance the practical utility and policy-relevance of the proposed methodologies, we perform an exploratory analysis testing $H_0: \TotEff_j \geq 0, H_a: \TotEff_j < 0$ based on \cref{thm:aLearning}. The results are shown in \cref{fig:pvals_toteffj}. Approximately $70\%$ of p-values are statistically significant, yielding evidence of statistically significant protective population effects when installing scrubbers at the majority of the power plants for significance level $\alpha=0.05$.

\subsubsection{Optimal scrubber policies under cost constraints}  \label{sec:optCostConstrainedPolicy}

Because, to our knowledge, policymaking entities have not specified exact budgets for scrubber installations, here we consider various total cost budgets defined as percents (10\%, 20\%,...,90\%) of the estimated cost of installing scrubbers at all power plants used in our analysis (\$\totalCost, as reported above). Under each cost budget, the optimal treatment policy, i.e., the set of plants that are selected for scrubber installation, and the subsequent average reduction in IHD hospitalizations per 10,000 person years are shown in \cref{fig:panelBudgetRate}. Equivalent analysis for counts are shown in Appendix \labelcref{sec:analysisCounts}. 
Under a budget constituting 10\% of the cost of universal scrubber installations at all US power plants, annual IHD hospitalizations can be reduced by \cntten (\rateten per 10,000 person-years) via the optimal scrubber installation policy. The plants selected for scrubbers under this budget are largely concentrated in the Ohio River Valley and Midwest. Under a budget constituting 90\% of universal installation costs, \cntninety (\rateninety per 10,000 person-years) IHD hospitalizations can be avoided annually. As shown in \cref{fig:panelBudgetRate}, the health benefits of additional scrubber installations seems to level off for budgets above approximately 60\% of universal installation cost, and the biggest reductions occur as the budget increases from 10\% -- 30\% of universal cost.

\begin{figure}[ht!]
\centering
\resizebox{\textwidth}{!}{
\begin{tabular}{cc}
     \includegraphics[width=.7\textwidth]{images_v1/senssubsetStates.pdf} 

     &  \includegraphics[width=.7\textwidth]{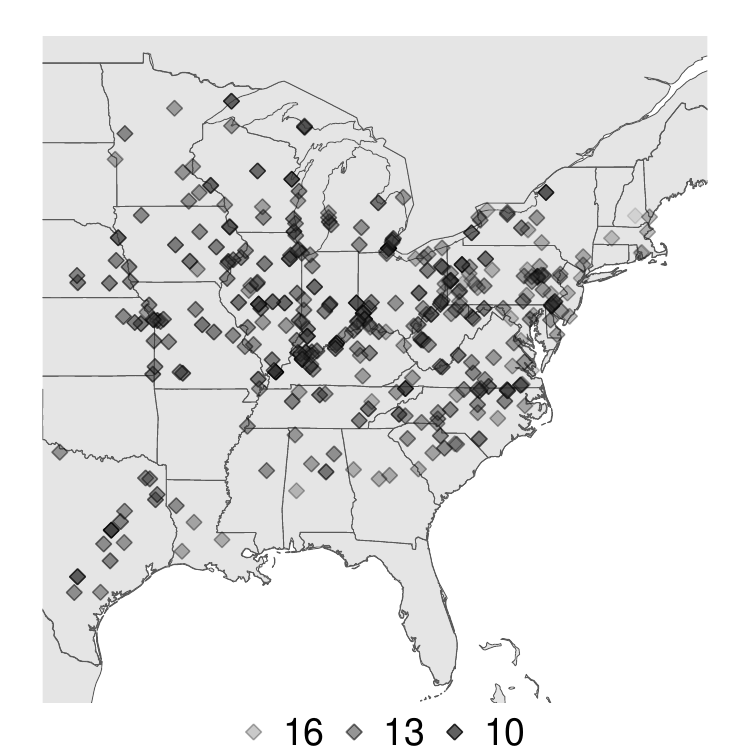} \\

     (a)  & (b)
\end{tabular}
}
\caption{\textmd{(a) displays the plot of $-\EstTotEff_j$ on a U.S. map, colored by intensity of  $\EstTotEff_j$. (b) displays an analagous plot for $-\log(-\EstCostBenefit_j)$.} 
\label{fig:rawVsProfit}}
\end{figure}

To investigate this, we map the $\EstTotEff_j$ alongside the $\EstCostBenefit_j$ in \cref{fig:rawVsProfit}. The scale of $\EstCostBenefit_j$ is much smaller, so we negate and log transform it for visualization purposes. This enables us to compare how `dark' different power plants are in \cref{fig:rawVsProfit}, to give more insights into the factors influencing selection for treatment in the optimal policies. As we can see, the power plants around the Midwest that have the greatest reductions in $\EstTotEff_j$ generally have competitive cost-benefit ratios. Interestingly, we can observe that several other power plants in the surrounding regions are also prioritized in the cost-constrained intervention strategies because, although their $\EstTotEff_j$ is relatively small, they also have cheaper scrubber installation costs, leading to a more competitive $\EstCostBenefit_j$.

\begin{figure}[ht!]
\centering
 \resizebox{\textwidth}{!}{
\begin{tabular}{ccc}
    \includegraphics[width=0.33\textwidth]{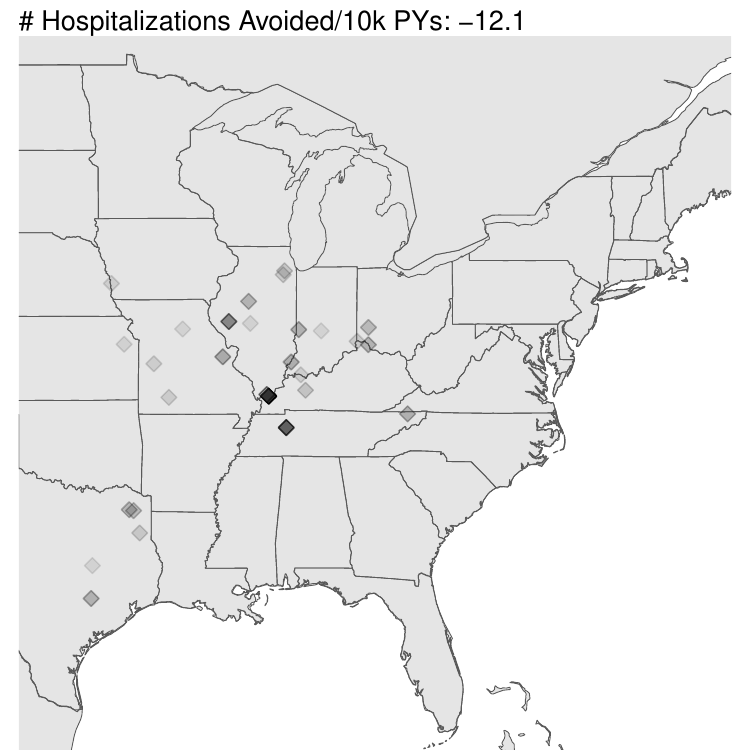}
     &
     \includegraphics[width=0.33\textwidth]{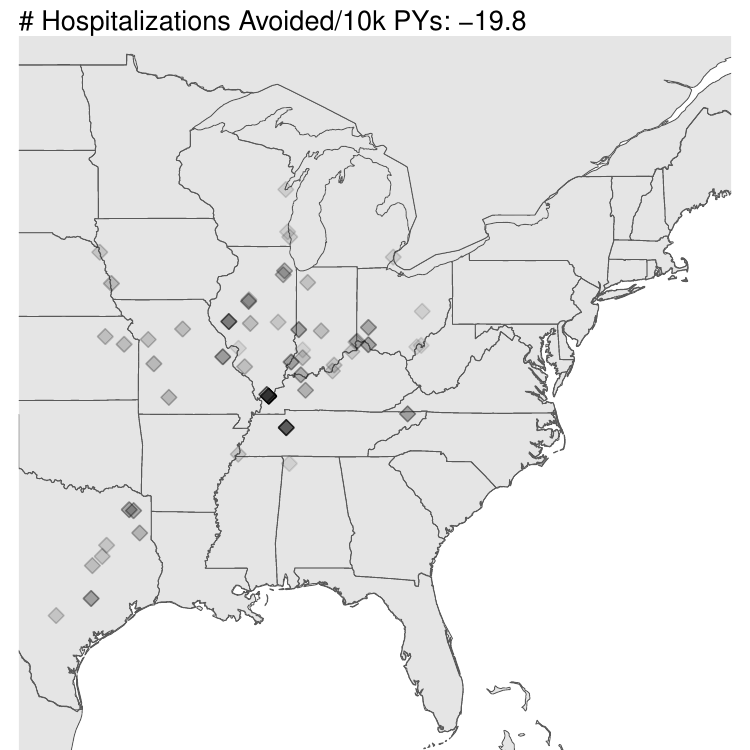} &

     \includegraphics[width=0.33\textwidth]{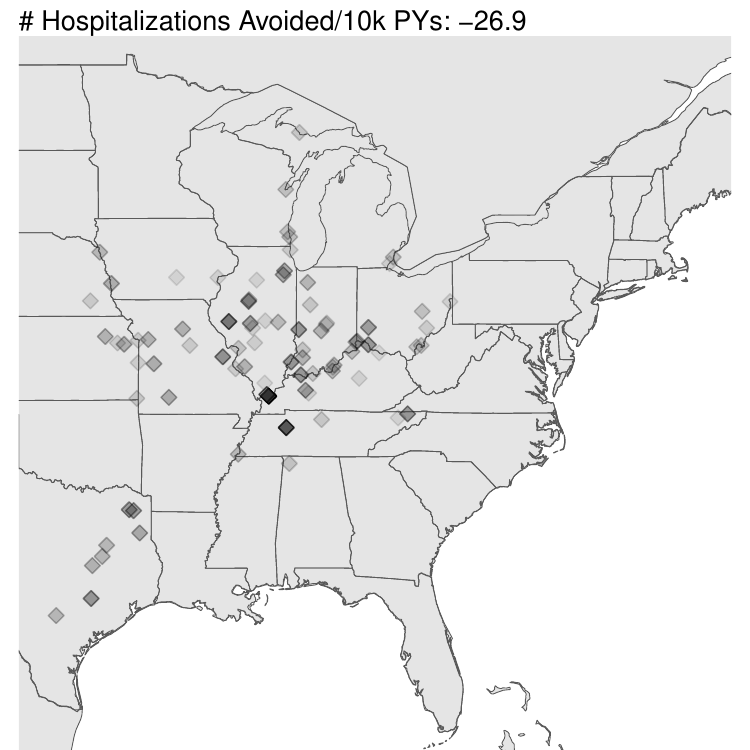}
     
     \\
     (a) 10\% Total Cost & (b) 20\% Total Cost  & (c) 30\% Total Cost 
\end{tabular}}

 \resizebox{\textwidth}{!}{
\begin{tabular}{ccc}
    \includegraphics[width=0.33\textwidth]{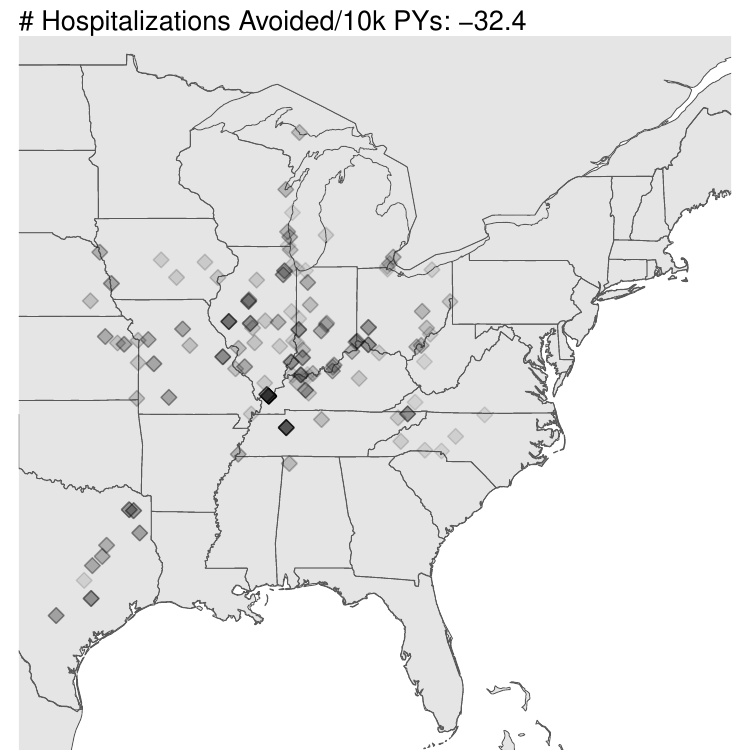}
     &
     \includegraphics[width=0.33\textwidth]{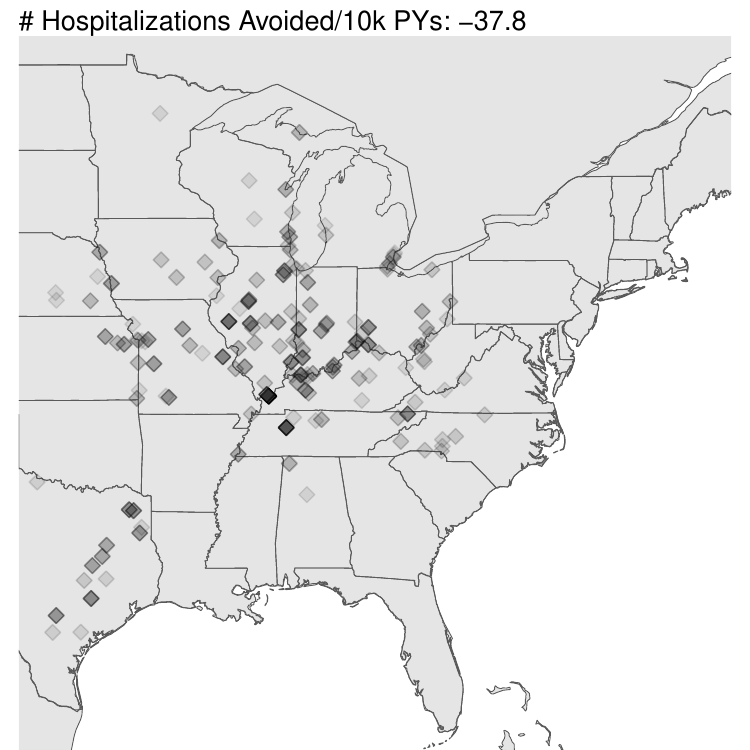} &

     \includegraphics[width=0.33\textwidth]{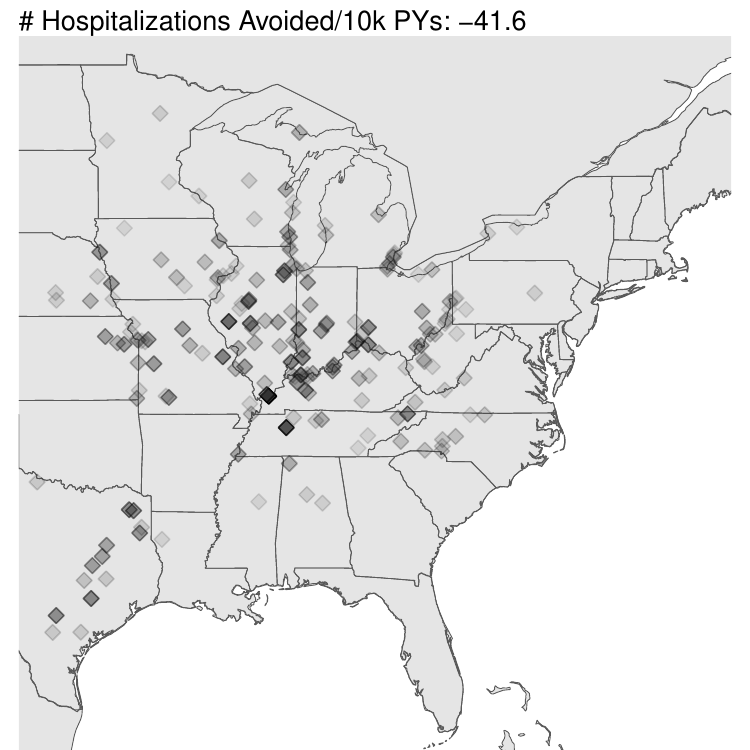}
     
     \\
     (a) 40\% Total Cost & (b) 50\% Total Cost  & (c) 60\% Total Cost 
\end{tabular}}

 \resizebox{\textwidth}{!}{
\begin{tabular}{ccc}
    \includegraphics[width=0.33\textwidth]{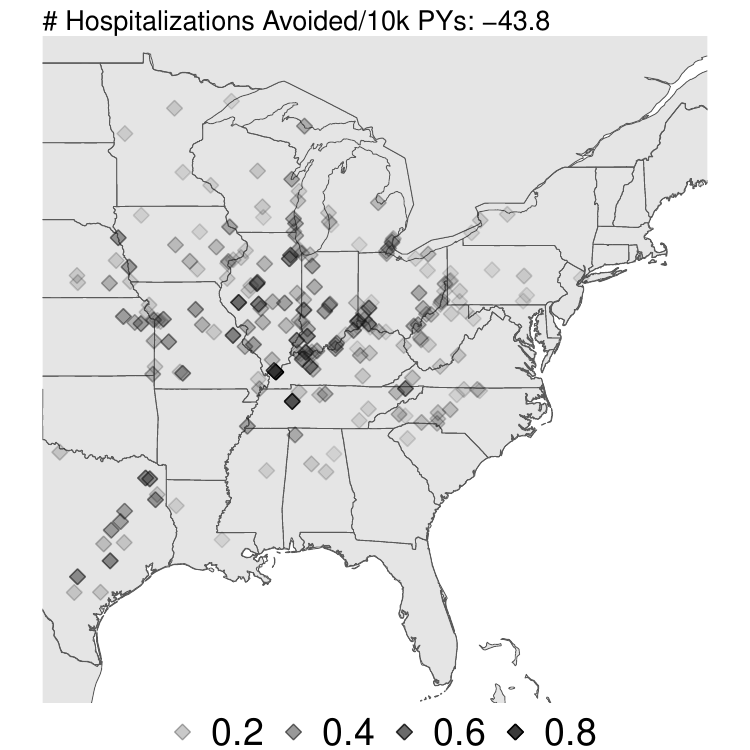}
     &
     \includegraphics[width=0.33\textwidth]{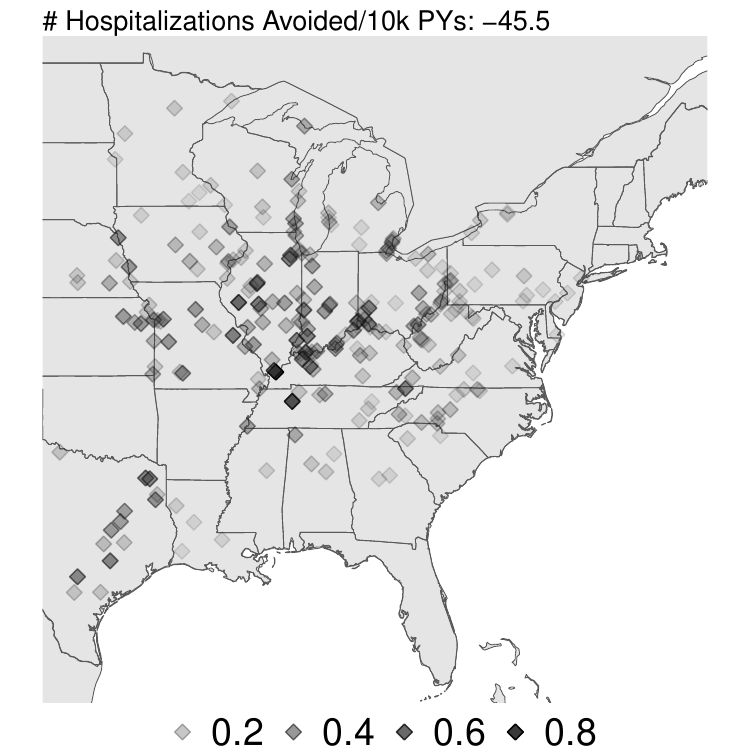} &

     \includegraphics[width=0.33\textwidth]{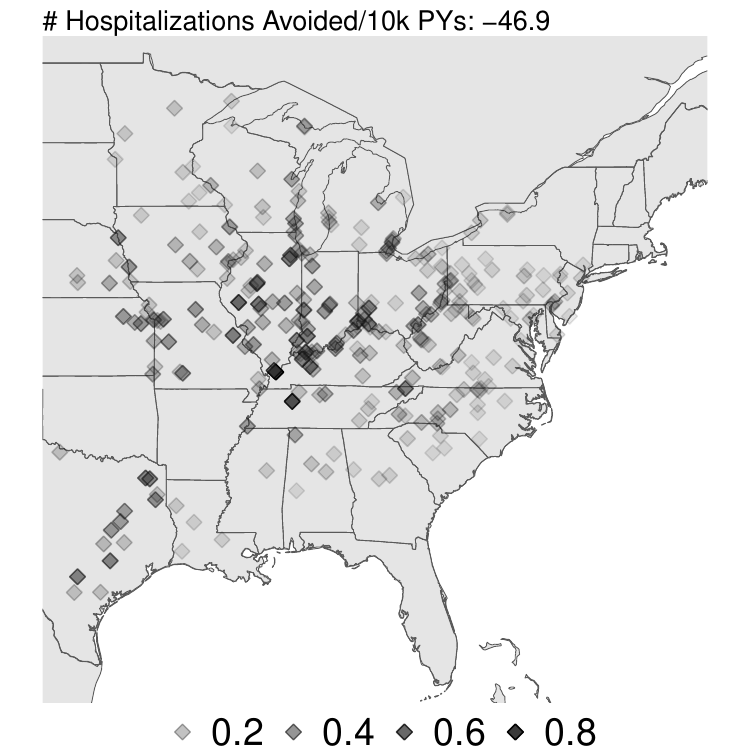}
     
     \\
     (a) 70\% Total Cost & (b) 80\% Total Cost  & (c) 90\% Total Cost 
\end{tabular}}
\caption{\textmd{Grid with the reduction of IHD Hospitalizations/10,000 Person Years, varying the spending from 10\%-90\% of budget, optimizing based on $\EstTotEff$ rather than $\EstCostBenefit$.}
\label{fig:panelBudgetRate_rawOpt}}
\end{figure}

In Section~\ref{ss:opt_vs_fact} of the Appendix, we directly compare the factual scrubber policy in 2005 with the optimal policy of the same cost, finding that the optimal policy would have treated a larger number of power plants and more than tripled the reductions in IHD hospitalization rates relative to the factual policy. We end with a comparison of the cost-constrained policies from \cref{fig:panelBudgetRate} to a more intuitive policy that policymakers may carry out under realistic cost constraints. Recall from \cref{sec:howToOptPolicy} that the optimal policy to \cref{eq:optObjCost} is obtained from the fractional knapsack solution. Informally, this means that one must: (a) estimate $\CostBenefit_j$, and (b) treat power plants $j$ with the lowest $\EstCostBenefit_j$ (as lower hospitalizations are better) so long as the cost constraint is satisfied. 
Intuitively, a policymaker may do the same as above, but instead treat power plants with the lowest $\EstTotEff_j$ (i.e. the greatest reduction in IHD hospitalizations) rather than the lowest $\EstCostBenefit_j$; \cref{fig:panelBudgetRate_rawOpt} shows the effect of enacting such a policy. Comparing \cref{fig:panelBudgetRate} to \cref{fig:panelBudgetRate_rawOpt}, we find \textit{uniformly lower} reductions (i.e. greater health benefits) from solving \cref{eq:optObjCost} compared to a more intuitive solution.

In \cref{sec:sensScrubbersPrior2005}, we check the sensitivity of our findings to potential violations to the temporal ordering between scrubber installations and measured outcomes in 2005. There are only 4 of 388 (1\%) of power plants that may violate this, and results are nearly identical whether we include or exclude these plants.


\section{Discussion}\label{sec:conclusions}

Understanding how to optimally assign an intervention is critically important to optimize policy effects while keeping the costs fixed. In this paper, we proposed a new methodology to produce optimal policies in the context of bipartite network interference.  Using our proposed methodology, we carry out the first statistical policy learning analyses for air pollution policy, finding up to \cntmax (\ratemax per 10,000 person-years) IHD hospitalizations can be avoided annually among Medicare fee-for-service beneficiaries. This analysis supports previous findings that reducing pollutants from large-scale emissions reduces public health burden \citep{Henneman2023}. The analysis of $\EstCostBenefit$ sheds light onto how nuanced this analysis can be under realistic cost constraints. Different budgets and policy choices can yield variable public health impacts. We consider a clean-slate analysis in which none of the power plants have scrubbers installed in order to provide insights into the potential benefits that could be experienced. In the appendix, we design optimal policies that build upon the factual 2005 scrubber landscape rather than a clean-slate. The contrast between the two approaches are shown in \cref{tab:addOnComparisonRate}, providing insight into how efficient the factual policy may be. If we were interested in decision-making on future power plants, we can leverage our rule from \ql or \al. The details of this procedure are outlined in Appendix \labelcref{sec:newUnits}.

From a methodological standpoint, we proposed the first BNI policy learning methodology based on Q- and \al under arbitrary BNI and validated our proposed methods in an extensive simulation study that closely mimics our real world setting. Our approach does not require clustering assumptions or defining direct and spillover estimands as commonly done in interference literature; such assumptions are untenable in this problem due to air pollution transport. The proposed methodologies hold general applicability for large environmental policy studies.

There are several limitations to note. We conduct our primary analysis assuming a linear exposure mapping and that HyADS is a primary effect modifier for power plant interventions. Such an assumption enables computationally feasible modeling and aligns with domain knowledge, but it is untestable in practice. Methodological limitations of this work are assuming costs $c_1 \dots c_J$ are known. 
Future work should address the limitations above. Further, given recent evidence of disproportionate burden to marginalized subgroups in air pollution regulatory policy \citep{Zeger2008,Hajat2015,Kioumourtzoglou2015,Di2017,jbaily2022air,Josey2023, chen2024environmental}, extending this work to ensure `fair' policy learning \citep{viviano2024fair} stands as an important future direction.

\section{Competing interests}
No competing interest is declared.

\section{Author contributions statement}
R.C.K., F.J.B.S., K.L.C., and R.C.N., contributed to writing and conception of the manuscript. R.C.K. and K.L.C. contributed to performing the analysis.

\section{Acknowledgments}
This work was funded by the Health Effects Institute award 4998-RFA22-2/24 and NIH grants T32ES714242, K01ES032458. No conflicts of interest to declare.

\section{Supplementary Material}
\label{sec6}
Information about where the data are obtained from, simulation code, and analysis code are available at 
\href{https://github.com/NSAPH-Projects/emissions-ihd-bni_optimal_tx}{\texttt{https://github.com/NSAPH-Projects/emissions-ihd-bni\_optimal\_tx}}. Supplementary material is available online at \url{http://biostatistics.oxfordjournals.org}. 

\bibliographystyle{abbrvnat}
\bibliography{refs}

\begin{thebibliography}{56}
\providecommand{\natexlab}[1]{#1}
\providecommand{\url}[1]{\texttt{#1}}
\expandafter\ifx\csname urlstyle\endcsname\relax
  \providecommand{\doi}[1]{doi: #1}\else
  \providecommand{\doi}{doi: \begingroup \urlstyle{rm}\Url}\fi

\bibitem[Aronow and Samii(2017)]{AronowSamii2017}
P.~M. Aronow and C.~Samii.
\newblock {Estimating average causal effects under general interference, with
  application to a social network experiment}.
\newblock \emph{The Annals of Applied Statistics}, 11\penalty0 (4):\penalty0
  1912 -- 1947, 2017.
\newblock \doi{10.1214/16-AOAS1005}.
\newblock URL \url{https://doi.org/10.1214/16-AOAS1005}.

\bibitem[Bang and Robins(2005)]{bangRobins2005}
H.~Bang and J.~M. Robins.
\newblock {{D}oubly robust estimation in missing data and causal inference
  models}.
\newblock \emph{Biometrics}, 61\penalty0 (4):\penalty0 962--973, Dec 2005.

\bibitem[Bargagli-Stoffi et~al.(2020)Bargagli-Stoffi, Cadei, Lee, and
  Dominici]{bargaglistofficre}
F.~J. Bargagli-Stoffi, R.~Cadei, K.~Lee, and F.~Dominici.
\newblock Causal rule ensemble: Interpretable discovery and inference of
  heterogeneous treatment effects, 2020.

\bibitem[Bargagli-Stoffi et~al.(2025)Bargagli-Stoffi, Tort{\'u}, and
  Forastiere]{bargagli2024heterogeneous}
F.~J. Bargagli-Stoffi, C.~Tort{\'u}, and L.~Forastiere.
\newblock {Heterogeneous treatment and spillover effects under clustered
  network interference}.
\newblock \emph{The Annals of Applied Statistics}, 19\penalty0 (1):\penalty0 28
  -- 55, 2025.
\newblock \doi{10.1214/24-AOAS1913}.
\newblock URL \url{https://doi.org/10.1214/24-AOAS1913}.

\bibitem[Belloni et~al.(2015)Belloni, Chernozhukov, Chetverikov, and
  Kato]{belloni2015}
A.~Belloni, V.~Chernozhukov, D.~Chetverikov, and K.~Kato.
\newblock Some new asymptotic theory for least squares series: Pointwise and
  uniform results, 2015.

\bibitem[Bernstein(1927)]{Bernstein1927}
S.~Bernstein.
\newblock Sur l'extension du théorème limite du calcul des probabilités aux
  sommes de quantités dépendantes.
\newblock \emph{Mathematische Annalen}, 97:\penalty0 1--59, 1927.
\newblock URL \url{http://eudml.org/doc/182666}.

\bibitem[Bradley(2005)]{BradleySurvey05}
R.~C. Bradley.
\newblock {Basic Properties of Strong Mixing Conditions. A Survey and Some Open
  Questions}.
\newblock \emph{Probability Surveys}, 2\penalty0 (none):\penalty0 107 -- 144,
  2005.
\newblock \doi{10.1214/154957805100000104}.
\newblock URL \url{https://doi.org/10.1214/154957805100000104}.

\bibitem[Chen et~al.(2024{\natexlab{a}})Chen, Bargagli-Stoffi, Kim, Henneman,
  and Nethery]{chen2024differenceindifferences}
K.~L. Chen, F.~J. Bargagli-Stoffi, R.~C. Kim, L.~R.~F. Henneman, and R.~C.
  Nethery.
\newblock Difference-in-differences under bipartite network interference: A
  framework for quasi-experimental assessment of the effects of environmental
  policies on health.
\newblock \emph{arXiv preprint arXiv:2404.13442}, 2024{\natexlab{a}}.

\bibitem[Chen et~al.(2024{\natexlab{b}})Chen, Bargagli-Stoffi, Kim, and
  Nethery]{chen2024environmental}
K.~L. Chen, F.~J. Bargagli-Stoffi, R.~C. Kim, and R.~C. Nethery.
\newblock Environmental justice implications of power plant emissions control
  policies: Heterogeneous causal effect estimation under bipartite network
  interference, 2024{\natexlab{b}}.

\bibitem[Cormen et~al.(2009)Cormen, Leiserson, Rivest, and
  Stein]{10.5555/1614191}
T.~H. Cormen, C.~E. Leiserson, R.~L. Rivest, and C.~Stein.
\newblock \emph{Introduction to Algorithms, Third Edition}.
\newblock The MIT Press, 3rd edition, 2009.
\newblock ISBN 0262033844.

\bibitem[Di et~al.(2017)Di, Wang, Zanobetti, Wang, Koutrakis, Choirat,
  Dominici, and Schwartz]{Di2017}
Q.~Di, Y.~Wang, A.~Zanobetti, Y.~Wang, P.~Koutrakis, C.~Choirat, F.~Dominici,
  and J.~D. Schwartz.
\newblock Air pollution and mortality in the {M}edicare population.
\newblock \emph{New England Journal of Medicine}, 376\penalty0 (26):\penalty0
  2513--2522, June 2017.
\newblock \doi{10.1056/nejmoa1702747}.
\newblock URL \url{https://doi.org/10.1056/nejmoa1702747}.

\bibitem[Dominici et~al.(2006)Dominici, Peng, Bell, Pham, McDermott, Zeger, and
  Samet]{Dominici2006}
F.~Dominici, R.~D. Peng, M.~L. Bell, L.~Pham, A.~McDermott, S.~L. Zeger, and
  J.~M. Samet.
\newblock Fine particulate air pollution and hospital admission for
  cardiovascular and respiratory diseases.
\newblock \emph{{JAMA}}, 295\penalty0 (10):\penalty0 1127, Mar. 2006.
\newblock \doi{10.1001/jama.295.10.1127}.
\newblock URL \url{https://doi.org/10.1001/jama.295.10.1127}.

\bibitem[Doudchenko et~al.(2020)Doudchenko, Zhang, Drynkin, Airoldi, Mirrokni,
  and Pouget-Abadie]{doudchenko2020causal}
N.~Doudchenko, M.~Zhang, E.~Drynkin, E.~Airoldi, V.~Mirrokni, and
  J.~Pouget-Abadie.
\newblock Causal inference with bipartite designs, 2020.

\bibitem[Dwyer-Lindgren et~al.(2014)Dwyer-Lindgren, Mokdad, Srebotnjak,
  Flaxman, Hansen, and Murray]{DwyerLindgren2014}
L.~Dwyer-Lindgren, A.~H. Mokdad, T.~Srebotnjak, A.~D. Flaxman, G.~M. Hansen,
  and C.~J. Murray.
\newblock Cigarette smoking prevalence in {US} counties: 1996-2012.
\newblock \emph{Population Health Metrics}, 12\penalty0 (1), Mar. 2014.
\newblock \doi{10.1186/1478-7954-12-5}.
\newblock URL \url{https://doi.org/10.1186/1478-7954-12-5}.

\bibitem[Giné and Nickl(2015)]{gineNickl}
E.~Giné and R.~Nickl.
\newblock \emph{Mathematical Foundations of Infinite-Dimensional Statistical
  Models}.
\newblock Cambridge University Press, 2015.
\newblock \doi{10.1017/cbo9781107337862}.

\bibitem[Guan et~al.(2021)Guan, Reich, and Laber]{Guan21}
Q.~Guan, B.~J. Reich, and E.~B. Laber.
\newblock A spatiotemporal recommendation engine for malaria control.
\newblock \emph{Biostatistics}, 23\penalty0 (3):\penalty0 1023--1038, 04 2021.
\newblock ISSN 1465-4644.
\newblock \doi{10.1093/biostatistics/kxab010}.
\newblock URL \url{https://doi.org/10.1093/biostatistics/kxab010}.

\bibitem[Hajat et~al.(2015)Hajat, Hsia, and O'Neill]{Hajat2015}
A.~Hajat, C.~Hsia, and M.~S. O'Neill.
\newblock Socioeconomic disparities and air pollution exposure: a global
  review.
\newblock \emph{Current Environmental Health Reports}, 2\penalty0 (4):\penalty0
  440--450, Sept. 2015.
\newblock \doi{10.1007/s40572-015-0069-5}.
\newblock URL \url{https://doi.org/10.1007/s40572-015-0069-5}.

\bibitem[Henneman et~al.(2023)Henneman, Choirat, Dedoussi, Dominici, Roberts,
  and Zigler]{Henneman2023}
L.~Henneman, C.~Choirat, I.~Dedoussi, F.~Dominici, J.~Roberts, and C.~Zigler.
\newblock Mortality risk from united states coal electricity generation.
\newblock \emph{Science}, 382\penalty0 (6673):\penalty0 941–946, 2023.
\newblock ISSN 1095-9203.
\newblock \doi{10.1126/science.adf4915}.
\newblock URL \url{http://dx.doi.org/10.1126/science.adf4915}.

\bibitem[Henneman et~al.(2019)Henneman, Choirat, Ivey, Cummiskey, and
  Zigler]{Henneman2019}
L.~R. Henneman, C.~Choirat, C.~Ivey, K.~Cummiskey, and C.~M. Zigler.
\newblock Characterizing population exposure to coal emissions sources in the
  united states using the {HyADS} model.
\newblock \emph{Atmospheric Environment}, 203:\penalty0 271--280, Apr. 2019.
\newblock \doi{10.1016/j.atmosenv.2019.01.043}.
\newblock URL \url{https://doi.org/10.1016/j.atmosenv.2019.01.043}.

\bibitem[Hudgens and Halloran(2008)]{Hudgens2008}
M.~G. Hudgens and M.~E. Halloran.
\newblock Toward causal inference with interference.
\newblock \emph{Journal of the American Statistical Association}, 103\penalty0
  (482):\penalty0 832--842, June 2008.
\newblock \doi{10.1198/016214508000000292}.
\newblock URL \url{https://doi.org/10.1198/016214508000000292}.

\bibitem[Jbaily et~al.(2022)Jbaily, Zhou, Liu, Lee, Kamareddine, Verguet, and
  Dominici]{jbaily2022air}
A.~Jbaily, X.~Zhou, J.~Liu, T.-H. Lee, L.~Kamareddine, S.~Verguet, and
  F.~Dominici.
\newblock Air pollution exposure disparities across {US} population and income
  groups.
\newblock \emph{Nature}, 601\penalty0 (7892):\penalty0 228--233, 2022.

\bibitem[Jenish and Prucha(2009)]{Jenish2009}
N.~Jenish and I.~R. Prucha.
\newblock Central limit theorems and uniform laws of large numbers for arrays
  of random fields.
\newblock \emph{J. Econom.}, 150\penalty0 (1):\penalty0 86--98, May 2009.

\bibitem[Jenish and Prucha(2012)]{JENISH12}
N.~Jenish and I.~R. Prucha.
\newblock On spatial processes and asymptotic inference under near-epoch
  dependence.
\newblock \emph{Journal of Econometrics}, 170\penalty0 (1):\penalty0 178--190,
  2012.
\newblock ISSN 0304-4076.
\newblock \doi{https://doi.org/10.1016/j.jeconom.2012.05.022}.
\newblock URL
  \url{https://www.sciencedirect.com/science/article/pii/S0304407612001340}.

\bibitem[Josey et~al.(2023)Josey, Delaney, Wu, Nethery, DeSouza, Braun, and
  Dominici]{Josey2023}
K.~P. Josey, S.~W. Delaney, X.~Wu, R.~C. Nethery, P.~DeSouza, D.~Braun, and
  F.~Dominici.
\newblock Air pollution and mortality at the intersection of race and social
  class.
\newblock \emph{New England Journal of Medicine}, Mar. 2023.
\newblock \doi{10.1056/nejmsa2300523}.
\newblock URL \url{https://doi.org/10.1056/nejmsa2300523}.

\bibitem[Kalnay et~al.(1996)Kalnay, Kanamitsu, Kistler, Collins, Deaven,
  Gandin, Iredell, Saha, White, Woollen, Zhu, Leetmaa, Reynolds, Chelliah,
  Ebisuzaki, Higgins, Janowiak, Mo, Ropelewski, Wang, Jenne, and
  Joseph]{Kalnay1996}
E.~Kalnay, M.~Kanamitsu, R.~Kistler, W.~Collins, D.~Deaven, L.~Gandin,
  M.~Iredell, S.~Saha, G.~White, J.~Woollen, Y.~Zhu, A.~Leetmaa, R.~Reynolds,
  M.~Chelliah, W.~Ebisuzaki, W.~Higgins, J.~Janowiak, K.~C. Mo, C.~Ropelewski,
  J.~Wang, R.~Jenne, and D.~Joseph.
\newblock The {NCEP}/{NCAR} 40-year reanalysis project.
\newblock \emph{Bulletin of the American Meteorological Society}, 77\penalty0
  (3):\penalty0 437--471, Mar. 1996.
\newblock \doi{10.1175/1520-0477(1996)077<0437:tnyrp>2.0.co;2}.
\newblock URL
  \url{https://doi.org/10.1175/1520-0477(1996)077<0437:tnyrp>2.0.co;2}.

\bibitem[Kioumourtzoglou et~al.(2015)Kioumourtzoglou, Schwartz, James,
  Dominici, and Zanobetti]{Kioumourtzoglou2015}
M.-A. Kioumourtzoglou, J.~Schwartz, P.~James, F.~Dominici, and A.~Zanobetti.
\newblock {PM}2.5 and mortality in 207 {US} cities.
\newblock \emph{Epidemiology}, page~1, Nov. 2015.
\newblock \doi{10.1097/ede.0000000000000422}.
\newblock URL \url{https://doi.org/10.1097/ede.0000000000000422}.

\bibitem[Koken et~al.(2003)Koken, Piver, Ye, Elixhauser, Olsen, and
  Portier]{Koken2003}
P.~J.~M. Koken, W.~T. Piver, F.~Ye, A.~Elixhauser, L.~M. Olsen, and C.~J.
  Portier.
\newblock Temperature, air pollution, and hospitalization for cardiovascular
  diseases among elderly people in {D}enver.
\newblock \emph{Environmental Health Perspectives}, 111\penalty0 (10):\penalty0
  1312--1317, Aug. 2003.
\newblock \doi{10.1289/ehp.5957}.
\newblock URL \url{https://doi.org/10.1289/ehp.5957}.

\bibitem[Kosorok(2008)]{kosorok2008introduction}
M.~R. Kosorok.
\newblock Introduction to semiparametric inference.
\newblock \emph{Introduction to Empirical Processes and Semiparametric
  Inference}, pages 319--321, 2008.

\bibitem[Kuznetsov and Mohri(2017)]{KuznetsovMohri17}
V.~Kuznetsov and M.~Mohri.
\newblock Generalization bounds for non-stationary mixing processes.
\newblock \emph{Mach. Learn.}, 106\penalty0 (1):\penalty0 93–117, Jan. 2017.
\newblock ISSN 0885-6125.
\newblock \doi{10.1007/s10994-016-5588-2}.
\newblock URL \url{https://doi.org/10.1007/s10994-016-5588-2}.

\bibitem[Laber et~al.(2018)Laber, Wu, Munera, Lipkovich, Colucci, and
  Ripa]{Labersafety18}
E.~B. Laber, F.~Wu, C.~Munera, I.~Lipkovich, S.~Colucci, and S.~Ripa.
\newblock {{I}dentifying optimal dosage regimes under safety constraints: {A}n
  application to long term opioid treatment of chronic pain}.
\newblock \emph{Stat Med}, 37\penalty0 (9):\penalty0 1407--1418, Apr 2018.

\bibitem[Leon et~al.(2003)Leon, Tsiatis, and
  Davidian]{tsiatisSemiparametricClass03}
S.~Leon, A.~A. Tsiatis, and M.~Davidian.
\newblock {Semiparametric Estimation of Treatment Effect in a Pretest-Posttest
  Study}.
\newblock \emph{Biometrics}, 59\penalty0 (4):\penalty0 1046--1055, 12 2003.
\newblock ISSN 0006-341X.
\newblock \doi{10.1111/j.0006-341X.2003.00120.x}.
\newblock URL \url{https://doi.org/10.1111/j.0006-341X.2003.00120.x}.

\bibitem[Massetti et~al.(2017)Massetti, Brown, Lapsa, Sharma, Bradbury,
  Cunliff, and Li]{Massetti2017}
E.~Massetti, M.~A. Brown, M.~V. Lapsa, I.~Sharma, J.~Bradbury, C.~Cunliff, and
  Y.~Li.
\newblock Environmental quality and the {U}.{S}. power sector: Air quality,
  land use and environmental justice.
\newblock \emph{Technical Report: ORNL/SPR-2016/772}, Jan. 2017.
\newblock \doi{10.2172/1339359}.
\newblock URL \url{https://doi.org/10.2172/1339359}.

\bibitem[Nethery et~al.(2020)Nethery, Mealli, Sacks, and Dominici]{Nethery2020}
R.~C. Nethery, F.~Mealli, J.~D. Sacks, and F.~Dominici.
\newblock Evaluation of the health impacts of the 1990 {C}lean {A}ir {A}ct
  {A}mendments using causal inference and machine learning.
\newblock \emph{Journal of the American Statistical Association}, 116\penalty0
  (535):\penalty0 1128--1139, Sept. 2020.
\newblock \doi{10.1080/01621459.2020.1803883}.
\newblock URL \url{https://doi.org/10.1080/01621459.2020.1803883}.

\bibitem[Ogburn et~al.(2024)Ogburn, Sofrygin, Díaz, and van~der
  Laan]{Ogburn24}
E.~Ogburn, O.~Sofrygin, I.~Díaz, and M.~van~der Laan.
\newblock Causal inference for social network data.
\newblock \emph{Journal of the American Statistical Association}, 119\penalty0
  (545):\penalty0 597--611, 2024.
\newblock \doi{10.1080/01621459.2022.2131557}.
\newblock URL \url{https://doi.org/10.1080/01621459.2022.2131557}.

\bibitem[Pinkse et~al.(2007)Pinkse, Shen, and Slade]{Pinkse07}
J.~Pinkse, L.~Shen, and M.~Slade.
\newblock A central limit theorem for endogenous locations and complex spatial
  interactions.
\newblock \emph{Journal of Econometrics}, 140\penalty0 (1):\penalty0 215--225,
  2007.
\newblock ISSN 0304-4076.
\newblock \doi{https://doi.org/10.1016/j.jeconom.2006.09.008}.
\newblock URL
  \url{https://www.sciencedirect.com/science/article/pii/S0304407606002296}.
\newblock Analysis of spatially dependent data.

\bibitem[Pouget-Abadie et~al.(2019)Pouget-Abadie, Aydin, Schudy, Brodersen, and
  Mirrokni]{PougetAbadieNeurips2019}
J.~Pouget-Abadie, K.~Aydin, W.~Schudy, K.~Brodersen, and V.~Mirrokni.
\newblock Variance reduction in bipartite experiments through correlation
  clustering.
\newblock In H.~Wallach, H.~Larochelle, A.~Beygelzimer, F.~d\textquotesingle
  Alch\'{e}-Buc, E.~Fox, and R.~Garnett, editors, \emph{Advances in Neural
  Information Processing Systems}, volume~32. Curran Associates, Inc., 2019.

\bibitem[Qiu et~al.(2022)Qiu, Carone, and Luedtke]{qiu2022individualized}
H.~Qiu, M.~Carone, and A.~Luedtke.
\newblock Individualized treatment rules under stochastic treatment cost
  constraints.
\newblock \emph{Journal of causal inference}, 10\penalty0 (1):\penalty0
  480--493, 2022.

\bibitem[Robins(2004)]{Robins2004}
J.~M. Robins.
\newblock \emph{Optimal Structural Nested Models for Optimal Sequential
  Decisions}, pages 189--326.
\newblock Springer New York, New York, NY, 2004.
\newblock ISBN 978-1-4419-9076-1.
\newblock \doi{10.1007/978-1-4419-9076-1_11}.

\bibitem[Samet et~al.(2000)Samet, Dominici, Curriero, Coursac, and
  Zeger]{Samet2000}
J.~M. Samet, F.~Dominici, F.~C. Curriero, I.~Coursac, and S.~L. Zeger.
\newblock Fine particulate air pollution and mortality in 20 {U}.{S}. cities,
  1987{\textendash}1994.
\newblock \emph{New England Journal of Medicine}, 343\penalty0 (24):\penalty0
  1742--1749, Dec. 2000.
\newblock \doi{10.1056/nejm200012143432401}.
\newblock URL \url{https://doi.org/10.1056/nejm200012143432401}.

\bibitem[Schulte et~al.(2014)Schulte, Tsiatis, Laber, and
  Davidian]{Schulte_2014}
P.~J. Schulte, A.~A. Tsiatis, E.~B. Laber, and M.~Davidian.
\newblock $\mathbf{Q}$- and $\mathbf{A}$-learning methods for estimating
  optimal dynamic treatment regimes.
\newblock \emph{Statistical Science}, 29\penalty0 (4), Nov. 2014.
\newblock ISSN 0883-4237.
\newblock \doi{10.1214/13-sts450}.
\newblock URL \url{http://dx.doi.org/10.1214/13-STS450}.

\bibitem[Song and Papadogeorgou(2024)]{song2024bipartite}
Z.~Song and G.~Papadogeorgou.
\newblock Bipartite causal inference with interference, time series data, and a
  random network, 2024.

\bibitem[Srivastava et~al.(2001)Srivastava, Jozewicz, and
  Singer]{Srivastava2001}
R.~K. Srivastava, W.~Jozewicz, and C.~Singer.
\newblock {SO}2 scrubbing technologies: A review.
\newblock \emph{Environmental Progress}, 20\penalty0 (4):\penalty0 219--228,
  Dec. 2001.
\newblock \doi{10.1002/ep.670200410}.
\newblock URL \url{https://doi.org/10.1002/ep.670200410}.

\bibitem[Stürmer et~al.(2010)Stürmer, Rothman, Avorn, and
  Glynn]{StrumerTrimSim2010}
T.~Stürmer, K.~J. Rothman, J.~Avorn, and R.~J. Glynn.
\newblock {Treatment Effects in the Presence of Unmeasured Confounding: Dealing
  With Observations in the Tails of the Propensity Score Distribution—A
  Simulation Study}.
\newblock \emph{American Journal of Epidemiology}, 172\penalty0 (7):\penalty0
  843--854, 08 2010.
\newblock ISSN 0002-9262.
\newblock \doi{10.1093/aje/kwq198}.
\newblock URL \url{https://doi.org/10.1093/aje/kwq198}.

\bibitem[Su et~al.(2019)Su, Lu, and Song]{suModelingEstimation}
L.~Su, W.~Lu, and R.~Song.
\newblock Modelling and estimation for optimal treatment decision with
  interference.
\newblock \emph{Stat}, 8\penalty0 (1):\penalty0 e219, 2019.

\bibitem[Tsai et~al.(2003)Tsai, Goggins, Chiu, and Yang]{Tsai2003}
S.-S. Tsai, W.~B. Goggins, H.-F. Chiu, and C.-Y. Yang.
\newblock Evidence for an association between air pollution and daily stroke
  admissions in {K}aohsiung, {T}aiwan.
\newblock \emph{Stroke}, 34\penalty0 (11):\penalty0 2612--2616, Nov. 2003.
\newblock \doi{10.1161/01.str.0000095564.33543.64}.
\newblock URL \url{https://doi.org/10.1161/01.str.0000095564.33543.64}.

\bibitem[{U.S. Environmental Protection
  Agency}(2022{\natexlab{a}})]{usepa2022a}
{U.S. Environmental Protection Agency}.
\newblock Policy assessment for the reconsideration of the national ambient air
  quality standards for particulate matter.
\newblock \emph{Technical Report: EPA-452/R-22-004}, May 2022{\natexlab{a}}.

\bibitem[{U.S. Environmental Protection
  Agency}(2022{\natexlab{b}})]{usepa2022b}
{U.S. Environmental Protection Agency}.
\newblock Regulatory impact analysis for the proposed reconsideration of the
  national ambient air quality standards for particulate matter.
\newblock \emph{Technical Report: EPA-452/P-22-001}, Dec. 2022{\natexlab{b}}.

\bibitem[Viviano(2024)]{viviano2024policy}
D.~Viviano.
\newblock Policy targeting under network interference.
\newblock \emph{Review of Economic Studies}, page rdae041, 2024.

\bibitem[Viviano and Bradic(2024)]{viviano2024fair}
D.~Viviano and J.~Bradic.
\newblock Fair policy targeting.
\newblock \emph{Journal of the American Statistical Association}, 119\penalty0
  (545):\penalty0 730--743, 2024.

\bibitem[Viviano and Rudder(2020)]{viviano2020policy2}
D.~Viviano and J.~Rudder.
\newblock Policy design in experiments with unknown interference.
\newblock \emph{arXiv preprint arXiv:2011.08174}, 2020.

\bibitem[Wu et~al.(2020)Wu, Braun, Schwartz, Kioumourtzoglou, and
  Dominici]{Wu2020}
X.~Wu, D.~Braun, J.~Schwartz, M.~A. Kioumourtzoglou, and F.~Dominici.
\newblock Evaluating the impact of long-term exposure to fine particulate
  matter on mortality among the elderly.
\newblock \emph{Science Advances}, 6\penalty0 (29), July 2020.
\newblock \doi{10.1126/sciadv.aba5692}.
\newblock URL \url{https://doi.org/10.1126/sciadv.aba5692}.

\bibitem[Xu et~al.(2024)Xu, Fu, and Qu]{qi2024jrssb}
Q.~Xu, H.~Fu, and A.~Qu.
\newblock {Optimal individualized treatment rule for combination treatments
  under budget constraints}.
\newblock \emph{Journal of the Royal Statistical Society Series B: Statistical
  Methodology}, page qkad141, 01 2024.
\newblock ISSN 1369-7412.
\newblock \doi{10.1093/jrsssb/qkad141}.
\newblock URL \url{https://doi.org/10.1093/jrsssb/qkad141}.

\bibitem[Zeger et~al.(2008)Zeger, Dominici, McDermott, and Samet]{Zeger2008}
S.~L. Zeger, F.~Dominici, A.~McDermott, and J.~M. Samet.
\newblock Mortality in the {M}edicare population and chronic exposure to fine
  particulate air pollution in urban centers (2000{\textendash}2005).
\newblock \emph{Environmental Health Perspectives}, 116\penalty0 (12):\penalty0
  1614--1619, Dec. 2008.
\newblock \doi{10.1289/ehp.11449}.
\newblock URL \url{https://doi.org/10.1289/ehp.11449}.

\bibitem[Zhang and Imai(2024)]{zhang2024individualized}
Y.~Zhang and K.~Imai.
\newblock Individualized policy evaluation and learning under clustered network
  interference, 2024.

\bibitem[Zigler et~al.(2023)Zigler, Liu, Forastiere, and Mealli]{Zigler2020}
C.~Zigler, V.~Liu, L.~Forastiere, and F.~Mealli.
\newblock Bipartite interference and air pollution transport: Estimating health
  effects of power plant interventions.
\newblock \emph{arXiv Preprint}, 2023.

\bibitem[Zigler and Papadogeorgou(2021)]{Zigler2021}
C.~M. Zigler and G.~Papadogeorgou.
\newblock Bipartite causal inference with interference.
\newblock \emph{Statistical Science}, 36\penalty0 (1), Feb. 2021.
\newblock \doi{10.1214/19-sts749}.
\newblock URL \url{https://doi.org/10.1214/19-sts749}.

\end{thebibliography}

\clearpage 


\begin{table}[!ht]
\begin{center}
\begin{tabular}
{ |p{2.25cm}p{1.35cm}p{1.35cm}p{3.5cm}p{2cm}p{2.1cm}|}
 \hline
 \multicolumn{6}{|c|}{Simulation Results} \\
 \hline
 Method & BS ($f_0$) & PS ($e$) & \Err & \TEErr  & Coverage \\
 \hline
 \ql & \cmark & -  & 0.21 & 0.73 & 94.84 \\
 \ql & \xmark & -  & 2.24 & 7.10 & 15.40 \\ 
  \hline
 \al & \cmark & \cmark & 0.24 & 1.32 & 98.57 \\
 \al & \cmark & \xmark & 0.24 & 1.32 & 98.37 \\
 \al & \xmark & \cmark & 0.57 & 1.74 & 95.23 \\
 \al & \xmark & \xmark & 0.58 & 1.75 & 94.56 \\
 \hline
\end{tabular}
\caption{\textmd{Simulation Results over 1000 iterations. BS denotes baseline model, $f_0$, and PS denotes propensity score model, $e$. See the definitions for \Err (\cref{eq:betaError}) and \TEErr (\cref{eq:teError}) above. For presentation, $10 \cdot \Err$ is displayed.}
\label{tab:simResults_func}}
\end{center}
\end{table}

\begin{figure}[ht!]
\centering
 \resizebox{\textwidth}{!}{
\begin{tabular}{ccc}
    \includegraphics[width=0.33\textwidth]{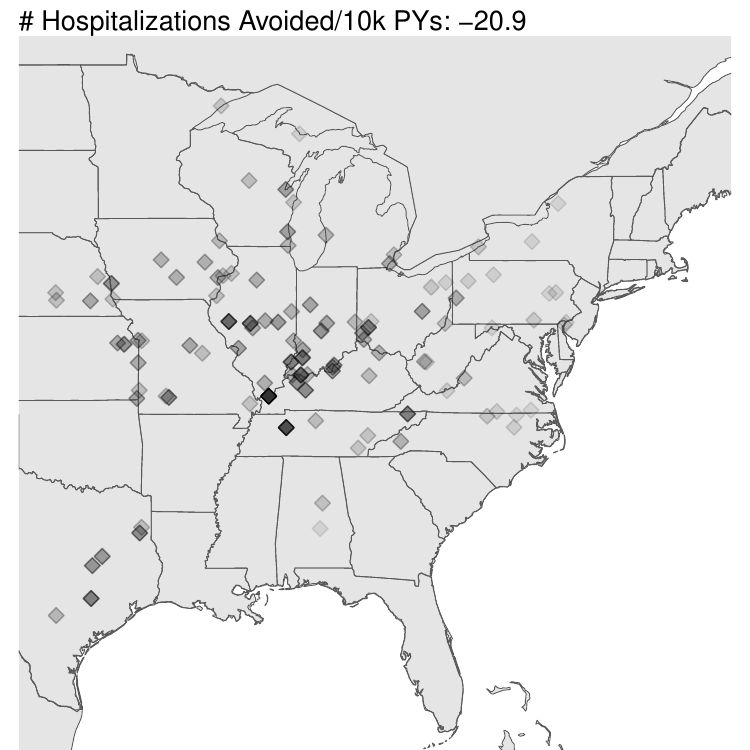}
     &
     \includegraphics[width=0.33\textwidth]{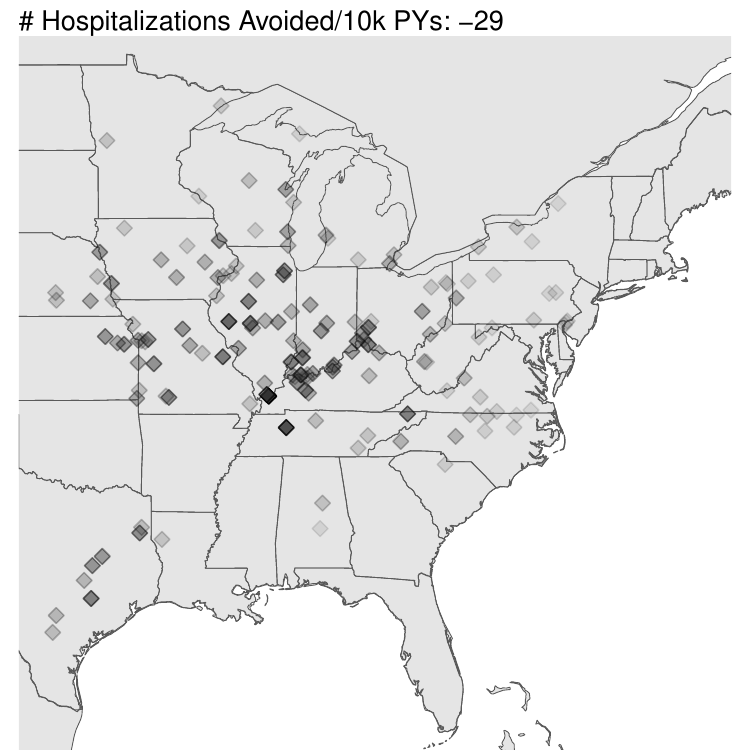} &

     \includegraphics[width=0.33\textwidth]{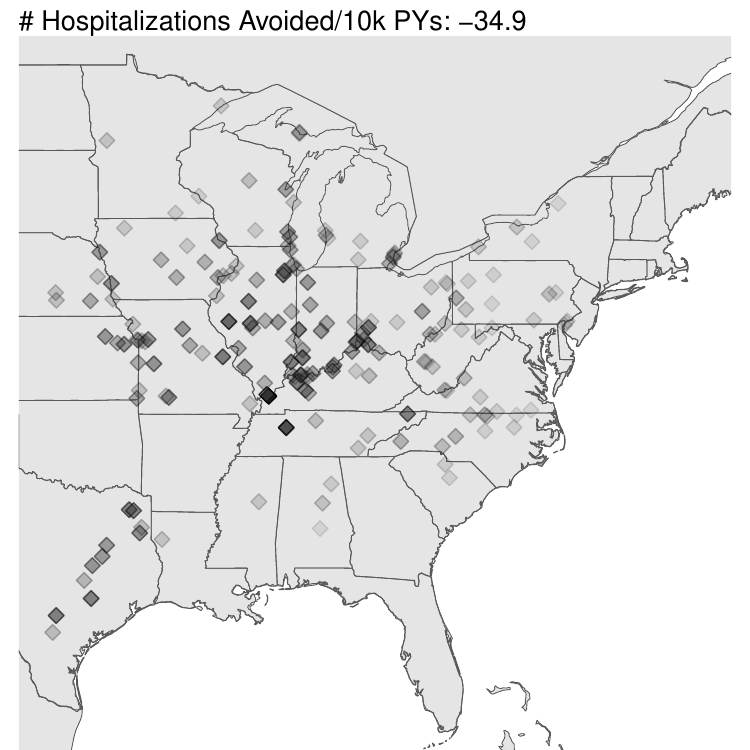}
     
     \\
     (a) 10\% Total Cost & (b) 20\% Total Cost  & (c) 30\% Total Cost 
\end{tabular}}

 \resizebox{\textwidth}{!}{
\begin{tabular}{ccc}
    \includegraphics[width=0.33\textwidth]{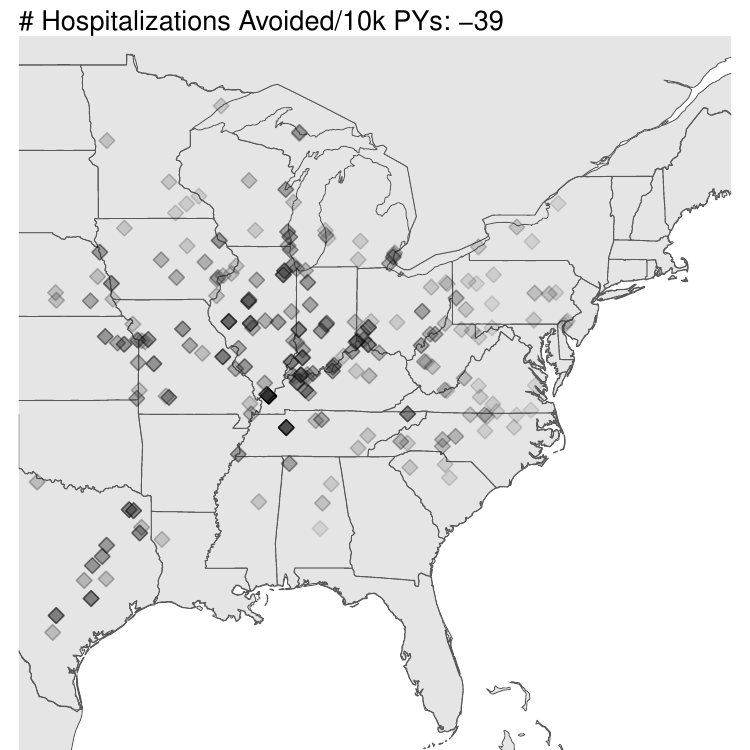}
     &
     \includegraphics[width=0.33\textwidth]{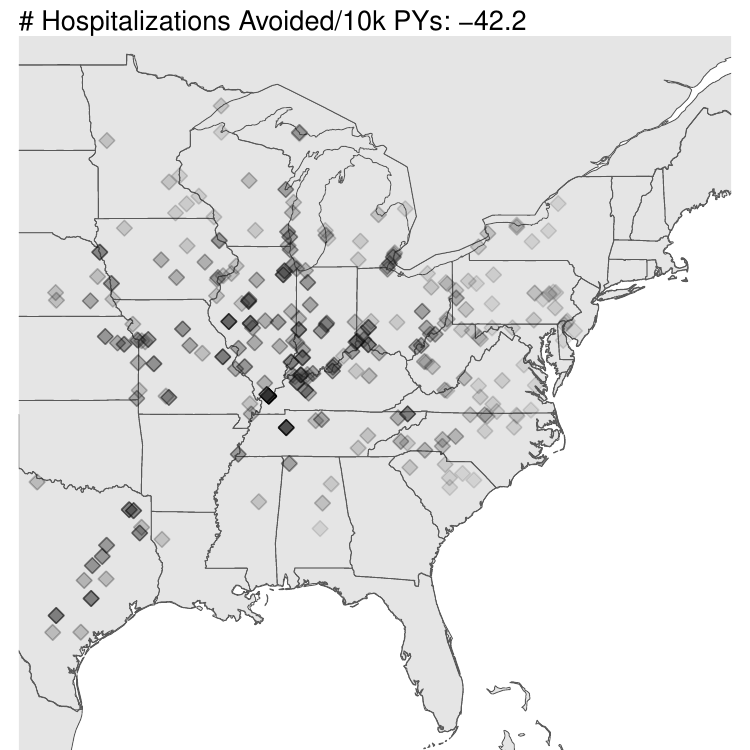} &

     \includegraphics[width=0.33\textwidth]{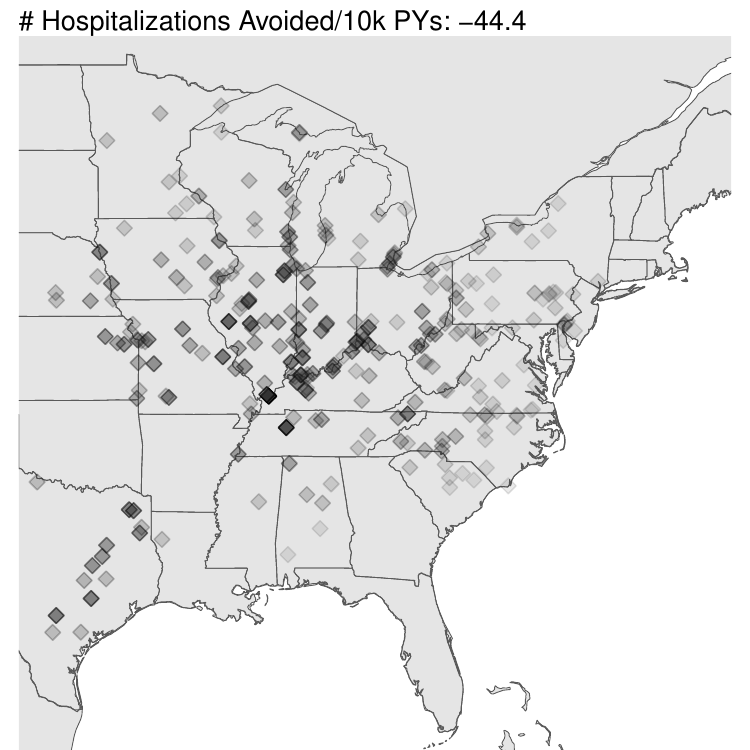}
     
     \\
     (a) 40\% Total Cost & (b) 50\% Total Cost  & (c) 60\% Total Cost 
\end{tabular}}

 \resizebox{\textwidth}{!}{
\begin{tabular}{ccc}
    \includegraphics[width=0.33\textwidth]{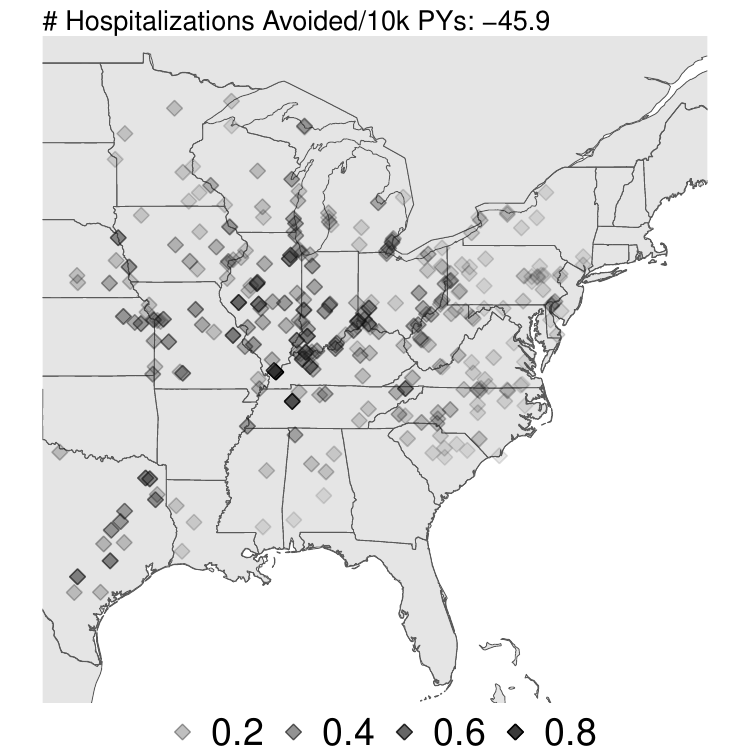}
     &
     \includegraphics[width=0.33\textwidth]{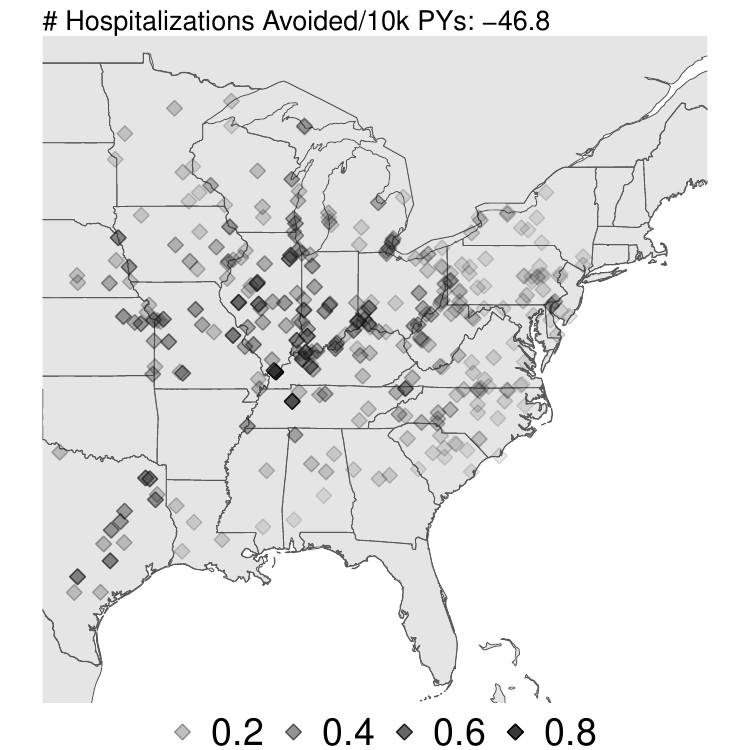} &

     \includegraphics[width=0.33\textwidth]{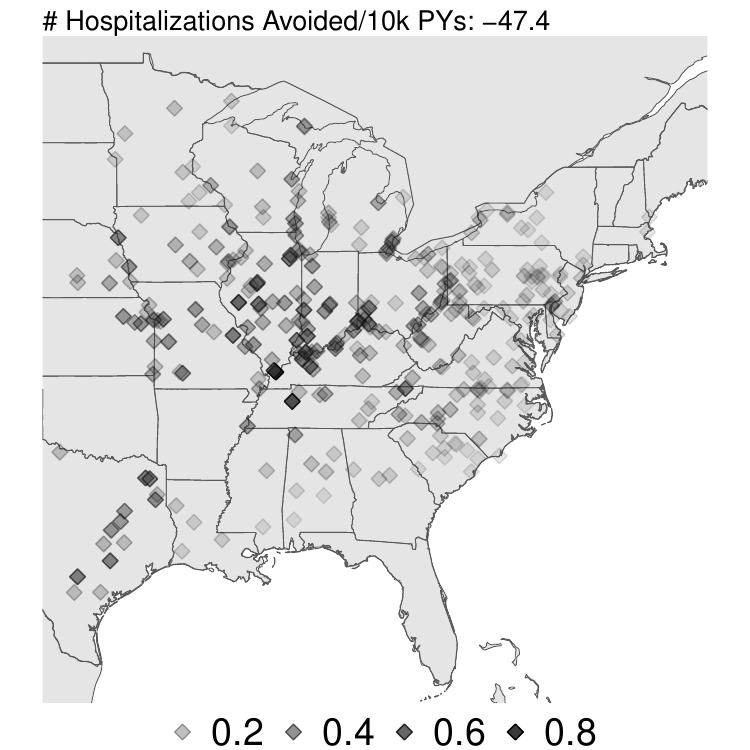}
     
     \\
     (a) 70\% Total Cost & (b) 80\% Total Cost  & (c) 90\% Total Cost 
\end{tabular}}
\caption{\textmd{Grid with the reduction of IHD Hospitalizations/10,000 Person Years, varying the spending from 10\%-90\% of budget } \label{fig:panelBudgetRate}}
\end{figure}

\label{LastPage}

\clearpage 

\markboth%
{Supplementary Material}{Supplementary Material}

\crefname{section}{Appendix}{Appendices}

\counterwithin{equation}{section}
\counterwithin{figure}{section}
\counterwithin{table}{section}
\pagenumbering{arabic}
\setcounter{page}{1}

\renewcommand{\thesection}{Appendix~\Alph{section}}

\begin{appendices}

\begin{center}
    \textbf{SUPPLEMENTARY MATERIAL TO\\ ``Towards Optimal Environmental Policies: Policy Learning under Arbitrary Bipartite Network Interference''}\\ \vspace{0.25cm}
    \normalsize Raphael C. Kim$^1$, Falco J. Bargagli-Stoffi$^2$, Kevin L. Chen$^1$, Rachel C. Nethery$^1$ \\
    {\small $^1$Department of Biostatistics, Harvard T.H. Chan School of Public Health, 677 Huntington Ave, 02215, MA, USA}\\
    {\small $^2$Department of Biostatistics, UCLA Fielding School of Public Health, 650 Charles E Young Dr S, 90095, CA, USA}
\end{center}

\section{Proofs}\label{sec:pfs}
The inferential results below employ the strategy from \cite{suModelingEstimation} and the regret bounds employ the strategy from \cite{KuznetsovMohri17}. However, we require results in a setting where there is arbitrary bipartite network interference. Since the setup is unique, we explicitly write out the results.

\subsection{Proof of \texorpdfstring{\cref{thm:identification}}{Theorem}} \label{pf:identification}
    \begin{align*}
        \mathbb{E}[Y_i(\ei)] &= \mathbb{E}[\mathbb{E}[Y_i(\ei) \mid \Xouti, \XJ, \T_i]] \\
        & \seq{\refAssump{I}{3}} \mathbb{E}[\mathbb{E}[Y_i(\ei) \mid \Xouti, \XJ, \T_i, \Ei=\ei]] \\
        & \seq{\refAssump{I}{4}} \mathbb{E}[\mathbb{E}[Y_i(\ei) \mid \Xouti, \T_i, \Ei=\ei]]  \\
        & \seq{\refAssump{I}{1}} \mathbb{E}[\mathbb{E}[Y_i \mid \Xouti, \T_i, \Ei=\ei]] 
    \end{align*}
        
\subsection{Proof of Q-Learning Inference}\label{pf:qLearning}


\paragraph{Inference Assumptions}
\begin{enumerate}[\bfseries ({R}1) ]
\label{ass:QLearning}
    \item \label{ass:bdd} The outcome covariates, intervention covariates, and interference map are bounded. $|| \Xouti ||_\infty < \infty$, $|| \Xintj ||_\infty < \infty, ||\T|| < \infty$
    \item \label{ass:regular}  Parameters $\outCoef$ lie in a compact space $\mathbf{\Theta}$ with $\outCoef_0$ as an inner point.
    \item \label{ass:limitsQLearning_a} $\Sigma_d=\Lim{n \rightarrow \infty} \frac{1}{n} \sum_{i=1}^n [-\frac{ \partial \phi(Y_i; \outCoef)}{\partial \outCoef}|_{\outCoef=\outCoef_0}]$ is finite and non degenerate
    \item \label{ass:limitsQLearning_psi} $\Sigma_\phi=\Lim{n \rightarrow \infty} \frac{1}{n} \sum_{i=1}^n  \phi(Y_i; \outCoef_0) \phi(Y_i; \outCoef_0)^\top$ is positive semi-definite.
\end{enumerate}

    \begin{proof}
        Using Taylor's theorem:  
        \begin{align*}
            \sqrt{n}(\hat{\outCoef}-\outCoef_0) &= [-\frac{1}{n}\sum_{i=1}^n \frac{\partial \phi(Y_i; \outCoef)}{\partial \outCoef^\top}|_{\outCoef=\outCoef_0}]^{-1}[\frac{1}{\sqrt{n}} \sum_{i=1}^n \phi(Y_i; \outCoef_0)] + o_p(1)
        \end{align*}

    Then invoking \ref{ass:limitsQLearning_a} and \ref{ass:limitsQLearning_psi}, by the multiplier central limit theorem \cref{thm:mclt}, we find that $\frac{1}{\sqrt{n}} \sum_{i=1}^n \phi(Y_i; \outCoef_0) \xrightarrow{d} N(0, \Sigma_\phi(\outCoef_0))$ where $\Sigma_\phi=\frac{1}{n}\phi(Y_i; \outCoef)\phi(Y_i; \outCoef)^\top=\frac{1}{n}\sum_{i=1}^n d_i(\outCoef) d_i(\outCoef)^\top (Y_i - \mu(\outCoef))^2$.

    Consistent variance estimation follows by $\hat{\Sigma}=\Sigma(\hat{\outCoef})=\frac{1}{n}\Sigma_d^{-1}(\hat{\outCoef}) \Sigma_\phi(\hat{\outCoef}) (\Sigma_d^{-1}(\hat{\outCoef}))^\top$
    for  $\Sigma_d(\outCoef)=\frac{1}{n} d_i(\outCoef) d_i(\outCoef)^\top$.
    \end{proof}

\subsection{Proof of \texorpdfstring{\cref{thm:unbiased}}{Lemma}} \label{pf:unbiased}
    \begin{proof}
    \textbf{Case I.} Baseline model $f_0$ is misspecified, propensity score model $e$ is correctly specified.
    \begin{align*}
        \mathbb{E}[\Phi_n(\outCoeftx_0,\outCoefb_0,\intCoef_0)] & \seq{\refAssump{I}{\cref{ass:consistency}}} \mathbb{E}[\mathbb{E}[\Phi_n(\outCoeftx_0,\outCoefb_0,\intCoef_0) \mid \Xj, \Xouti, \T_i, Y_i(\ei)]]
    \end{align*}

    Consider a particular summand of $\E[\Phi_n(\outCoeftx_0,\outCoefb_0,\intCoef_0)]$,
    \begin{align*}
        \E[\E[\phi_{i}(\outCoeftx_0,\outCoefb_0,\intCoef_0) & \mid \Xj, \Xouti, \T_i,  Y_i(\ei) ]]\\
        &=\E[\lambda(\Xouti, \T_i; \outCoeftx) (Y_i- \muTrue)  \\
            &\quad \quad \quad \quad \quad \cdot \E[\sum_{j=1}^J \Tinst \E[(A_j - \pi(\Xj, \intCoef_0)) \mid \Xj, \Xouti, Y_i(\Ei),\T_i]] ] 
    \end{align*}
    Since $\pi$ is correctly specified, the inner expectation evaluates to 0.

   \textbf{Case II.} Baseline model $f_0$ is correctly specified, propensity score model $p$ is incorrectly specified.
    \begin{align*}
        \mathbb{E}[\Phi_n(\outCoeftx_0,\outCoefb_0,\intCoef_0)] &= \mathbb{E}[\mathbb{E}[\Phi_n(\outCoeftx_0,\outCoefb_0,\intCoef_0) \mid \Xouti, \XJ, \T_i]] \\
        &\seq{\refAssump{I}{3}} \mathbb{E}[\mathbb{E}[\Phi_n(\outCoeftx_0,\outCoefb_0,\intCoef_0) \mid \Xouti, \XJ, \T_i, \Ei=\ei]] \\
        &\seq{\refAssump{I}{4}} \mathbb{E}[\mathbb{E}[\Phi_n(\outCoeftx_0,\outCoefb_0,\intCoef_0) \mid \Xouti, \T_i, \Ei=\ei]]  
    \end{align*}

    Consider a particular summand: 
    \begin{align*}
        \mathbb{E}[\mathbb{E}[\phi_{ni}(\outCoeftx_0,\outCoefb_0,\intCoef_0) & \mid \Xouti, \T_i, \Ei=\ei]] \\
        &=  \E[\lambda_i(\Xouti, \T_i; \outCoeftx) (\Ei -  \EiHat) \\
        &\quad \quad \quad \quad \quad \cdot \E[\E[(Y_i- \muTrue) \mid \Xouti, \T_i, \Ei=\ei]]]
    \end{align*}
    Since $\mu$ is correctly specified, the inner expectation evaluates to 0.
    \end{proof}
    
\subsection{Proof for \texorpdfstring{\cref{thm:aLearning}}{Theorem}}\label{pf:aLearning}
To perform inference of $\E[Y_i(A; \outCoef_0)]$ under A Learning, we will require the following assumptions:

\paragraph{Inference Assumptions}
    \begin{enumerate}[{\bfseries ({R}1') }]\label{ass:ALearning}
        \item \label{ass:bdd_al} The outcome covariates, intervention covariates, and interference map are bounded. $|| \Xouti ||_\infty < \infty$, $|| \Xintj ||_\infty < \infty, ||\T|| < \infty$
        \item \label{ass:regular_param_space_gamma} Regularity of parameter space. Parameters $(\intCoef, \outCoef)$ lie in a compact space $\mathbf{\Gamma \cup \Theta}$ with $(\intCoef_0, \outCoef_0)$ as an inner point.
        \item \label{ass:limitsALearning_a} $\Sigma_{\beta,\alpha}$, $\Sigma_\phi$ and $\Sigma_\gamma$, defined below in \cref{pf:aLearning}, is finite and has a non-degenerate limit
        \item \label{ass:limitsALearning_psi} $\Omega_\psi$, defined below in \cref{pf:aLearning}, is positive semi-definite.
        \item \label{ass:limitsgamma_a} $V_d=\Lim{J \rightarrow \infty} \frac{1}{J} \sum_{j=1}^J [-\frac{ \partial \phi_{\intCoef} (A_j; \intCoef)}{\partial \intCoef}|_{\intCoef=\intCoef_0}]$ is finite and non degenerate
        \item \label{ass:limitsgamma_psi} $V_\phi=\Lim{J \rightarrow \infty} \frac{1}{J} \sum_{j=1}^J  \phi(A_j; \intCoef_0) \phi(A_j; \intCoef_0)^\top$ is positive semi-definite.
        \item \label{ass:sampleRatios}
        As $n,J \rightarrow \infty$, $\frac{J}{n} \rightarrow \ratioJn$ a fixed constant.
    \end{enumerate}


     Assumptions \refAssump{R}{\labelcref{ass:limitsALearning_a}'}-\refAssump{R}{\labelcref{ass:limitsALearning_psi}'} are for a well-defined covariance matrix similar to 
     \refAssump{R}{\labelcref{ass:limitsQLearning_a}}-\refAssump{R}{\labelcref{ass:limitsQLearning_psi}} in \cref{thm:qLearning}. Assumptions 
     \refAssump{R}{\labelcref{ass:limitsgamma_a}'}-\refAssump{R}{\labelcref{ass:limitsgamma_psi}'} in \cref{thm:qLearning} are primarily for showing \cref{thm:can_intCoef}, which shows we have asymptotic normality of estimating $\intCoef_0$. 

    \begin{proof}

    Let $\PhiAug$ be $\Phi_n$ augmented with \cref{eq:alearning_alpha}. For example, if $f_0$ is linear in $\Xouti$, we will augment with $(1, \Xouti)$. I.e.

    \begin{align}
    \PhiAug &= \begin{bmatrix}
           \sum_{i=1}^n &\frac{\partial}{\partial \outCoefb} f_0(\Xouti, \outCoefb) (Y_i- \muGeneralAB) \\
           \sum_{i=1}^n \sum_{j=1}^J & \lambda(\Xouti, \T_i; \outCoeftx) (Y_i- \muGeneralAB )) \\
    &\quad\quad\quad\quad\quad \cdot \frac{1}{J} \Tinst(A_j - \pi(\Xintj, \intCoef)) 
         \end{bmatrix}
    \end{align}

    Take a Taylor expansion to find
    \begin{align*}
    \PhiAug(\hat{\outCoeftx}, \hat{\outCoefb}, \hat{\intCoef}) &= \PhiAug(\outCoeftx_0, \outCoefb_0, \intCoef_0) + \frac{\partial \PhiAug(\outCoeftx, \outCoefb, \intCoef_0)}{\partial (\outCoeftx^\top, \outCoefb^\top)}|_{\outCoeftx=\outCoeftx_0, \outCoefb=\outCoefb_0} 
    \thetaMatrix \\
    &\quad\quad\quad+ \frac{\partial \PhiAug(\outCoeftx_0, \outCoefb_0, \intCoef)}{\partial \intCoef^\top}|_{\intCoef=\intCoef_0} (\hat{\intCoef}-\intCoef_0) + \frac{1}{2} T_2 + R_{nJ} 
    \end{align*}
    where 
    \begin{align*}
        T_2 &= \thetaMatrix^\top \frac{\PhiAug(\outCoeftx, \outCoefb, \intCoef_0)}{\partial(\outCoeftx^\top, \outCoefb^\top) \partial(\outCoeftx, \outCoefb)}|_{\outCoeftx=\outCoeftx_0, \outCoefb=\outCoefb_0} \thetaMatrix  \\
        & \quad\quad\quad +  2 \thetaMatrix^\top \frac{\PhiAug(\outCoeftx, \outCoefb, \intCoef)}{\partial(\outCoeftx^\top, \outCoefb^\top) \partial(\intCoef^\top)}|_{\outCoeftx=\outCoeftx_0, \outCoefb=\outCoefb_0, \intCoef=\intCoef_0} (\hat{\intCoef}-\intCoef_0) \\
        & \quad\quad\quad +  (\hat{\intCoef}-\intCoef_0)^\top \frac{\PhiAug(\outCoeftx_0, \outCoefb_0, \intCoef)}{\partial \intCoef^\top \partial \intCoef}|_{\intCoef=\intCoef_0} (\hat{\intCoef}-\intCoef_0)
    \end{align*}
    and 

    \begin{align*}
        R_{nJ} &= ||\thetaMatrix, \hat{\intCoef}-\intCoef_0 ||^3 \frac{ \sqrt{2}^3}{3!}\cdot \sup_{(\outCoeftx,\outCoefb), \intCoef} |\frac{\partial^3}{\partial^a (\outCoeftx,\outCoefb) \partial^b \intCoef} \PhiAug(\outCoeftx,\outCoefb,\intCoef)| , \quad a+b = 3
    \end{align*}

    Under regularity conditions 
    \refAssump{R}{\labelcref{ass:bdd_al}'}-\refAssump{R}{\labelcref{ass:limitsALearning_psi}'}, we have consistency of $\hat{\outCoeftx}, \hat{\outCoefb}, \hat{\intCoef}$ and boundedness of these derivatives. Hence, $T_2=R_{nJ}=o_P(\sqrt{n})$, noting that remainder terms corresponding to $\intCoef$ are $o_P(\sqrt{J})=o_P(\sqrt{n})$ by \refAssump{R}{\labelcref{ass:sampleRatios}'}.
        
    Rearranging, we have
    \begin{align*}
    & \thetaMatrix= \\
    &\quad\quad (\frac{\partial \PhiAug(\outCoeftx, \outCoefb, \intCoef_0)}{\partial (\outCoeftx^\top, \outCoefb^\top)}\Bigr|_{\substack{\outCoeftx=\outCoeftx_0 \\ \outCoefb=\outCoefb_0}})^{-1} [\PhiAug(\outCoeftx_0, \outCoefb_0,\intCoef_0) 
    +\frac{\partial \PhiAug(\outCoeftx_0, \outCoefb_0, \intCoef)}{\partial \intCoef^\top}\Bigr|_{\intCoef=\intCoef_0} (\hat{\intCoef}-\intCoef_0)] \\
    &\quad\quad\quad\quad + o_p(1/\sqrt{n})
    \end{align*}

    Let $\Sigma_{\beta, \alpha}= \Lim{n \rightarrow \infty}\frac{1}{n}\frac{\partial \PhiAug(\outCoeftx, \outCoefb, \intCoef_0)}{\partial (\outCoeftx^\top, \outCoefb^\top)}\Bigr|_{\substack{\outCoeftx=\outCoeftx_0 \\ \outCoefb=\outCoefb_0}} $ and $\Sigma_\gamma=\Lim{n \rightarrow \infty}\frac{1}{n}\frac{\partial \PhiAug(\outCoeftx_0, \outCoefb_0, \intCoef)}{\partial \intCoef^\top}\Bigr|_{\intCoef=\intCoef_0} $. 
    
    Invoking \refAssump{R}{\labelcref{ass:limitsALearning_a}'}, we have
    \begin{align*}
         \thetaMatrix &=  \Sigma_{\beta, \alpha}^{-1} [\PhiAug(\outCoeftx_0, \outCoefb_0,\intCoef_0) 
        +\Sigma_\gamma (\hat{\intCoef}-\intCoef_0)] + o_p(1/\sqrt{n})
    \end{align*}

    Then we have,
    \begin{align*}
        & \sqrt{n} \thetaMatrix = \Sigma_{\beta,\alpha}^{-1} [\frac{1}{\sqrt{n}} \PhiAug(\outCoeftx_0, \outCoefb_0,\intCoef_0) 
        +\Sigma_\gamma \sqrt{n} (\hat{\intCoef}-\intCoef_0)] + o_p(1)
    \end{align*}

    Let a summand of $\PhiAug$ be denoted by 
    \begin{align}
    \phi_{i}^\mathrm{aug} &= \begin{bmatrix}
           &\frac{\partial}{\partial \outCoefb} f_0(A; \Xouti, \outCoefb) (Y_i- \mu(A; \Xouti, \T_i, \outCoefb, \outCoeftx)) \\
           &\lambda(\Xouti, \T_i; \outCoeftx) (Y_i- \mu(A; \Xouti, \T_i, \outCoefb, \outCoeftx) )) \\
    &\quad\quad\quad\quad\quad \cdot \frac{1}{J} \sum_{j=1}^J \Tinst (A_j -\pi(\Xintj, \intCoef)) 
         \end{bmatrix}
    \end{align}

    Then, by \cref{thm:can_intCoef},

    \begin{align*}
         \sqrt{n} \thetaMatrix &= \Sigma_{\beta, \alpha}^{-1} \frac{1}{\sqrt{n}}\sum_{i=1}^n\phi_{i}^{\mathrm{aug}} + \Sigma_{\beta, \alpha}^{-1}\Sigma_\gamma \frac{1}{\sqrt{\ratioJn}} \frac{1}{\sqrt{J}} \sum_{j=1}^J \varepsilon_j + o_p(1)
    \end{align*}
    Note that conditioning on $\Xouti, \Xintj,\T_i, \mathbf{A}$, each term on the RHS is mean 0. 

    Let $\Sigma_\phi=\Lim{n \rightarrow \infty}\sum_{i=1}^n \phi_{i}^\mathrm{aug} (\phi_{i}^\mathrm{aug})^\top$, and define 

    \begin{equation}
        \Omega_\phi=(\Sigma_{\beta,\alpha}^{-1})\Sigma_\phi(\Sigma_{\beta,\alpha}^{-1})^\top
    \end{equation}

    \begin{equation}
        \Omega_\gamma=(\frac{1}{\sqrt{\ratioJn}}\Sigma_{\beta,\alpha}^{-1}\Sigma_\gamma)\Omega_\varepsilon(\frac{1}{\sqrt{\ratioJn}}\Sigma_{\beta,\alpha}^{-1}\Sigma_\gamma)^\top
    \end{equation}
    
    Then we can apply the multiplier central limit theorem \ref{thm:mclt} to each term to conclude asymptotic normality with variance covariance given by 
    \begin{equation}
        \Omega=\Omega_\phi+\Omega_\gamma
    \end{equation}

    In practice, we need to estimate $\ratioJn$ using $\frac{J}{n}$. By Slutsky's, the result follows. 
\end{proof}

\subsection{Proof for \texorpdfstring{\cref{thm:TE_Consistency}}{Theorem}}\label{pf:TE_Consistency}

\begin{proof}
    Below, proofs will be conditional on $\T_i$. 
    By Jensen's, it suffices to bound $|\E [\etaEst_j-\eta_j]|$.
    \begin{align*}
        \E \frac{1}{n}(\etaEst_j - \eta_j) &= \E_n[\Tinst (f_A(\Xouti, \hat{\outCoeftx})-\trueFA)] \\
        &= \E_n [\Tinst (f_A(\Xouti, \hat{\outCoeftx})- \E f_A(\Xouti, \hat{\outCoeftx}))+ \Tinst (\E f_A(\Xouti, \hat{\outCoeftx}) - f_A(\Xouti, \outCoeftx))]\\
        &\leq \E \sup_{\outCoeftx_s} \frac{1}{n} \sum_i (\Tinst (f_A(\Xouti, \outCoeftx_s)-\E f_A(\Xouti, \outCoeftx_s))) 
    \end{align*}
    where the 3rd line follows from our unbiased estimation equation. 
    We choose to bound this quantity using symmetrization and spatial mixing assumptions combined with independent blocking from \cite{Bernstein1927}. In this way, we handle the spatial dependence and present an argument easily extensible to more general, nonparametric function classes.
    
    We formalize our spatial setup as in \cite{Jenish2009}. Let $D_n \subseteq D \subseteq \Rtwo$ for $D_n$ a finite subset. $\Xouti$ lie in $D_n$. Define $L(i)$ as the location function; $L(i)$ returns the 2d coordinate location of unit $i$ in $D$.
    Take metric $\rho(i, i') =\max_{1 \leq k \leq 2} |L(i)_k - L(i')_k|$, which induces a norm $|i| = \max_{1 \leq k \leq 2} |L(i)_k|$, for $L(i)_k$ denoting the $k$-th component of the coordinate location of unit $i$. The distance between two subsets $U,V \subset D$ is defined as $\rho(U,V) = \inf \{\rho(i, i') : i \in U, i' \in V \}$, and the cardinality of a finite subset $U$ of $D$ is denoted $|U|$. Further define the spatial mixing coefficient (overloading notation on $\beta$):

        For $U \subset D_n, V \subset D_n$, define $\sigma_n(U)=\sigma(\Xouti: i \in U)$, the sigma field over $U$, and $\beta_n(U,V)=\beta(\sigma_n(U), \sigma_n(V))=\frac{1}{2} \sup \sum_{U'_s} \sum_{V'_{r}} |\pr(U'_s \cap V'_r)-\pr(U'_s)\pr(V'_r)|$ where $\{ U'_s \}, \{ V'_r\}$ are finite partitions of $U$ and $V$ respectively \citep{BradleySurvey05}. Then $\beta$ mixing coefficient on $D_n$ is given by:
        $$\beta_{s,s'}(r)=\sup\{ \beta_n(U,V): |U| \leq s, |V| \leq s', \rho(U,V) \geq r \}$$

        Also define $\overline{\beta}_{s,s'}(r)=\sup_n \beta_{s,s',n}(r)$ to account for sampling region variability.

    \paragraph{Spatial Assumptions}
    \begin{enumerate}[\bfseries ({S}1) ]
        \item \label{ass:samplingRegime} \textit{Infinite Sampling Regime}: \label{ass:infiniteSampling} $D$ is infinitely countable and all elements are at least $\rho_0 > 1$ apart for all $i, i' \in D$.
        \item \label{ass:mixing} \textit{Spatial Mixing}: For $\tau > 2$,
        $r \overline{\beta}_{s,s'}(r) \leq (s+s') \cdot r^{-\tau}$
    \end{enumerate}    
    These assumptions are standard in spatial statistics literature \citep{Jenish2009}. 
    With the setup from (\textbf{S\labelcref{ass:samplingRegime})}-\textbf{(S\labelcref{ass:mixing})}), we are equipped to bound this quantity. 
    
    First partition the grid $D_n$ into blocks with area $d_n^2$ yielding $L_n$ blocks. 
    For $z \in \{ 0, 1 \}^2$, let 
    $type(i)$ be whether or not sample $i$ lies in some block $l$ corresponding to `type z', or spatial locations with block $l$ having the 1st coordinate being even/odd (0/1) and the second coordinate being even/odd (0/1). This permits us to employ the independent blocking from Bernstein on $\B_{z,l}=\{ \Xouti \in \B_{l} \mbox{ and }type(i)=z, \forall l  \}$, or the data samples that are of type $z$ and lie in $\B_l$. We follow the strategy from \cite{KuznetsovMohri17}: We first show that the average deviation can be bounded in terms of independent blocks, and thus i.i.d. tools can be readily applied. Mcdiarmid's inequality and Rademacher averages along with Dudley's integral will be employed to control block function complexities.

    \textbf{Bound on Independent Block Copy in terms of mixing quantities.}
    Let $\B_{z,l}'$ be independent copies of $\B_{z,l}$, and $\B_z$ be the blocks $\B_{z',l}$ with $z=z'$ (and $\B_z'$ similarly).
    By Proposition 1 of \cite{KuznetsovMohri17}, we have for an $M-$bounded function $h$ and fixed $z$,
    \begin{align*}
        \E[h(\B_{z}')-h(\B_{z})] \leq L_n \cdot |D_n| \sup_{s,s'}\overline{\beta}_{s,s'}(d_n) 
    \end{align*}
     Take $h$ to be the block averages for block type $z$, or $h(\B_{z})=\E[\frac{1}{|\B_{z}|}\sum_{l} \sum_{i \in \B_{z,l}} \Tinst (f_A(\Xouti, \outCoeftx_s)-\E f_A(\Xouti, \outCoeftx_s))]$. Then it follows from the union bound that
    \begin{align*}
        \pr[\frac{1}{n} \sum_{i =1}^n \Tinst &(f_A(\Xouti, \outCoeftx_s) - \E f_A(\Xouti, \outCoeftx_s) > \delta] \\
        & \leq \sum_z \pr[h(\B_z')-\E h(\B_z') > \delta - \E h(\B_z')] + L_n \cdot \sup_{s,s'}\overline{\beta}_{s,s'}(d_n) 
    \end{align*}
    Now operating on these independent blocks, we can prove deviation inequalities using tools from the independently distributed setting. In particular, we apply a standard concentration bound and then an empirical process bound.

    \textbf{Concentration Inequality.}
    First, a one \textit{block} difference yields an empirical deviation of $M \frac{d_n}{n}$. By Mcdiarmid's inequality, 
    \begin{align*}
        \pr[\frac{1}{|\B_z'|} \sum_{i \in \B_z'} \Tinst (f_A(\Xouti, \outCoeftx_s) - \E f_A(\Xouti, \outCoeftx_s) > \delta] & \leq \exp{(\frac{-2 |D_n|^2 (\delta+\E f_A(\Xouti, \outCoeftx_s))^2}{L_n d_n^2 M^2})}
    \end{align*} 
    
    \textbf{Empirical Process Bound.}
    It remains to bound $\E h(\B_z')$. By symmetrization and Dudley's integral,
    \begin{align*}
        \E h(\B_z') &\leq \E \sup_{\outCoeftx_s} \frac{1}{|\B_z'|} \sum_{l} \sum_{i \in B_{z,l}' }\sigma_i f_A(\Xouti, \outCoeftx_s) \\
        & \lesssim \frac{1}{\sqrt{n}} 
    \end{align*}
    for $\sigma_i$ i.i.d. Rachemacher r.v.'s. The last line follows since $\outCoeftx_s$ is finite-dimensional.

    Integrating out the tail and employing \refAssump{S}{\labelcref{ass:mixing}}, we have :
    $$\E_n[\etaEst_j -\eta_j] \lesssim \sqrt{\frac{1}{|D_n|}} + d_n \sqrt{\frac{1}{|D_n|}}$$

    We can choose $d_n$ such that $\frac{d_n}{|D_n|}=o_p(1)$, or $d_n$ grows slower than $D_n$. Noting that we took $\rho_0 = 1$, the result follows. 

    
\end{proof}

     This (and our Section \labelcref{thm:qLearning}, \cref{thm:aLearning}) can be easily extended to nonparametric regression models to encompass splines, polynomial, fourier, and other bases. If we consider nonparametric models, we require our function class to be Donsker and formalized conditions on the approximation error from the optimal linear projection onto $k$ bases \cite{belloni2015}.

\subsection{Proof for \texorpdfstring{\cref{thm:RegretBd}}{Theorem}}\label{pf:RegretBd}
\begin{proof}
    \begin{align*}
        \E[Y_i(\hat{\pi}_j^*)-Y_i(\pi_j^*)] &= \E[\frac{1}{J} \sum_{j=1}^J \Tinst f_A(\Xouti; \outCoeftx)(\hat{\pi}_j^*-\pi_j^*)] \\
        &= \E[\frac{1}{J} \sum_{j=1}^J \Tinst f_A(\Xouti; \outCoeftx)[\I{\etaEst_j < 0}- \I{\eta_j < 0}] \\
        &= \frac{1}{J} \sum_{j=1}^J \Tinst \E[f_A(\Xouti; \outCoeftx)\I{\etaEst_j < 0}]-\E[f_A(\Xouti; \outCoeftx)\I{\eta_j < 0}] 
    \end{align*}
    We consider the inner expectation:
    \begin{align*}
        \E[f_A(\Xouti; \outCoeftx)\I{\etaEst_j < 0}] &-\E[f_A(\Xouti; \outCoeftx)\I{\eta_j < 0}] \\
        & = \E[f_A(\Xouti; \outCoeftx) \mid \I{\etaEst_j < 0}] \cdot \pr[\etaEst_j < 0] - \E[f_A(\Xouti; \outCoeftx) \mid \eta_j < 0] \cdot \pr[\eta_j < 0]] 
    \end{align*}
    By \cref{thm:TE_Consistency}, if $\eta_j < 0$, $\I{\etaEst_j<0}$ occurs with high probability (converging at $\sqrt{n}$ rate)
    . Thus, our estimated policy converges to the optimal policy.

\end{proof}


\subsection{Propensity Score Helper Lemma}
For \cref{thm:aLearning}, we require consistent estimation and asymptotic normality of propensity score model $e$. We state this in the following helper lemma.

\begin{lemma}\label{thm:can_intCoef}

Assume the setup from \cref{thm:aLearning}. Then we have consistent estimation and asymptotic normality of propensity score model $\pi$. I.e.
        $$\sqrt{J}(\hat{\intCoef}-\intCoef_0) \xrightarrow{d} N(0, \Omega_\varepsilon) $$ 
        where for $\Omega_\varepsilon = \Lim{J \rightarrow \infty} \sum_{j=1}^J \varepsilon_j \varepsilon_j^\top$ and $\varepsilon_j$ independent mean zero random vectors.
\end{lemma}
\begin{proof}
    The proof follows identically to \cref{thm:qLearning}.
\end{proof}

\subsection{Additional Helper Results}
The following is Lemma 10.5 from \citep{kosorok2008introduction}, restated here for convenience.
\begin{theorem}[Multiplier Central Limit Theorem]\label{thm:mclt}
    Let $Z_1 \dots Z_n$ be i.i.d. euclidean random vectors with $\E Z=0$ and finite second moment, independent of i.i.d. sequence of real random variables $\zeta_1 \dots \zeta_n$ mean 0 and variance 1. Then, conditionally on $Z_1 \dots Z_n$, $\frac{1}{\sqrt{n}} \sum_{i=1}^n \zeta_i Z_i \xrightarrow{d} N(0, \mathrm{Cov}(Z))$ for almost all sequences $Z_1 \dots Z_n$
\end{theorem}
\clearpage

\section{Descriptive Statistics}
We compute basic descriptive statistics on our outcome and intervention dataset.

\setcounter{table}{0}
\begin{table}[ht!]
\begin{tabular}{lll}
Variable                             & Mean   & Range          \\ \hline
Total NO$_x$ controls               & 2.94  & (0, 24)        \\
log(Heat input)                     & 14.45 & (8.98, 17.32) \\
log(Operating time)                  & 7.20  & (5.39, 8.93)   \\
\% Operating capacity                 & 0.64  & (0.07, 1.17)   \\
\% Selective non-catalytic reduction  & 0.26  & [0, 1]         \\
ARP Phase II                        & 0.71  & \{0, 1\}  \\ \hline
Scrubbed                        & 0.19  & \{0, 1\}  \\ \hline
\end{tabular}
\caption{\textmd{Summary of intervention level covariates from power plant data.}
\label{tab:pp_covs}}
\end{table}

\begin{table}[ht!]
\begin{tabular}{lll}
Variable                            & Mean   & Range          \\ \hline
\% White                                  & 0.89  & (0, 1)         \\ 
\% Female                                    & 0.56  & (0, 1)         \\
\% Urban                                    & 0.38  & (0, 1)         \\
\% High school graduate                  & 0.36  & (0, 1)         \\
\% Poor                                & 0.13  & (0, 1)   \\
\% Moved in last 5 years                     & 0.42  & (0, 1)         \\
\% Households occupied                      & 0.87  & (0.015, 1)      \\
\% Smoke                                & 0.26  & (0.13, 0.44)   \\
Mean Medicare age                           & 74.94 & (68.96, 96.26)    \\
Mean temperature (K)                        & 287.69 & (275.34, 301.14) \\
Mean relative humidity (\%)                & 0.0087 & (0.0045, 0.016) \\
log(Population)                                & 8.16  & (1.39, 11.65)  \\
Population per Square Mile                             & 1123.78  & (0.067, 158,503.38)  \\
\hline
IHD Hosp. Rate (per 10k PY)                           & 0.29  & (0, 0.35)         \\ 
IHD Hosp. Count                            & 22.50  & (0, 557)         
\end{tabular}
\caption{\textmd{Summary of outcome level covariates from Medicare Beneficiary data.} 
\label{tab:zip_covs}}
\end{table}

\section{Additional Simulation Details}\label{sec:simDetails}

\subsection{Simulation Parameters}
The simulation parameters are found in the supplement code, and summarized below:
\begin{align*}
    \outCoef=( &-0.000955,0.0288,0.0382,-0.000148,-0.00227,0.0167,-0.0199,\\
    &0.0396,0.0152,0.0173,-0.0119,0.0161,0.0329,0.0365,\\
    &-0.0203,0.0221,-0.0181,-0.0262,2.170e-05,0.0305,-0.0310,\\
    &0.0357,-0.0187,0.00968,0.0204,0.0269,0.00420,0.00231,\\
    &-0.000344,-0.000490,0.000127,0.00115,0.00140,0.00118,0.00133,\\
    &0.00120,-0.000423,-0.000867,0.000361,-0.00135,-0.001362,5.567e-05,\\
    &0.000982,-0.000606,0.000586,-0.00121,-0.000864,-0.000517,\\
    &-0.00135,-0.000558,0.00103,0.00106,0.00117,-0.000676)
    \\
    \intCoef=(&-0.681,0.131,-0.704,0.386,0.334,0.424,\\
    &0.00141,-0.00471,0.010,-0.0140,-0.0107,-1.449)
\end{align*}

\subsection{Simulation Metrics}
The specific formulas for the metrics in the simulation are as follows:

The parameter error for the treatment effect function $f_A$ is given by
        \begin{equation} \label{eq:betaError}
            \Err = ||\hat{\outCoeftx}-\outCoeftx_0||_2
        \end{equation}
and the RMSE
        \begin{equation} \label{eq:teError} 
            \TEErr = \mbox{RMSE}(\EstTotEff, \TotEff)
        \end{equation}






\section{Data Preprocessing}\label{sec:data_preprocessing}

\subsection{Intervention Unit Trimming}
\begin{figure}[ht!]
\centering
    \includegraphics[width=.95\textwidth,trim={0 1.3cm 0 0},clip]{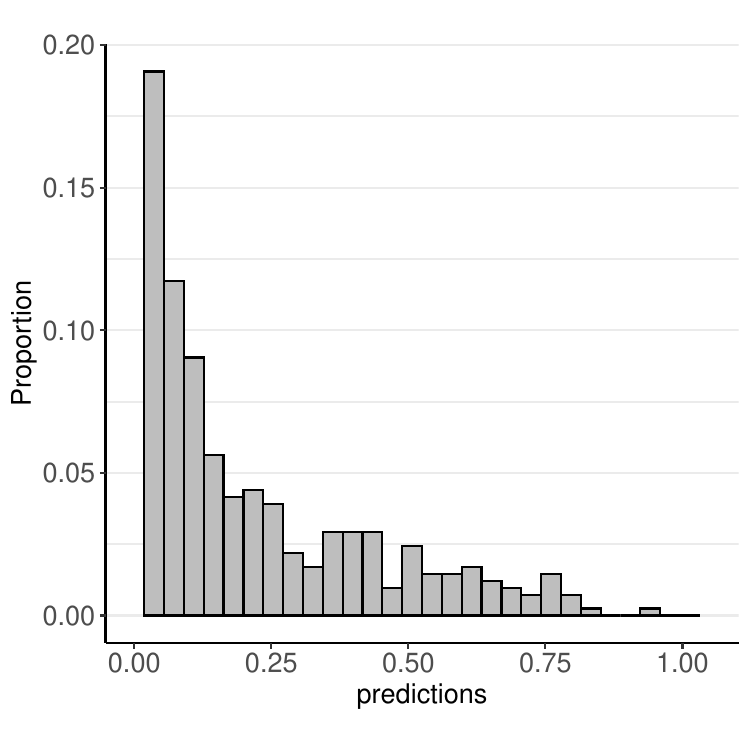} 
    \caption{\textmd{Plot of the estimated intervention unit propensities.}\label{fig:piHatHist}}
\end{figure}

We trim intervention units with low treatment probabilities. The histogram for fitted intervention unit probabilities are shown in Fig. \labelcref{fig:piHatHist}.

\clearpage 
\subsection{Cost Model Training}\label{sec:CostModel}
We select our model based on Normalized Mean Absolute Error (NMAE). For a model $\M$ with predictions $\hat{C}_j=\E[C_j \mid \Xintj; \M]$, the NMAE is calculated as follows.


\begin{equation}\label{eq:NMAE}
    NMAE(\M)=\frac{1}{J}\sum_{j=1}^J \frac{|C_j-\hat{C}_j|}{|\hat{C}_j|}
\end{equation}

\cref{tab:NMAE} shows the NMAE for each cost model. 

\begin{table}[ht!]
\begin{center}
\begin{tabular}
{ |p{2.25cm} p{2.25cm}|  }
 \hline
 \multicolumn{2}{|c|}{Cost Model Results} \\
 \hline
 $\M$ & NMAE \\
 \hline
 LM &  0.91  \\
 SVM &  0.55  \\ 
 RF &  0.54   \\ 
 \hline
\end{tabular}
\caption{\textmd{Cost model NMAE on the validation set.}\label{tab:NMAE}}
\end{center}
\end{table}

\begin{figure}[ht!]
\centering
     \includegraphics[width=.9\textwidth]{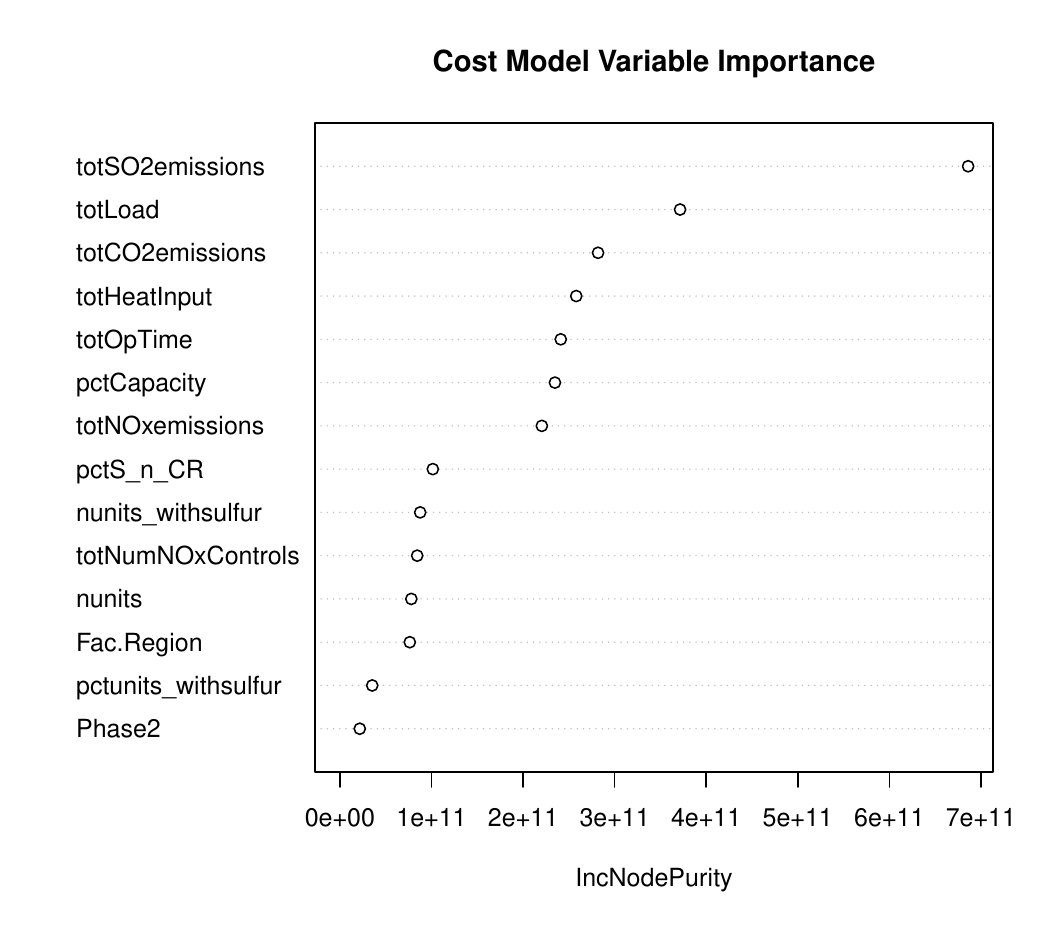} 
     \caption{\textmd{Plot of the variable importances for our cost model.}\label{fig:varImpPlot}}
\end{figure}

The random forest model performs best on our validation dataset, so random forest is used to generate cost predictions. The variable importance plot (under Increase in Node Purity) is shown in Fig. \labelcref{fig:varImpPlot}.
We find \TotSOt and  and \TotLoad yield high variable importance for predicting cost, which aligns with the scientific domain.

\begin{figure}[ht!]
\centering
\resizebox{\textwidth}{!}{
\begin{tabular}{cc}
     \includegraphics[width=.75\textwidth]{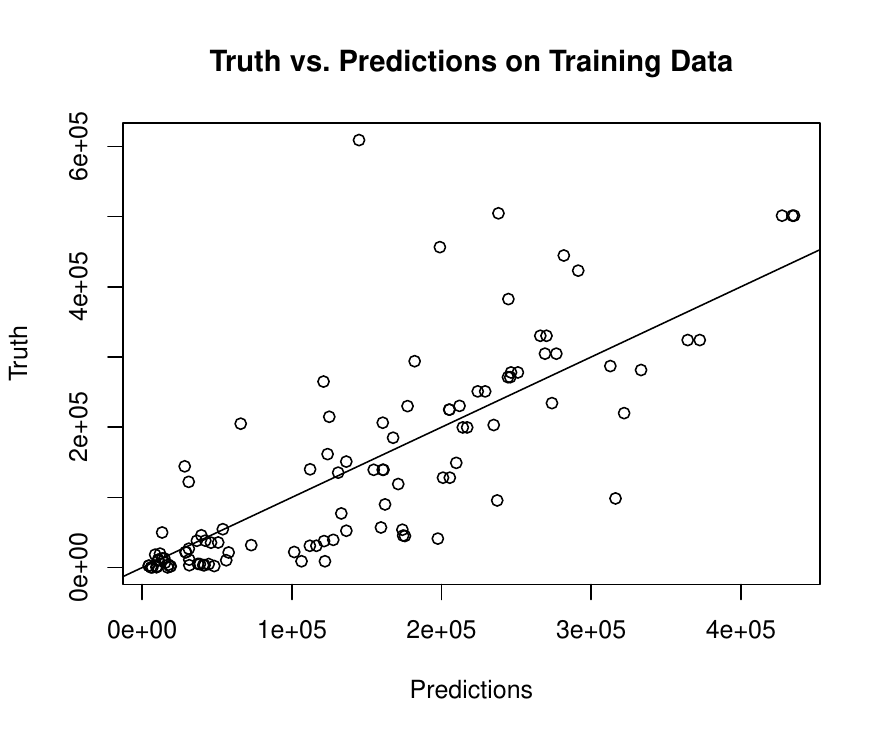} 

     &  \includegraphics[width=.75\textwidth]{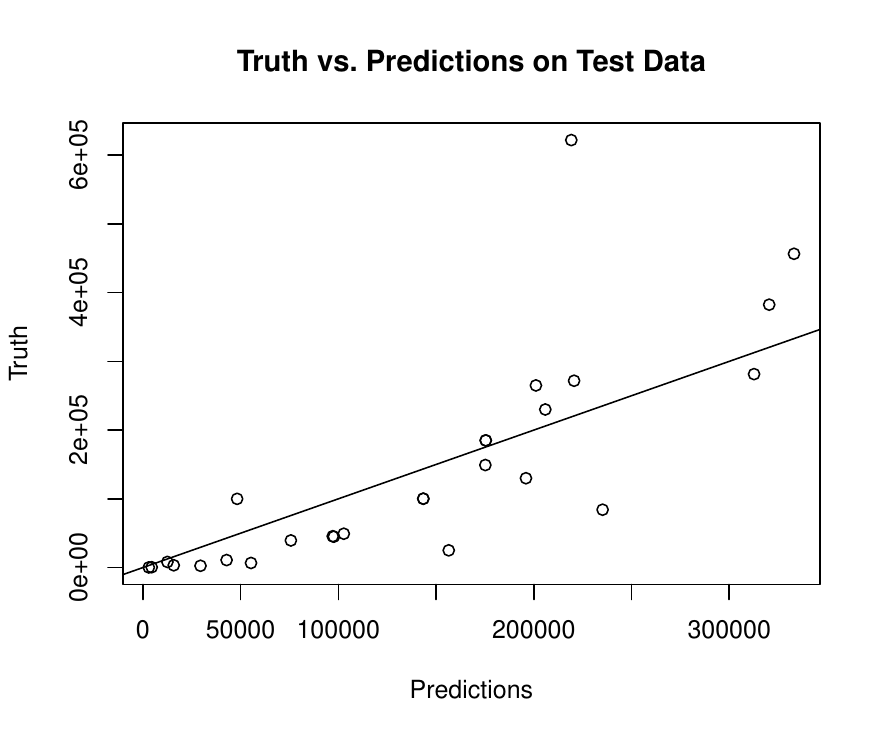} \\

     (a)  & (b)
\end{tabular}
}
\caption{\textmd{(a) displays the predictions against truth in our training set and (b) displays the predictions against the truth in our validation set.}\label{fig:predsTruth}}
\end{figure}

In \cref{fig:predsTruth}, we display the random forest model predictions against the true values, on the training and validation set. 

\clearpage

\section{Additional Analysis Details}
\subsection{Exploratory Analysis of \texorpdfstring{$\EstTotEff$}{}}
\label{sec:pvals_toteffj}
\begin{figure}[ht!]
\centering
\includegraphics[width=.7\textwidth]{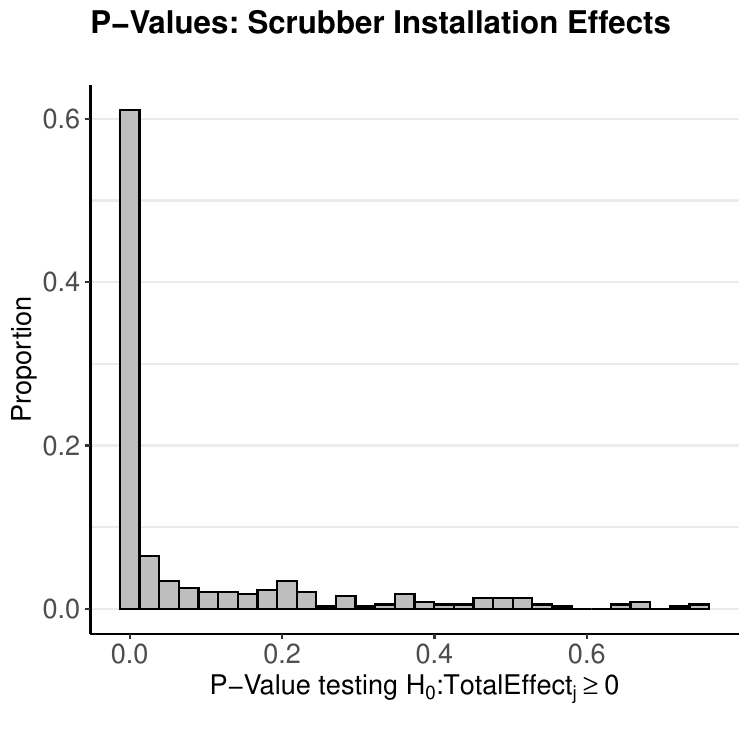} 
\caption{\textmd{P-Values from testing $H_0: \TotEff_j \geq 0$.}
\label{fig:pvals_toteffj}}
\end{figure}

We perform an exploratory analysis of $\TotEff_j$ testing $H_0: \TotEff_j \geq 0, H_a: \TotEff_j < 0 $. The p-values for each of the $J$ power plants are shown in \cref{fig:pvals_toteffj}. Approximately 70\% of p-values are $< 0.05$. Caution should be taken in interpreting these findings, as dependence and false discovery should be accounted for in a formal statistical test. Regardless, these findings support that there are statistically significant population treatment effects with each scrubber installation.

\subsection{Nonlinear Policy Analysis}
\label{sec:sens_analysis_generalize}

In Section \labelcref{sec:rwd}, we performed policy analysis using linear models akin to Section \labelcref{sec:simulation}. In this section, we present a brief overview of results using nonlinear models. As shown in \cref{tab:Nonlinear_rwd}, the potential reduction in hospitalization rates and counts do not increase with the complexity.

\begin{table}[!ht]
\begin{center}
\begin{tabular}
{ |p{2.5cm}p{2.25cm}p{2cm}|  }
 \hline
 \multicolumn{3}{|c|}{Estimated Reductions} \\
 \hline
 Model & Rates & Counts \\
 \hline
 Linear &  \ratemax & \cntmax \\
 Quadratic & 33.58 & 22,309 \\  
 Cubic & 45.87 & 31,958 \\ 
 Sin/Cos & 67.28 & 50,990 \\
 \hline
\end{tabular}
\caption{\textmd{Estimated reductions in IHD hospitalization rates per 10k person-years and IHD hospitalization counts, varying the model link, with no budget constraints.}\label{tab:Nonlinear_rwd}}
\end{center}
\end{table}

\subsection{Count Analysis}
\label{sec:analysisCounts}
The following are figures akin to \cref{fig:panelBudgetRate}, but for IHD Hospitalization Counts. The counts are obtained by multiplying the rates by the number of person-years observed for each zip code. There are more plants that are similarly protective in the northeast and southeastern US. Similar to the cost-constrained rates analysis, large gains are experienced from 10\% -- 30\% of universal scrubber treatment cost. In addition, from 60\%, much of the gain stabilizes.

\begin{figure}[ht!]
\centering
\resizebox{\textwidth}{!}{
\begin{tabular}{cc}
     \includegraphics[width=.7\textwidth]{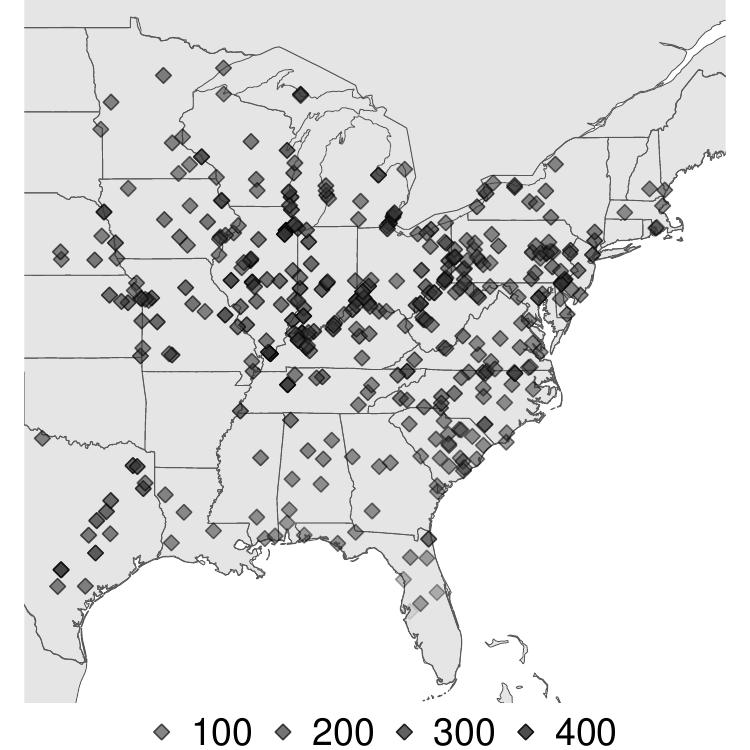} 

     &  \includegraphics[width=.75\textwidth]{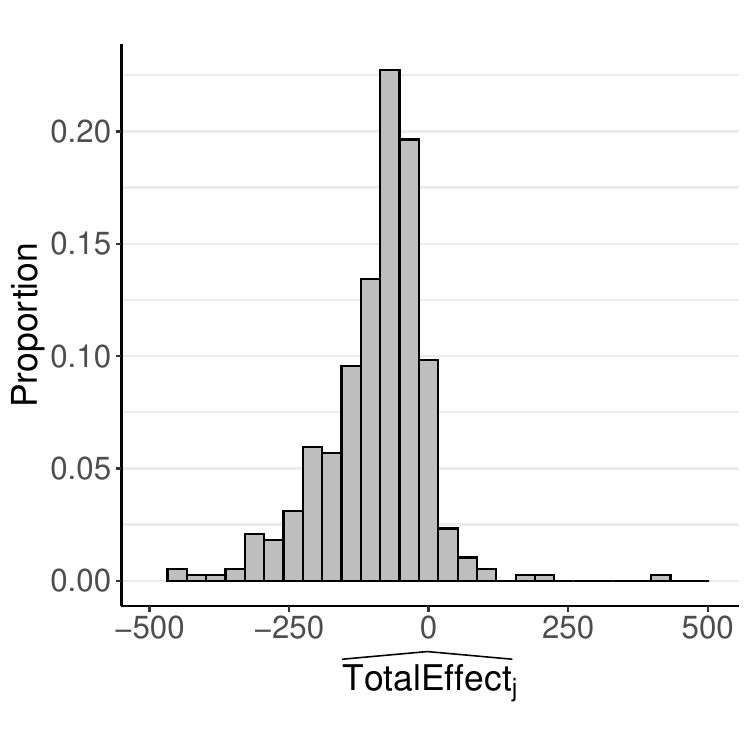} \\

     (a)  & (b)
\end{tabular}
}
\caption{\textmd{Count Results. (a) displays the plot of $\EstTotEff_j$ on a US map, colored by intensity of  $\EstTotEff_j$. (b) displays the histogram of $\EstTotEff_j$.} \label{fig:TE_Counts}}
\end{figure}

\clearpage
\begin{figure}[ht!]
\centering
\resizebox{\textwidth}{!}{
\begin{tabular}{ccc}
    \includegraphics[width=0.33\textwidth]{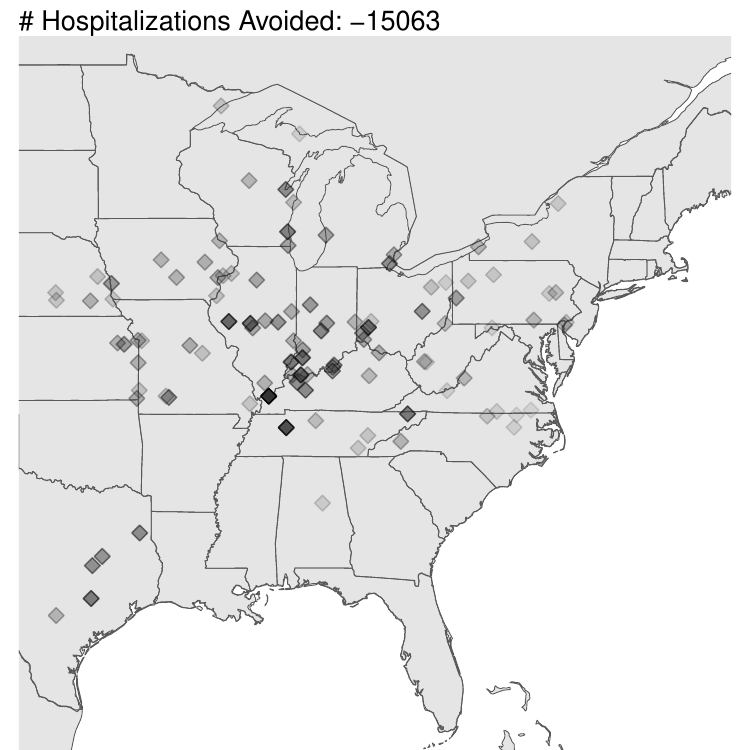}
     &
     \includegraphics[width=0.33\textwidth]{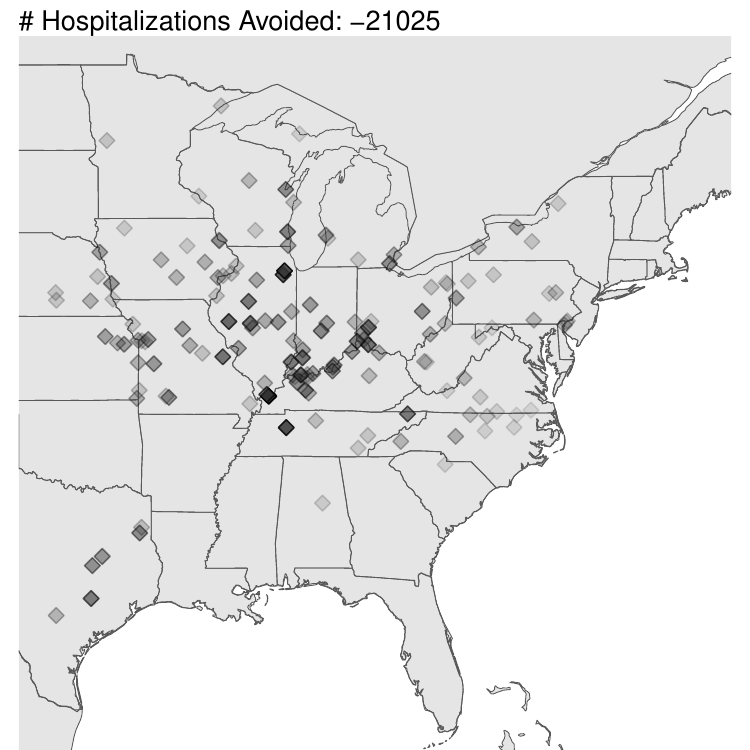} &

     \includegraphics[width=0.33\textwidth]{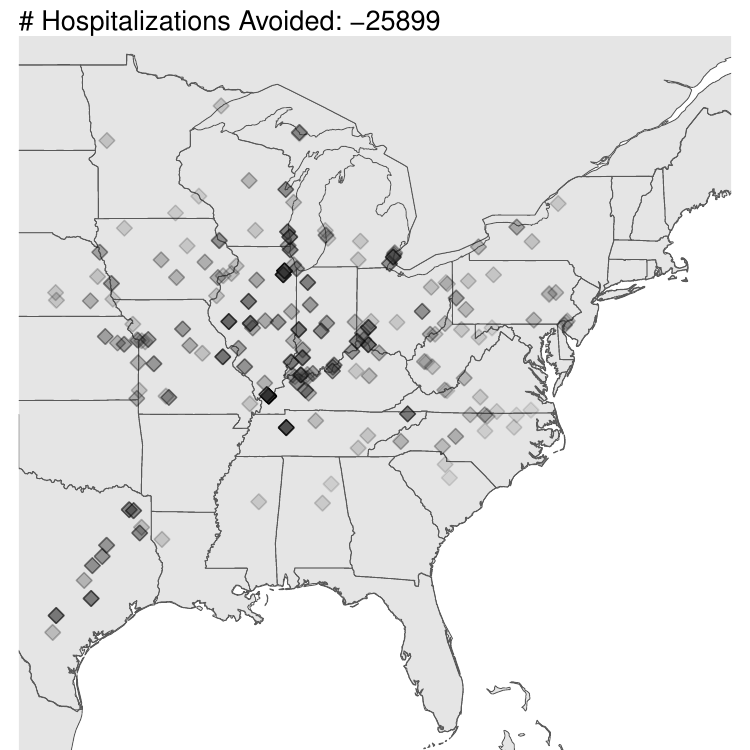}
     
     \\
     (a) 10\% Total Cost & (b) 20\% Total Cost  & (c) 30\% Total Cost 
\end{tabular}}

 \resizebox{\textwidth}{!}{
\begin{tabular}{ccc}
    \includegraphics[width=0.33\textwidth]{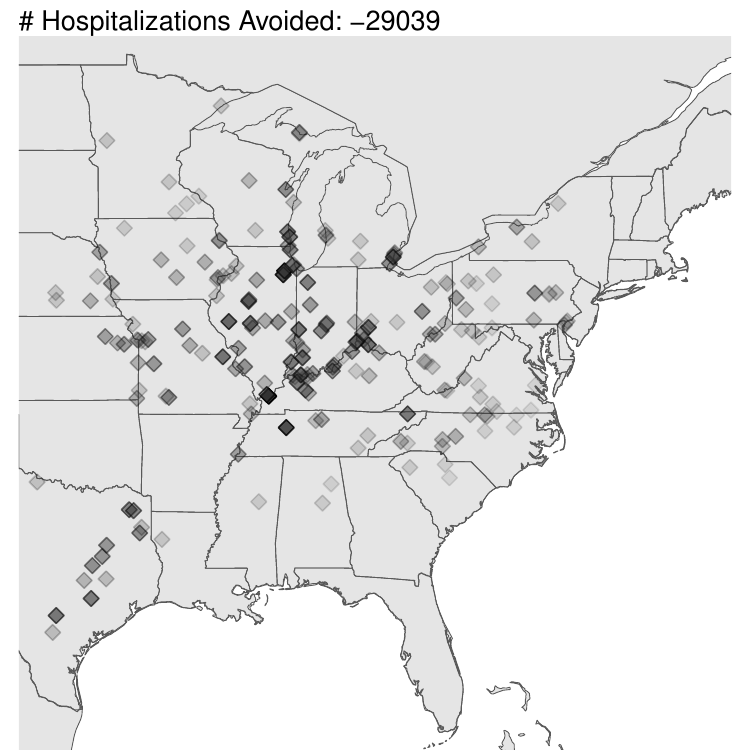}
     &
     \includegraphics[width=0.33\textwidth]{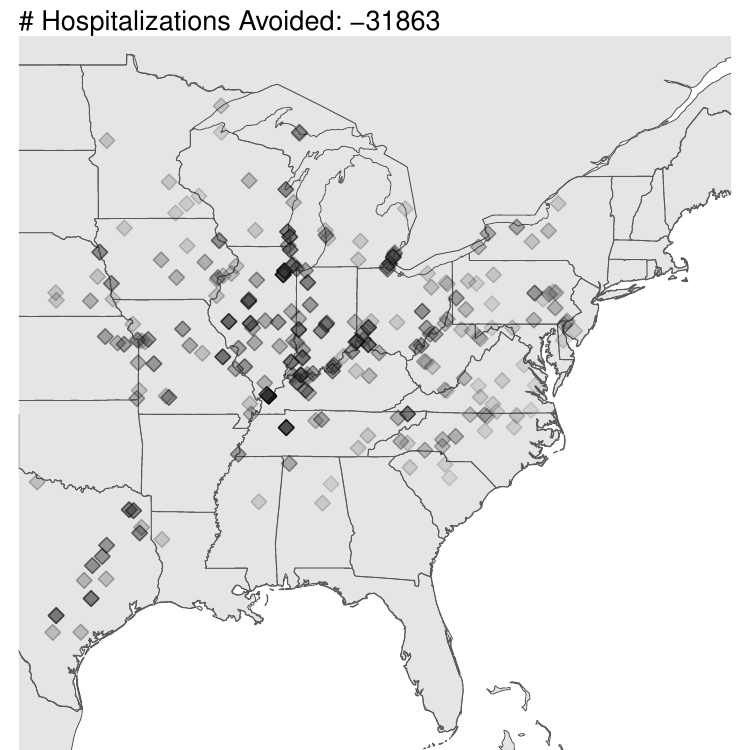} &

     \includegraphics[width=0.33\textwidth]{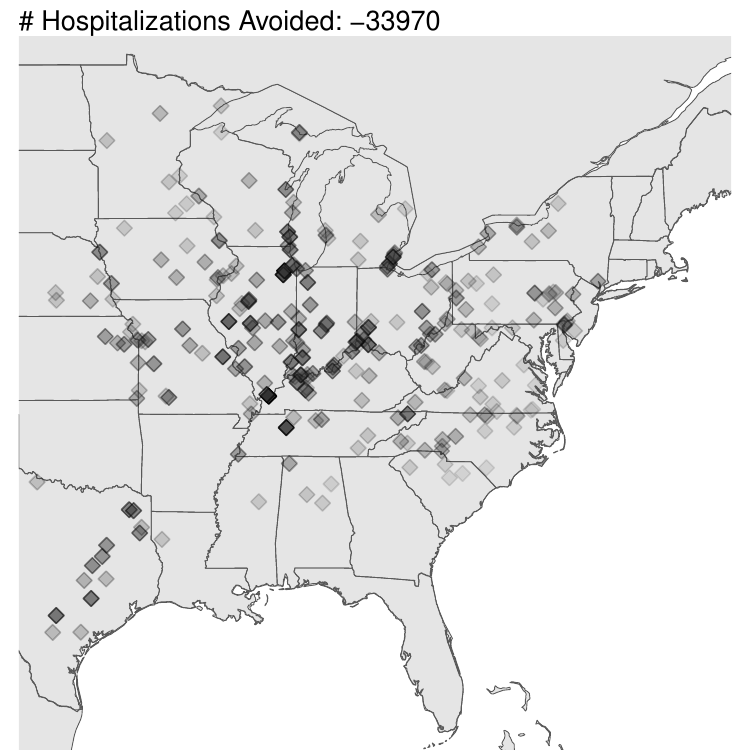}
     
     \\
     (a) 40\% Total Cost & (b) 50\% Total Cost  & (c) 60\% Total Cost 
\end{tabular}}

 \resizebox{\textwidth}{!}{
\begin{tabular}{ccc}
    \includegraphics[width=0.33\textwidth]{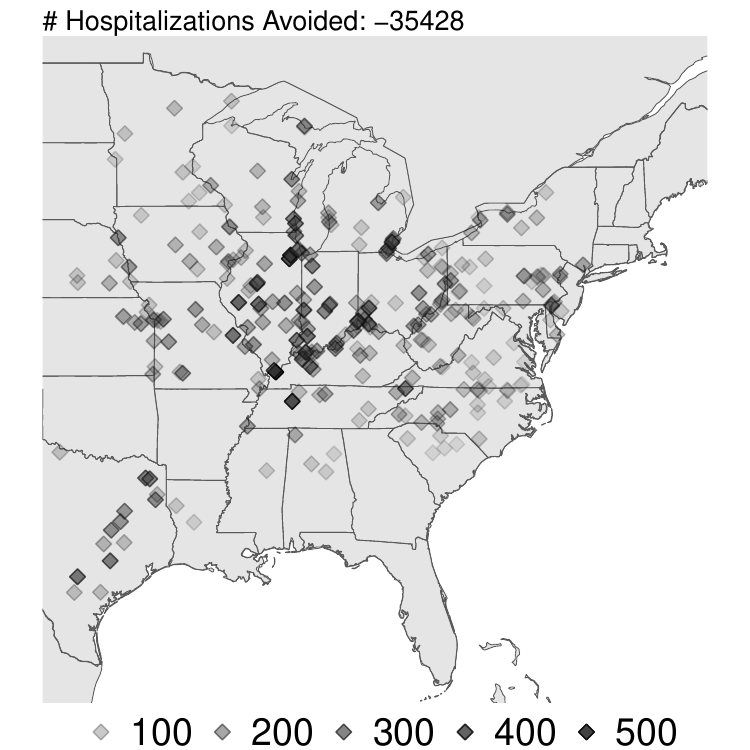}
     &
     \includegraphics[width=0.33\textwidth]{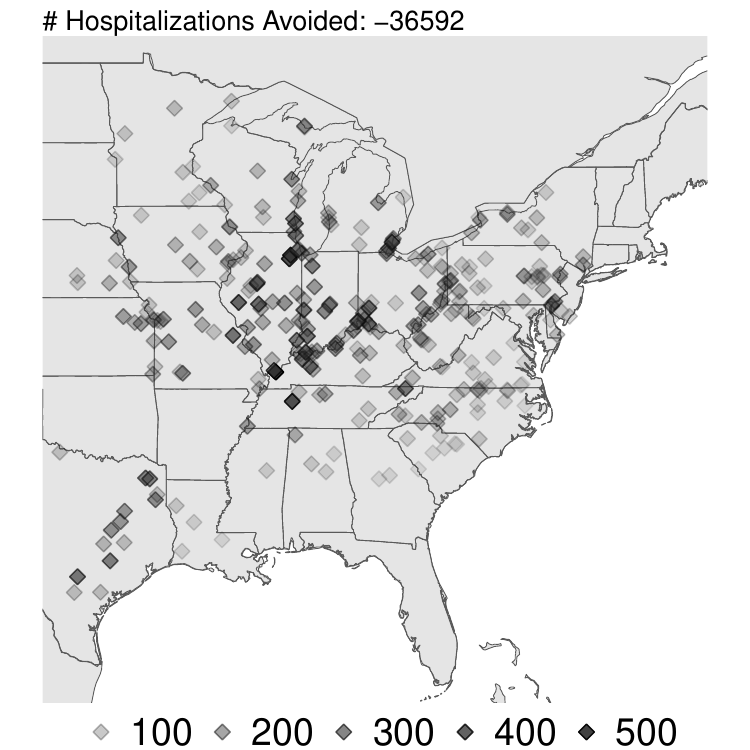} &

     \includegraphics[width=0.33\textwidth]{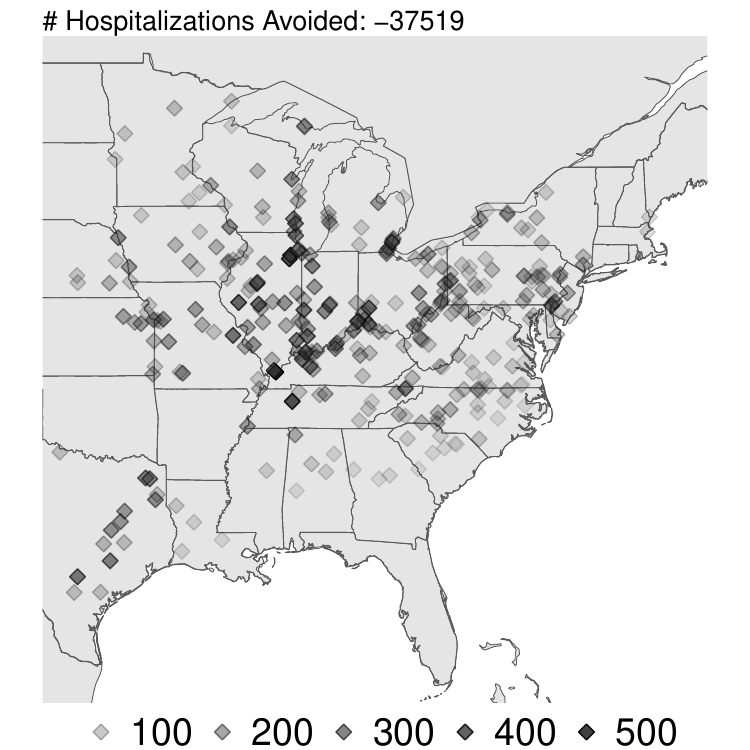}
     
     \\
     (a) 70\% Total Cost & (b) 80\% Total Cost  & (c) 90\% Total Cost 
\end{tabular}}
\caption{\textmd{Count Results. Grid with the reduction of IHD Hospitalizations, varying the spending from 10\%-90\% of budget.} \label{fig:panelBudgetCount}}
\end{figure}

\clearpage
\subsection{Optimal policy building upon the factual 2005 scrubber landscape}\label{sec:addOnAnalysis}
\begin{figure}[ht!]
\centering
 \resizebox{\textwidth}{!}{
\begin{tabular}{ccc}
     \includegraphics[width=0.33\textwidth]{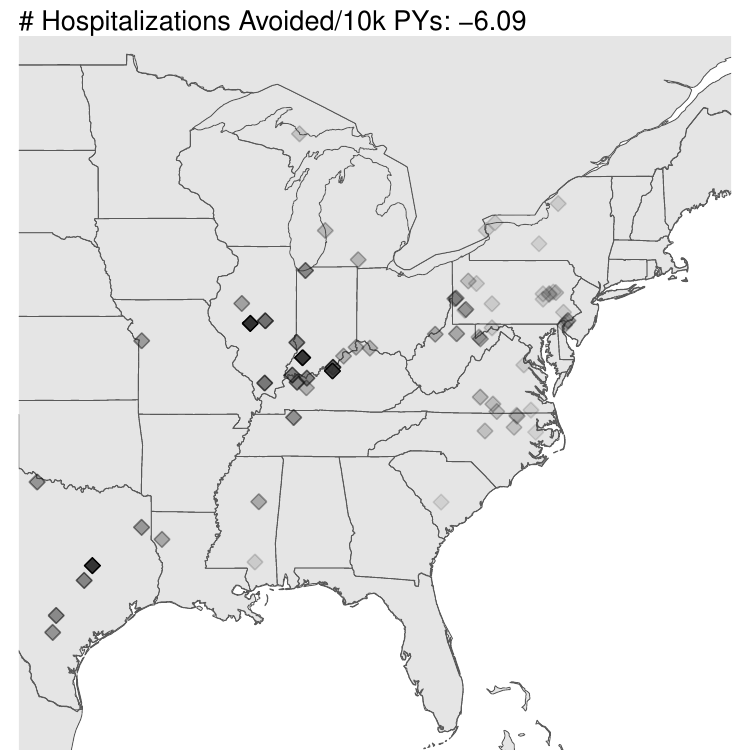}
     &
     \includegraphics[width=0.33\textwidth]{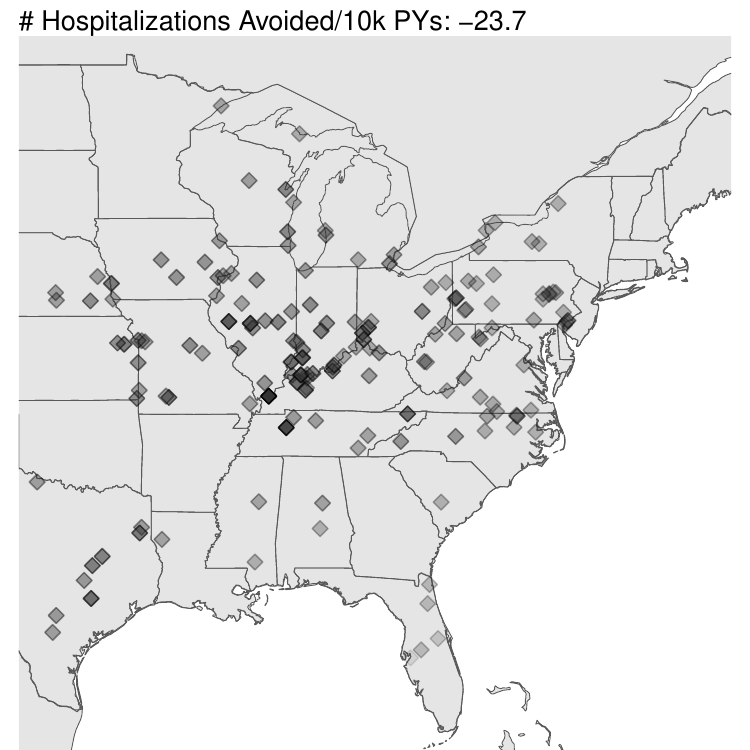}
    &
     \includegraphics[width=0.33\textwidth]{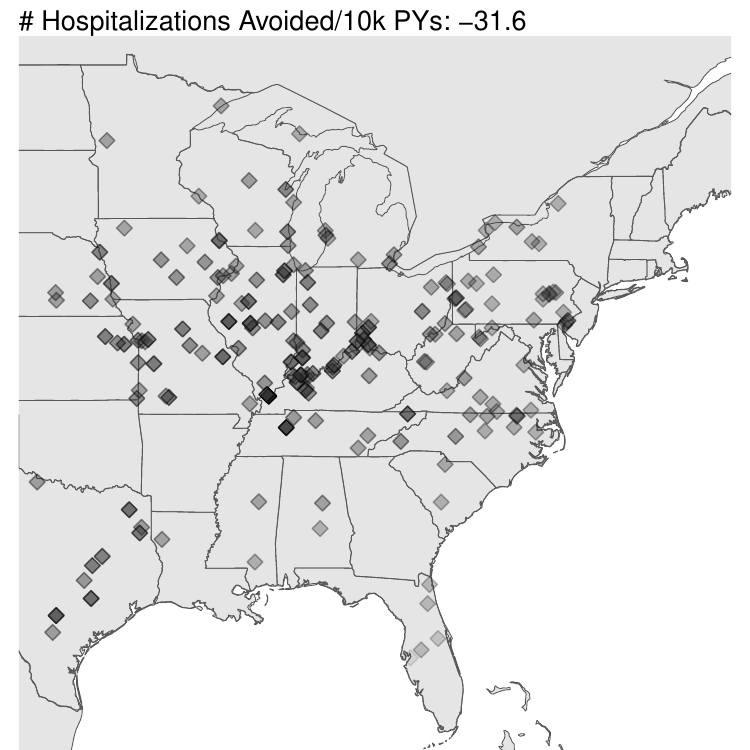}
     \\
     (a) 10\% Total Cost & (b) 20\% Total Cost  & (c) 30\% Total Cost 
\end{tabular}}

 \resizebox{\textwidth}{!}{
\begin{tabular}{ccc}
     \includegraphics[width=0.33\textwidth]{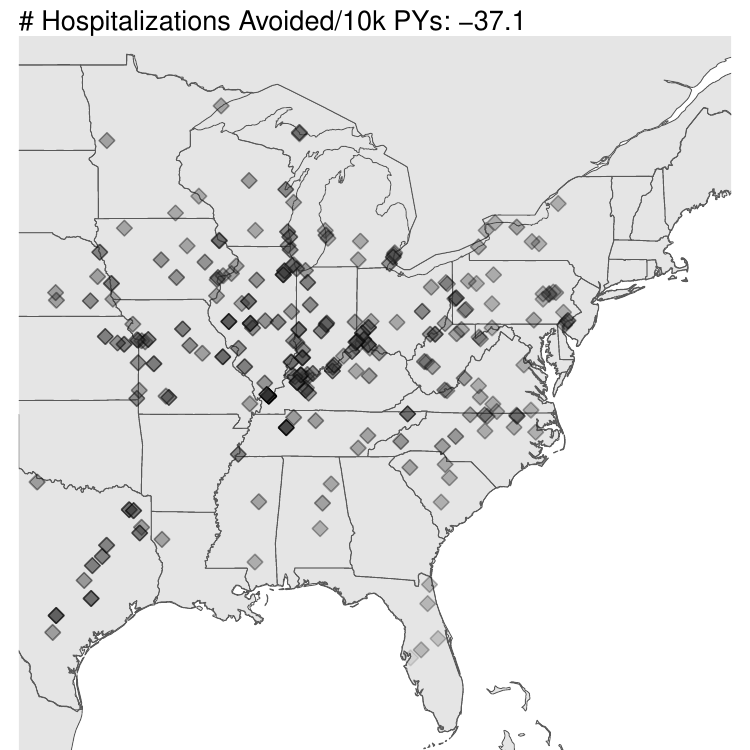}
     &
     \includegraphics[width=0.33\textwidth]{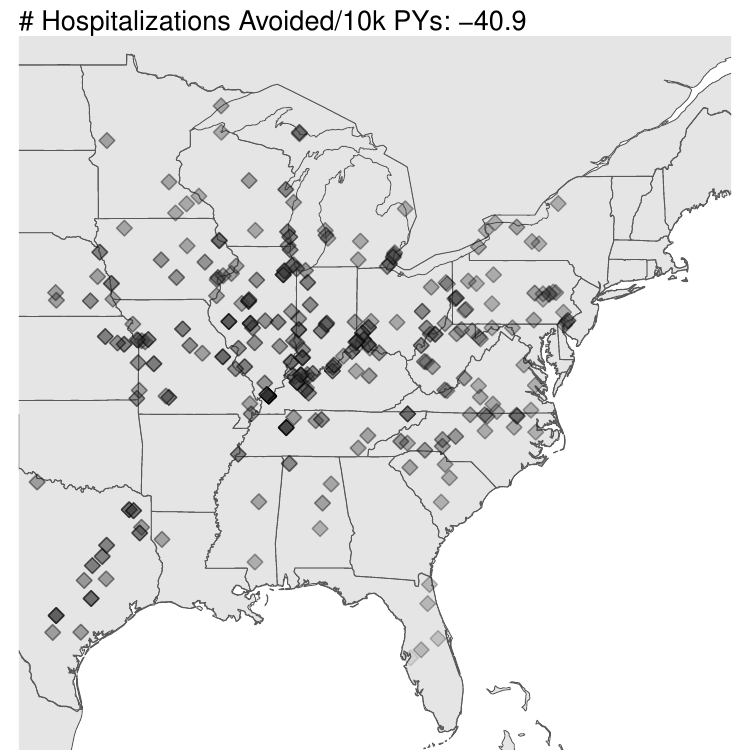}
    &
     \includegraphics[width=0.33\textwidth]{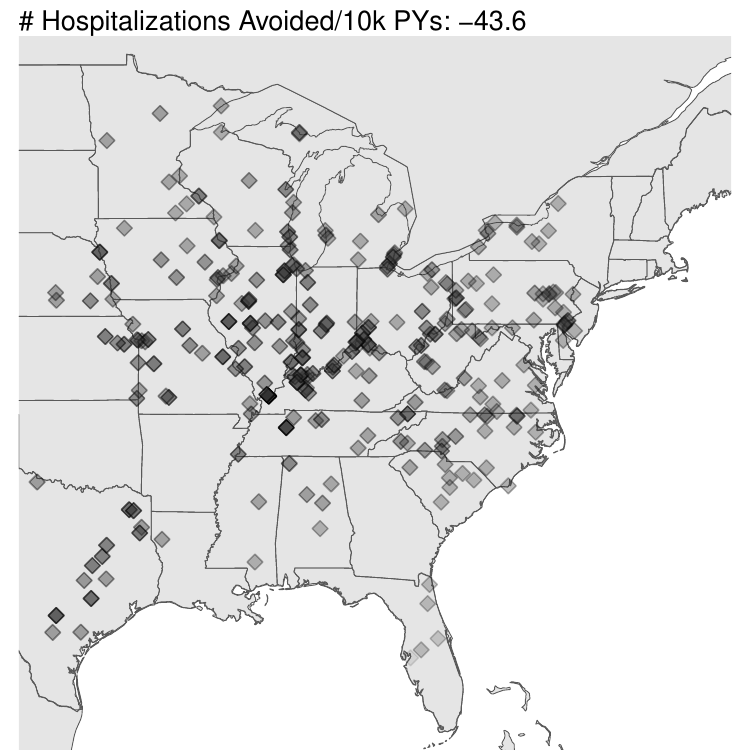}
     
     \\
     (a) 40\% Total Cost & (b) 50\% Total Cost  & (c) 60\% Total Cost 
\end{tabular}}

 \resizebox{\textwidth}{!}{
\begin{tabular}{ccc}
     \includegraphics[width=0.33\textwidth]{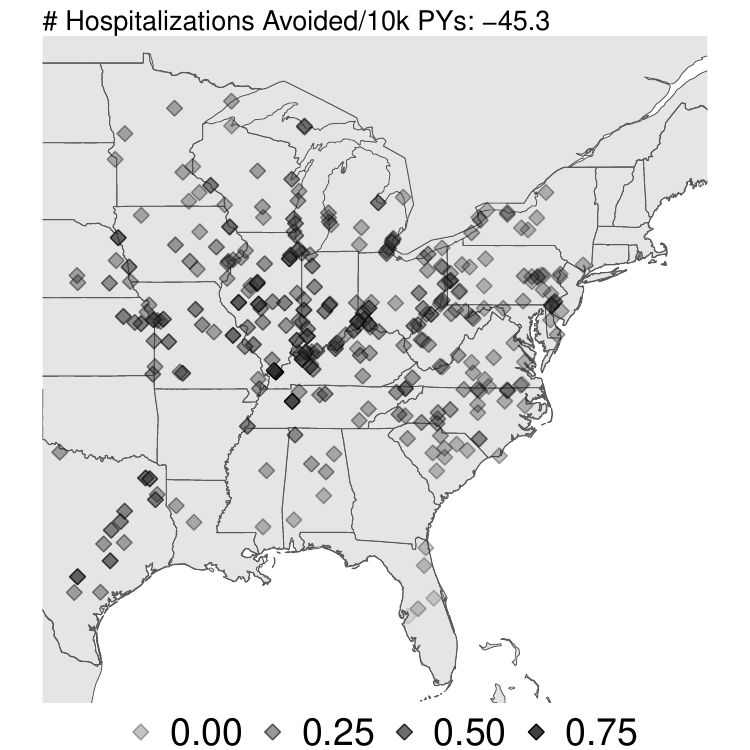}
     &
     \includegraphics[width=0.33\textwidth]{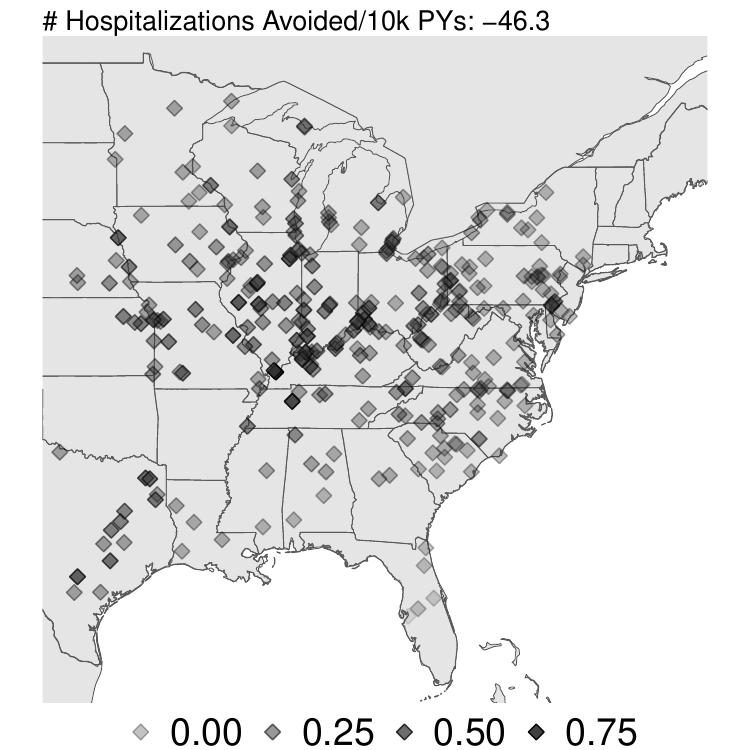}
    &
     \includegraphics[width=0.33\textwidth]{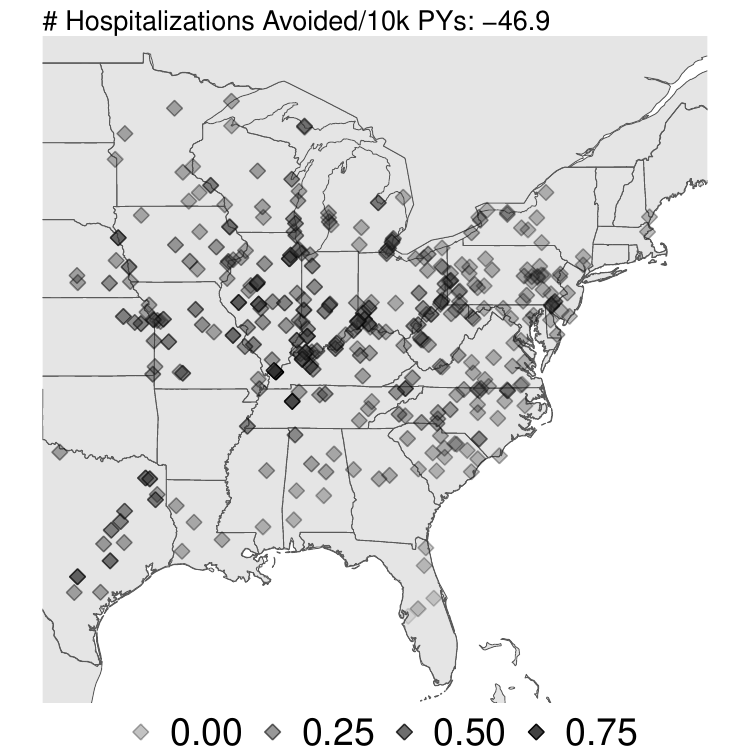}
     
     \\
     (a) 70\% Total Cost & (b) 80\% Total Cost  & (c) 90\% Total Cost 
\end{tabular}}
\caption{\textmd{Grid with the reduction of IHD Hospitalizations/10,000 Person Years, varying the spending from 10\%-90\% of budget, when considering the augmentation policy.} \label{fig:panelBudgetRateAddOn}}
\end{figure}

In this section, we consider the alternative to analysis presented in Section \labelcref{sec:rwd}, examining policies starting from the factual scrubber distribution, only allowing for addition of new scrubbers. We refer to this as the \textit{augmentation policy}. We do not consider the possibility of removing existing scrubber installations because (1) we do not have data on the cost and effects of doing so, and (2) policymakers may not consider such an action since it may be unethical given the established links between air pollution and adverse public health outcomes. 
Because the factual scrubber policy costs approximately 10\% of the universal scrubber installation cost, we consider the optimal cost-constrained policy under 20\%, 30\% \dots 90\% of the universal cost in \cref{fig:panelBudgetRateAddOn}. The factual policy is listed in the 10\% slot, and the 100\% treated policy is identical to the 100\% policy in our main analysis.

For further comparison, we contrast the clean-slate analysis and augmentation policy in \cref{tab:addOnComparisonRate}. We see that the augmentation policy is inefficient along the range of cost constraints considered, with notable gains experienced using the clean slate analysis under lower cost constraints. As we increase the cost to around 40-100\%, the rate reductions are very similar.

\begin{table}[ht!]
\centering
\begin{tabular}{|lll|}
\hline
\%$C$  &  Clean Slate &  Augmentation \\ \hline
10\%  & 20.9  & 6.1 \\
20\%  & 29.0  & 23.7  \\
30\%  & 34.9  & 31.8  \\
40\%  & 39.0  & 37.1  \\
50\%  & 42.2  & 40.9  \\
60\%  & 44.4  & 43.6  \\
70\%  & 45.9  & 45.3  \\
80\%  & 46.8  & 46.3  \\
90\%  & 47.4  & 46.9 \\
\hline
\end{tabular}
\caption{\textmd{Rate reductions by cost constraint (\% of universal scrubber cost $C$) for the clean-slate analysis and the augmentation policy.}
\label{tab:addOnComparisonRate}}
\end{table}

\subsection{Optimal Policy vs. Factual Policy}\label{ss:opt_vs_fact}
\begin{figure}[ht!]
    \centering
    \resizebox{\textwidth}{!}{
    \begin{tabular}{cc}
         \includegraphics[width=.7\textwidth]{images_v1/senssubsetStates_rates_empirical_0.1.pdf}
    
         &  \includegraphics[width=.7\textwidth]{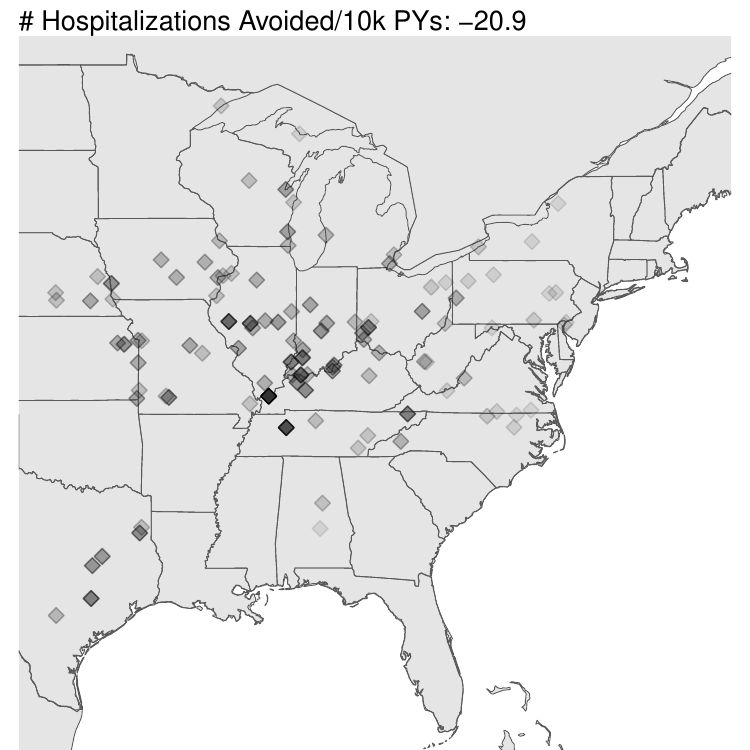} \\
    
         (a) Empirical Policy & (b) Optimal Policy
    \end{tabular}
    }
    \caption{\textmd{Empirical and Optimal Policy Comparison, under equivalent budget constraints.}
    \label{fig:empiricalPolicyCmp}}
    \end{figure}

In this subsection, we compare the factual 2005 scrubber installation scheme and the optimal ``clean slate'' policy under the same cost constraint ($\sim 10\%$ of the universal cost). As shown in \cref{tab:addOnComparisonRate} and \cref{fig:empiricalPolicyCmp}, the clean-slate analysis yields notable improvement in IHD hospitalization reductiosn compared to the factual policy. We can see on the map in \cref{fig:empiricalPolicyCmp}, the factual policy treats 73 plants, compared to the optimal clean-slate policy, which treats 117 plants in similar regions and yields higher IHD hospitalization rate reductions. This indicates lower efficiency in the factual policy, since it chooses more expensive and less health benefiting power plants. In fact, there are 27 power plants overlapping between the optimal policy and factual policy, so the factual policy did not choose many that the optimal policy would have chosen under this cost constraint.

\subsection{Sensitivity Analyses on Plants without Scrubbers newly installed in 2005.}\label{sec:sensScrubbersPrior2005}
Of the 74 power plants in our dataset with scrubber installations by 2005, only 4 (1\%) of these power plants had scrubbers newly installed in year 2005. This raises the potential concern for temporality of outcomes occurring prior to treatment since outcomes were measured in 2005. In this section, we carry out sensitivity analyses excluding these 4 power plants to test how sensitive our results may be to potential violation of this assumption. Specifically, we analyze how our $\TotEff_j$ estimation differs. 

\begin{figure}[ht!]
    \centering
    \resizebox{\textwidth}{!}{
    \begin{tabular}{cc}
         \includegraphics[width=.7\textwidth]{images_v1/sensHistogram_TEs.pdf}
    
         &  \includegraphics[width=.7\textwidth]{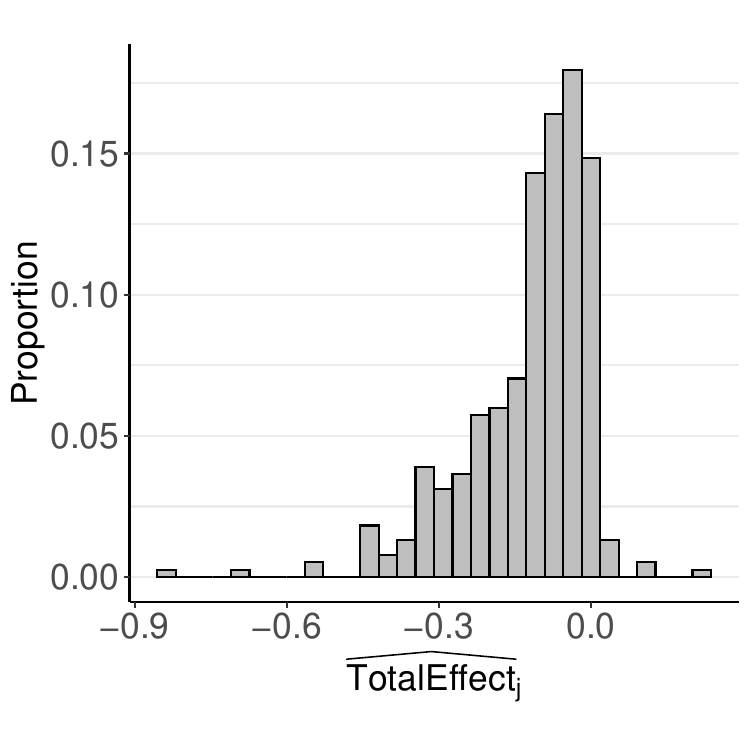} \\
    
         (a) Include 4 plants (current analysis) & (b) Exclude 4 plants (sensitivity analysis)
    \end{tabular}
    }
    \caption{\textmd{Comparison of $\EstTotEff_j$ when we include the 4 plants that have scrubbers installed in the year 2005 (left), and exclude the 4 plants (right).} \label{fig:sens4Plants}}
    \end{figure}

Rerunning our analysis without these plants, we obtain nearly identical rate and count reductions (identical up to 0.1 for rates, and identical up to 1 for counts) -- \cref{fig:sens4Plants} supports this claim showing nearly identical histogram plots of $\EstTotEff_j$.
    
Based on this analysis, we have reasonable confidence that the temporal, causal order is respected since the majority of power plants have scrubbers installed prior to year 2005 (when we measure outcomes) and the results are not sensitive to exclusion of these 4 power plants.
    
\section{Implementing the Policy on Future Units}\label{sec:newUnits}
Suppose we were interested in recommending scrubber installations for a new set of power plants. In order to carry out this analysis, we will need an updated, or even possibly new, bipartite adjacency matrix $\T'$ (HyADS matrix) that holds information on the dependencies between the new units $(\Xintj)', (\Xouti)'$ yielding sample sizes $J', n'$. Then we can take the following approaches under no cost-constraints and cost-constraints.

\subsection{Policies under No Cost Constraints.}
    Recall the policy for power plant $j$ under no cost constraints is given by 
    $$
    \hat{\pi}_j^*(\TotEff(\hat{\outCoeftx})_j) = \I{\TotEff(\hat{\outCoeftx})_j < 0}
    $$

    For a future power plant $j'$, we can estimate the policy in the same manner:
    \begin{equation}\label{eq:futurePolicy}
    \hat{\pi}_{j'}^*(\TotEff(\hat{\outCoeftx})_{j'}) = \I{\TotEff(\hat{\outCoeftx})_{j'} < 0}
    \end{equation}
    where $\TotEff(\hat{\outCoeftx})_{j'}= \frac{1}{J'} \sum_{i=1}^{n'}  H_{ij'}' f_A((\Xouti)'; \hat{\outCoeftx}) $

\subsection{Policies under Cost Constraints.} Recall that under cost-constraints, we solve \cref{eq:optObjCost}. This is equivalent to choosing to treat power plants with largest $\EstCostBenefit_j$ until our cost-constraint $C$ is hit. Denote the rank of power plant $j$'s $\EstCostBenefit_j$ by $(j)$. Denote the last power plant that is able to be treated under cost constraint $C$ by $j^*_C$. Then the power plant that has the next highest $\EstCostBenefit$ (power plant $(j^*_C)+1$) is treated with probability proportional to the remaining budget. In other words, the policy is given by:
\[
    \hat{\pi}_{j}^*(\CostBenefit(\hat{\outCoeftx})_j) = \begin{cases}
        1, & \text{if }(j) \leq (j^*_C) \\
        \frac{C-\sum_{l: l \leq (j^*_C)}c_{l}}{c_{j}}, & \text{if }(j) = (j^*_C) + 1\\
        0, & \text{Otherwise.}
    \end{cases}
\]

Then, we may consider a similar adaptation to policy making on a future power plant $j'$, recalculating the last power plant that is able to be treated under cost-constraint $C$ accounting for $j'$. Denote this power plant by $j^{**}_{C}$. 
\[
    \hat{\pi}_{j'}^*(\CostBenefit(\hat{\outCoeftx})_{j'}) = \begin{cases}
        1, & \text{if }(j) \leq (j^{**}_{C}) \\
        \frac{C-\sum_{l: l \leq (j^{**}_{C})}c_{l}}{c_{j}}, & \text{if }(j) = (j^{**}_{C}) + 1\\
        0, & \text{Otherwise.}
    \end{cases}
\]



\section{Model Generalizations and Extensions}\label{sec:generalizations}

\subsection{Multiple Timepoints}
If we have $n$ samples over $t \in [T]$ timepoints, we may follow the same approach as in \cite{Schulte_2014}. 
For illustration purposes, we once again assume the outcome model at each timepoint factorizes by intervention unit as in \cref{eq:meanModel}:
    \begin{align*}
    Y_{it}(\bar{a}_{it}) &= \mu_t(\bar{a}_{it}; \mathbf{X}_{it}^{out}, \T_i, \outCoef_{0t})+\epsilon_{it}\\
        &= f_{0t}(\mathbf{X}_{it}^{out}, \outCoefb_{0t})+ \bar{a}_{it} f_{At}(\mathbf{X}_{it}^{out}, \outCoeftx_{0t}) +\epsilon_{it} \numberthis \label{eq:meanModel_k}
    \end{align*}
Above, each quantity is the analog as described in \cref{eq:meanModel}, but indexed a second time for timepoint $t \in [T]$.


\subsection{Q-Learning} We propose to solve the following estimating equation backwards in time $t=T, T-1 \dots 1$.
\begin{equation}\label{eq:qLearningEE_k}
     \frac{1}{n} \sum_{i=1}^n \phi_t(Y_{it}; \outCoef)=0 
\end{equation}

\noindent where $\phi_t(Y_{it}; \outCoef_t)=(Y_{it} - \mu_t(\bar{A}_{it}; \mathbf{X}_{it}^{out}, \T_i, \outCoef_t))d_{it}(\outCoef_t)$, $d_{it}(\outCoef_t) = \frac{\partial \mu_t(\bar{A}_{it}, \mathbf{X}_{it}, \T_i, \outCoef_t)}{\partial \outCoef_t}$.

Specifically, at any timepoint $t$, we obtain $\hat{\pi}^*_t(\EstTotEffjk)$. Thus at time $t-1$, we have the optimal policy value assuming policymakers make the optimal policy choice at time $t$. With backwards induction, model-correctness, and straightforward extensions of the assumptions mentioned throughout Section\labelcref{sec:methods}, this strategy can be shown to be valid.

\subsection{\al}
The result follows similarly to above. Extending the estimating equations in question \cref{eq:alearning_alpha}, \cref{eq:alearning_gamma}, \cref{eq:alearning_beta} to $T$ timepoints and solving for these backwards in time yields a multi-time point extension to A Learning.

\subsection{Nonparametric Models} 
The approaches above can be extended to more flexible models assuming nuisance parameters are estimated at $\sqrt{n}$ consistent rate. 
Valid inference would follow from Section \labelcref{thm:qLearning} and \cref{thm:aLearning} since higher order terms would decay to 0 fast enough. In particular, one may employ nonparametric series estimators for functions that satisfy smoothness conditions \citep{gineNickl}. 

\subsection{Mean Outcome Form} 
The proposed framework permits use of more general mean outcome models than that considered in \cref{eq:meanModel}. \cref{eq:meanModel} enables us to obtain closed-form or efficient solutions to \cref{eq:optObjCost}. In general, we may require higher computational burden depending on the assumed outcome form. Interpretability also goes down since we would not be able to directly interpret \TotEff in terms of the outcome model as in \cref{eq:TEj}.


\end{appendices}

\end{document}